\documentclass[11pt]{article}

\usepackage[final]{acl}


\usepackage{times}
\usepackage{latexsym}
\usepackage{fontawesome5}
\usepackage{amsfonts}
\usepackage{graphicx}
\usepackage{subcaption}
\usepackage{tabularx}
\usepackage{booktabs}
\usepackage{array}
\usepackage{amsmath}
\usepackage{longtable}
\usepackage{booktabs}
\usepackage{todonotes}
\usepackage{multirow}
\usepackage{enumitem}
\usepackage[most]{tcolorbox}
\usepackage{booktabs}
\usepackage{multirow}    % only if used elsewhere
\usepackage{subcaption}
\usepackage{graphicx}    % for \resizebox

\usepackage{todonotes}

\newtcolorbox{promptbox}[1][Prompt]{%
  enhanced,
  breakable,
  colback=gray!5,
  colframe=gray!55,
  boxrule=0.4pt,
  arc=2pt,
  left=5pt,right=5pt,top=4pt,bottom=4pt,
  fonttitle=\bfseries\footnotesize,
  coltitle=black,
  colbacktitle=gray!18,
  attach boxed title to top left={xshift=6pt,yshift=-2pt},
  boxed title style={colframe=gray!55,boxrule=0.4pt,arc=1pt},
  title=#1,
}
\newtcolorbox{responsebox}[1][Model Response]{%
  enhanced,
  breakable,
  colback=blue!2,
  colframe=blue!35!gray,
  boxrule=0.4pt,
  arc=2pt,
  left=5pt,right=5pt,top=4pt,bottom=4pt,
  fonttitle=\bfseries\footnotesize,
  coltitle=black,
  colbacktitle=blue!8,
  attach boxed title to top left={xshift=6pt,yshift=-2pt},
  boxed title style={colframe=blue!35!gray,boxrule=0.4pt,arc=1pt},
  title=#1,
}

\usepackage[T1]{fontenc}
\usepackage{booktabs}
\usepackage[utf8]{inputenc}

\usepackage{microtype}
\usepackage[table]{xcolor}
\providecommand{\nacell}{}
\renewcommand{\nacell}{\cellcolor{gray!15}\textit{N/A}}

\newcommand{\dpos}[1]{\textcolor{blue}{\textbf{+#1}}}
\newcommand{\dneg}[1]{\textcolor{red}{\textbf{#1}}}
\newcommand{\dzero}{0.00}
\usepackage{inconsolata}

\usepackage{graphicx}

\title{The Bottleneck Is Retrieval, Not Reading: Diagnosing Evidence-Based Generation with RAG for Long Context Question Answering}
\author{
 \textbf{Debopriyo Banerjee\textsuperscript{1\thanks{Equal contribution.}}},
 Owais Aijaz\textbf{\textsuperscript{2\footnotemark[1]}},
 Ahmer Tabassum\textbf{\textsuperscript{2\footnotemark[1]}},
 \textbf{Zhuohan Xie\textsuperscript{2}}
\\
 \textbf{Preslav Nakov\textsuperscript{2}},
 \textbf{Veselin Stoyanov\textsuperscript{2}}
\\
 \textsuperscript{1}Inception42
 \quad
 \textsuperscript{2}MBZUAI\\
 \texttt{
\{deb.ban89, vesko.st\}@gmail.com
 }
\texttt{preslav.nakov@mbzuai.ac.ae} \\
 \faGlobe~\href{https://debobanerjee.github.io/LC-EBG-project/}{Project}
 ~~\faGithub~\href{https://github.com/debobanerjee/LC-EBG}{Code}
}

\begin{document}
\maketitle

\begin{abstract}

% version - 3
Retrieval-augmented generation (RAG) is the dominant paradigm for grounding large language models in external corpora, but its similarity-driven, chunked pipeline has structural blind spots: it cannot anticipate which passages a question requires and discards document-level context that resolves cross-passage references. We introduce \emph{evidence-based generation} (EBG), requiring models to return an answer with line-level citations, and a joint \texttt{Ans+Evid} metric that scores correctness and grounding simultaneously. We reformulate two corpora into evidence-grounded multi-document QA benchmarks~-~\textsc{MultiHop-RAG} and \textsc{QASPER-MP}, a multi-paper adaptation of QASPER with realistic short paper cues, and pair them with a controlled synthetic two-hop haystack and an evidence-augmented NoLiMa spanning 10K--1M characters. Across ten proprietary frontier and three open-weight LLMs, EBG isolates two RAG limitations that long-context models avoid: similarity retrieval surfaces \emph{any} relevant chunk far more reliably than \emph{all} required chunks for multi-fact queries, and chunking destroys document meta-information needed to resolve identity-bearing queries. EBG also exposes grounding failures invisible to answer-only evaluation, including incomplete citations in two-needle settings and a widening gap between \texttt{Ans} and \texttt{Ans+Evid} when context contradicts parametric knowledge. Data and code are available at \url{https://github.com/debobanerjee/LC-EBG}.

\end{abstract}

\section{Introduction}

% Large Language Models (LLMs) possess a remarkable capability of rapidly processing vast amounts of text and identifying the information most relevant to a given query. This is particularly well suited to applications such as legal discovery, medical record analysis, financial due diligence, literature review, and technical support, all of which require synthesizing targeted answers from evidence distributed across large document collections.

% Large Language Models (LLMs) can rapidly process vast amounts of text and identify the information most relevant to a given query. This capability suits applications such as legal discovery, medical record analysis, financial due diligence, literature review, and technical support, all of which require synthesizing targeted answers from evidence distributed across large document collections.

Large Language Models (LLMs) can rapidly process large text volumes to identify information relevant to a query. This capability suits applications such as legal discovery, medical analysis, financial due diligence, literature review, etc., which require synthesis from large document collections.

% The predominant paradigm for grounding LLM outputs in external knowledge is retrieval-augmented generation \citep[RAG;][]{lewis2020retrieval}. A RAG pipeline retrieves a small set of passages via embedding similarity, then conditions the generator on those passages alone. While effective for \emph{single-fact lookups}, this design suffers from several structural limitations. First, each retrieved chunk is stripped off its surrounding context: the model has no visibility into where a paragraph falls within a document or how it relates to neighboring sections. Second, document-level metadata (authorship, date, section hierarchy) is typically lost during chunking. Third, and most critically, a similarity-based retriever cannot anticipate which combination of passages will be needed to answer a question that requires integrating information from multiple locations in the corpus. This deficiency is particularly acute for \emph{multi-hop questions}, where the answer depends on chaining facts that may appear in distant, semantically dissimilar passages \citep{karpukhin-etal-2020-dense, li-etal-2024-retrieval}.

The predominant paradigm for grounding LLM outputs in external knowledge is retrieval-augmented generation \citep[RAG;][]{lewis2020retrieval}, which retrieves passages via embedding similarity and conditions the generator on those passages alone. While effective for \emph{single-fact lookups}, this design has several structural limitations. First, each retrieved chunk is stripped of its surrounding context: the model cannot see where a paragraph falls within a document or how it relates to neighboring sections. Second, document-level metadata (authorship, date, section hierarchy) is typically lost during chunking. Third, and most critically, similarity-based retrieval cannot anticipate which combination of passages will be needed for questions requiring integration across multiple locations. This is particularly acute for \emph{multi-hop questions}, where the answer depends on chaining facts that may appear in distant, semantically dissimilar passages \citep{karpukhin-etal-2020-dense, li-etal-2024-retrieval}.

% Our goal in this paper is not to advocate replacing RAG with long-context (LC) models outright, but to make these structural limitations measurable. RAG systems are typically evaluated on answer correctness alone, which conflates two very different failure modes: the retriever fails to surface the right passages, or the reader fails to use them. To separate the two, we propose \emph{evidence-based generation (EBG)}, a paradigm in which the model must produce both an answer and explicit references to the input lines that support it. EBG inherits the verifiability that RAG provides by construction~\citep{lewis2020retrieval}, but extends it to long-context settings where the model is given the full document neighborhood and must \emph{select} its own evidence rather than rely on a pre-filtered chunk set. The joint \texttt{Ans+Evid} metric this enables lets us locate where, exactly, a pipeline breaks: in retrieval, in grounding, or in reading. Recent work has shown that long-context models can match or exceed RAG accuracy when given sufficient context~\citep{li-etal-2024-retrieval, laitenberger-etal-2025-stronger}, but without provenance these comparisons are inconclusive; EBG closes that gap and lets us treat LC as a controlled probe of what RAG misses.

Our goal in this paper is not to advocate replacing RAG with long-context (LC) models outright, but to make these structural limitations measurable. RAG systems are typically evaluated on answer correctness alone, which conflates two very different failure modes: the retriever fails to surface the right passages, or the reader fails to use them. To separate the two, we propose \emph{evidence-based generation (EBG)}, a paradigm where the model must produce both an answer and explicit references to the input lines that support it. EBG inherits RAG's built-in verifiability~\citep{lewis2020retrieval}, but extends it to long-context settings where the model is given the full document neighborhood and must \emph{select} its own evidence. The joint \texttt{Ans+Evid} metric pinpoints where a pipeline breaks: in retrieval, in grounding, or in reading. Recent work shows LC models can match or exceed RAG accuracy given sufficient context~\citep{li-etal-2024-retrieval, laitenberger-etal-2025-stronger}, but without provenance these comparisons are inconclusive; EBG closes that gap.

\noindent
We make the following contributions:
\begin{itemize}[leftmargin=1em, itemsep=0pt, topsep=0pt, parsep=0pt, partopsep=0pt]
    \item \textbf{EBG framework.} We define the evidence-based generation task and a joint \texttt{Ans+Evid} metric, and evaluate ten proprietary frontier and three open-weight LLMs under matched LC and RAG conditions across 10K--1M character contexts.
    % \item \textbf{Evidence-grounded benchmarks.} We reformulate two existing corpora into evidence-grounded multi-document QA benchmarks~-~a re-instrumented \textsc{MultiHop-RAG} with line-level evidence labels, and \textsc{QASPER-MP}, a multi-paper adaptation of QASPER with realistic short paper cues~-~and pair them with a controlled synthetic two-hop haystack and an evidence-augmented version of NoLiMa. The first two stress the retrieval pipeline; the last two stress long-context reading under controlled depth and contradiction.
    \item \textbf{Evidence-grounded benchmarks.} We reformulate two corpora into evidence-grounded multi-document QA benchmarks—a re-instrumented \textsc{MultiHop-RAG} with line-level evidence labels, and \textsc{QASPER-MP}, a multi-paper QASPER adaptation with realistic paper cues, and pair them with a controlled two-hop synthetic haystack and an evidence-augmented \textsc{NoLiMa}. The first two stress retrieval; the last two stress long-context reading.
    % \item \textbf{Characterizing RAG's limitations.} Using EBG, we isolate two concrete failure modes of similarity-based RAG that LC sidesteps: (i)~similarity retrieval recovers \emph{any} relevant chunk far more reliably than \emph{all} required chunks for multi-fact queries, and (ii)~chunking discards the document meta-information that identity-bearing queries depend on. We reproduce the RAG baselines on \textsc{MultiHop-RAG} and show that the retrieval ceiling at realistic top-$k$ budgets, not the reader, is the binding constraint.
    \item \textbf{Characterizing RAG's limitations.} Using EBG, we isolate two failure modes of similarity-based RAG that LC sidesteps: (i)~retrieval recovers \emph{any} relevant chunk far more reliably than \emph{all} required chunks for multi-fact queries, and (ii)~chunking discards document meta-information that identity-bearing queries depend on. On \textsc{MultiHop-RAG}, the retrieval ceiling at realistic top-k budgets, not the reader, is the binding constraint.
    \item \textbf{Diagnostic analysis of LC grounding.} Beyond the RAG comparison, EBG reveals that citation acts as a reasoning scaffold, that models cite incomplete evidence in two-needle settings, and that grounding degrades sharply when in-context evidence contradicts parametric knowledge.
\end{itemize}

\section{Related Work}

\subsection{Long-Context Language Models}
Advances in transformer architectures have expanded context windows from thousands to hundreds of thousands of tokens through improved positional encoding such as ALiBi~\citep{press:2022:AliBi}, RoPE extensions~\citep{Su:2024:RoPE}, and memory-augmented attention~\citep{wu:2022:memformer}. Training data matters equally, as pretraining corpora rarely contain long-range dependencies; \citet{zhuang2025scaling} construct data with long-distance references encouraging use of distant context. Yet larger windows do not automatically yield better reasoning. \citet{du-etal-2025-context} show accuracy degrades with context length even under perfect retrieval, suggesting a processing bottleneck. Similarly, \citet{liu-etal-2024-lost} show performance drops when information appears mid-input rather than at boundaries.

\subsection{Evaluating Long-Context Reasoning}
Early long-context evaluations relied on needle-in-a-haystack tests \citep{kamradt2024needle}, which mainly measure retrieval of planted information rather than reasoning. Subsequent work argues that such evaluations set the bar too low, since strong performance often reflects shallow heuristics rather than genuine understanding \citep{liu-etal-2024-lost,thoma:2025:NIAHCRT,modarressi2025nolima,zhang-etal-2024-bench,yen2025helmet}. Recent benchmarks shift toward more demanding evaluations: $\infty$Bench~\citep{zhang-etal-2024-bench} studies degradation beyond 100K characters, HELMET~\citep{yen2025helmet} shows that benchmark choice substantially alters rankings, and NoLiMa~\citep{modarressi2025nolima} removes lexical shortcuts by requiring semantic matching between queries and evidence. However, these benchmarks still score models on final answers alone, leaving open whether correct answers stem from actual use of relevant context or from priors, surface cues, or chance. Unlike these answer-only benchmarks, our work focuses on grounding by requiring models to return both answers and supporting input lines, distinguishing genuine reasoning from lucky guessing and enabling a more faithful assessment of whether models use long contexts effectively and reliably.

\subsection{Retrieval-Augmented Generation}
Retrieval-augmented generation (RAG)~\citep{lewis2020retrieval} remains the dominant approach for grounding LLMs in external knowledge. Built on dense retrieval~\citep{karpukhin-etal-2020-dense} and multi-passage readers such as Fusion-in-Decoder~\citep{izacard-grave-2021-leveraging}, RAG provides provenance by construction since retrieved passages serve as evidence. Recent work revisits RAG alongside long-context models, highlighting a trade-off between the two paradigms \citep{li-etal-2024-retrieval,laitenberger-etal-2025-stronger}: long-context prompting generally improves accuracy but at substantially higher cost, motivating hybrid systems that route queries by difficulty \citep{li-etal-2024-retrieval}. Controlled comparisons also show well-tuned single-stage RAG can match elaborate multi-stage pipelines once token budgets are equalized \citep{laitenberger-etal-2025-stronger}, while adaptive retrieval and self-reflection improve factuality and citation quality \citep{asai2024selfrag}. Existing work primarily evaluates these approaches on accuracy or efficiency. We instead focus on grounding, distinguishing whether evidence was available from whether the model actually used it.

\section{Evidence-based Generation (EBG)}

We formalize \textbf{evidence-based generation (EBG)} as a grounded question-answering paradigm that extends standard answer-only evaluation by requiring models to generate both answers and explicit citations to the supporting input evidence.

\subsection{Answer-only vs Evidence-based Generation}

Existing long-context evaluations, including NoLiMa~\citep{modarressi2025nolima}, use answer-only evaluation, omitting the grounding needed to validate model reasoning.
EBG extends this formulation by requiring models to generate both explicit \textbf{evidence} references to supporting input lines and the final \textbf{answer}, yielding a stricter evaluation setting in which correctness depends on both accurate reasoning and faithful grounding. 

To illustrate the distinction, consider a haystack containing the sentence \textit{``Then Stuart mentioned that he has been vegan for years.''} at line 1,284 and the question \textit{``Which character cannot eat Brandade?''}~-~a two-hop query requiring the inference that \textit{Brandade} contains \textit{fish} and is therefore incompatible with \textit{veganism}. The answer-only and EBG settings differ in both the system instruction and the required output format.

\begin{promptbox}[Answer-only Prompt (User)]
\footnotesize\ttfamily
You will answer a question based on the following book snippet:\\[2pt]
\{haystack\}\\[4pt]
Use the information provided in the book snippet to answer the question. Your answer should be short and based on either explicitly stated facts or strong, logical inferences.\\[4pt]
Question: Which character cannot eat Brandade?\\[4pt]
Return only the final answer with no additional explanation or reasoning.
\end{promptbox}

\begin{responsebox}
\footnotesize\ttfamily
Stuart
\end{responsebox}

Under answer-only evaluation, the response \texttt{``Stuart''} is simply marked correct. But under EBG, the same response is evaluated using three complementary metrics: answer accuracy (\texttt{Ans}), evidence accuracy (\texttt{Evid}), and their conjunction (\texttt{Ans+Evid}). While \texttt{Ans} preserves standard answer correctness, \texttt{Ans+Evid} additionally requires the model to cite the supporting evidence at line 1,284. Together, these metrics separate answer correctness from evidence-grounded correctness.

\begin{promptbox}[EBG Prompt (System + User)]
\footnotesize\ttfamily
\textbf{[system]} Your job is to answer the question entirely from the context and provide a reference. Your answer should cite all lines the answer is based on.\\[4pt]
\textbf{[user]} <context>\{haystack\}</context>\\[2pt]
Answer the question based on information only from the context. If the question is not answerable from the context, answer NA. Your response should comprise only the answer and all lines the answer is based on in json format. For example:\\[2pt]
\{\,``lines'': [25, 412],``answer'': ``John''\,\}\\[4pt]
Question: Which character cannot eat Brandade?
\end{promptbox}
\begin{responsebox}
\footnotesize\ttfamily
\{\,``lines'': [1284], ``answer'': ``Stuart''\,\}
\end{responsebox}
This decomposition exposes silent failures hidden by answer-only evaluation: a model may arrive at the correct answer while citing irrelevant lines, or retrieve the correct evidence while producing the wrong answer.
The joint \texttt{Ans+Evid} metric therefore serves as the strictest measure of evidence-based reasoning, requiring both accurate reasoning and faithful grounding.

We also evaluate vanilla Chain-of-Thought (CoT) prompting, following \citet{modarressi2025nolima}, and a CoT-based EBG prompting variant. The prompt templates are in Appendix \ref{app:prompt_templates}.

% \paragraph{Metrics.}For \textit{answer-only} prompting, we report only answer accuracy (\texttt{Ans*}). For \textit{EBG} prompting, we report \texttt{Ans}, \texttt{Evid}, and \texttt{Ans+Evid} accuracies.

% Importantly, EBG does not merely add an auxiliary output; it changes the nature of the task. As shown in Section~\ref{sec:answer_only_vs_ans_evid}, requiring evidence often \emph{improves} answer accuracy, suggesting that citation acts as a reasoning scaffold rather than a post-hoc explanation.

% Importantly, EBG does not merely add an auxiliary output; it changes the nature of the task. We hypothesize that evidence (or citation) acts as a reasoning scaffold rather than a post-hoc explanation.

% \subsection{Comparison with Retrieval-Augmented Generation (RAG)}

% EBG is complementary to, but orthogonal to, retrieval-augmented generation (RAG). A standard RAG pipeline follows:
% \begin{center}
% \textit{Query} $\rightarrow$ \textit{Retrieve top-$k$ chunks} $\rightarrow$ \textit{Generate answer}
% \end{center}

% In contrast, EBG introduces an explicit grounding requirement:
% \begin{center}
% \textit{Query} $\rightarrow$ \textit{(Context provided or retrieved)} $\rightarrow$ \textit{Generate evidence} $\rightarrow$ \textit{Generate answer}
% \end{center}

% The key distinction is that \textbf{RAG defines how context is obtained}, while \textbf{EBG defines how that context is used and evaluated}. These two dimensions are therefore composable.

\subsection{RAG under EBG}
In this work, we evaluate RAG systems \emph{within the EBG framework} by requiring models to not only answer using retrieved chunks, but also to explicitly cite the supporting evidence. This yields two RAG variants:
\begin{itemize}[leftmargin=1em, itemsep=0pt, topsep=0pt, parsep=0pt, partopsep=0pt]
    \item \textbf{Answer-only RAG:} Standard formulation requiring answer correctness only.
    \item \textbf{Answer + Evidence RAG:} EBG formulation requiring correct answer and correct citation.
\end{itemize}

\noindent
This unified view allows us to decompose end-to-end performance into two factors:
\begin{itemize}[leftmargin=1em, itemsep=0pt, topsep=0pt, parsep=0pt, partopsep=0pt]
    \item \textbf{Evidence availability:} whether the retriever surfaces the relevant needle(s) (Hit@k),
    \item \textbf{Evidence utilization:} whether the model correctly uses and cites the retrieved evidence (long-context accuracy).
\end{itemize}

% \paragraph{Key differences.}
% \begin{itemize}
%     \item \textbf{Where grounding occurs:} RAG enforces grounding through retrieval, whereas EBG enforces grounding through explicit evidence generation.
%     \item \textbf{Faithfulness:} RAG alone does not guarantee that retrieved content is used; EBG explicitly requires the model to identify supporting spans.
%     \item \textbf{Evaluation signal:} RAG is typically evaluated on answer accuracy, while EBG introduces a stricter joint correctness criterion (answer + evidence).
% \end{itemize}

Overall, our formulation shows that RAG and EBG are not competing paradigms but \textbf{complementary components}: retrieval determines \emph{what evidence is available}, while EBG determines \emph{whether that evidence is correctly used and cited}.

% \subsection{EBG as a Reasoning and Evaluation Paradigm}

% Overall, EBG reframes long-context question answering as a \textbf{generate-then-ground} problem rather than a single-step prediction task. By requiring explicit evidence, it provides:
% \begin{itemize}
%     \item \textbf{Interpretability:} outputs can be directly verified against the input,
%     \item \textbf{Diagnosability:} errors can be attributed to retrieval, reasoning, or grounding failures,
%     \item \textbf{Stronger evaluation:} distinguishing correct answers from correctly grounded answers.
% \end{itemize}

% This makes EBG particularly suitable for long-context settings, where models must not only find relevant information but also demonstrate that they have done so.

\section{Benchmarks}
\label{sec:benchmarks}

To systematically evaluate EBG and RAG under EBG, we construct four complementary benchmarks: a fully synthetic benchmark, an evidence-augmented version of NoLiMa, MultiHop-RAG, and QASPER-MP (an adapted version of QASPER). Together, they evaluate EBG under both controlled synthetic settings and realistic long-context scenarios.

\subsection{Synthetic Dataset}
\label{sec:benchmark_synthetic}

The synthetic benchmark provides precise control over haystack content and evidence placement. It consists of a distractor fact pool, a curated needle set, and a haystack generation procedure. This controlled construction allows us to vary context length independently of task difficulty while retaining exact ground-truth evidence locations.

\paragraph{Distractor fact pool.}
We construct a fixed pool of \textbf{66{,}834 unique single-sentence facts} using iterative LLM prompting followed by deduplication, ensuring that distractors are not confusable with inserted needles. The large pool also limits repeated distractors across independently generated trials and provides sufficient diversity for constructing long haystacks.

\paragraph{Needle set.}
We manually curate \textbf{10 two-hop needles}, each containing: i) two \emph{atomic} supporting facts, ii) a compositional question requiring \emph{both} facts, and iii) accepted answer surface forms. The needles span relations such as \emph{company--product} associations, \emph{founder--specialty} links, and \emph{birthplace} lookups. By requiring composition across two independently placed facts, the benchmark tests both evidence retrieval and integration rather than simple single-fact recall.

\paragraph{Haystack assembly.}
For a target length $L$, distractor facts are shuffled and accumulated until reaching $L$ characters, after which the two needle facts are inserted at independently sampled positions and recorded as gold evidence. Haystacks are formatted as line-numbered \texttt{i:~fact} entries and evaluated under the EBG prompt template (\S~\ref{app:prompt_templates}). Each trial uses a newly generated haystack and a uniformly sampled needle. Consequently, performance differences cannot be attributed to a fixed ordering of distractors or evidence positions. Appendix~\ref{app:synth_dataset_example} provides a representative example.

\paragraph{Sanity check.}
At $L{=}10\text{K}$ (${\approx}2\text{K}$ tokens), all evaluated frontier models achieve at least $0.98$ joint \texttt{Ans+Evid} accuracy over $N{=}100$ trials, confirming that the inserted needles are unambiguous and readily solvable when context length is modest. Any degradation at larger $L$ therefore reflects failures in long-context evidence retrieval or integration rather than intrinsic task ambiguity. This makes the synthetic benchmark a controlled probe of how model reliability changes as relevant evidence becomes increasingly sparse within the surrounding context.

\subsection{NoLiMa-based Dataset}
\label{sec:benchmark_nolima_data}

We adapt the NoLiMa haystack construction procedure to create long-context inputs ranging from $L{=}10\text{K}$ to $1\text{M}$ characters and require models to both answer questions and cite supporting needles as evidence. This converts NoLiMa from an answer-only benchmark into an \textbf{evidence-grounded evaluation} that tests whether models truly use the provided context.

\paragraph{Needle set.}
We use the original \textbf{single-needle} (SN) setting from NoLiMa~\citep{modarressi2025nolima}, specifically the \texttt{0402} and \texttt{0405} needle groups with both default and inverse variants under the \textit{commonsense} category. Each needle contains a character name and a single factual attribute, and supports both one-hop and two-hop questions. 
% For one-hop questions (e.g., \textit{Which character is lactose intolerant?''}), the answer is stated directly in the needle. Two-hop questions (e.g., \textit{Which character cannot drink a cappuccino?''}) additionally require external world knowledge, such as the fact that \emph{cappuccino} contains \emph{milk}.

We further introduce a \textbf{two-needle} (\textit{TN}) setting, in which the \textit{explicit secondary needle} is consistent with world knowledge, and a \textbf{contradictory two-needle} (\textit{CTN}) variant, in which it conflicts with world knowledge. 
% Together, these settings enable evaluation of strictly context-faithful reasoning in two-hop scenarios.
Appendix~\ref{app:nolima_needle_setting_details} provides additional details and templates for \textit{SN}, \textit{TN}, and \textit{CTN} settings.

\paragraph{Haystack assembly.}

We adopt the random book generation procedure from NoLiMa~\citep{modarressi2025nolima} to construct haystacks ranging from 10K to 1M characters (10K--50K in 10K increments and 100K--1M in 100K increments), generating three random books ${B^c_1, B^c_2, B^c_3}$ for each context length $c$. Because tokenization is model-dependent, we vary input length in characters rather than tokens. Each haystack is split into line-indexed entries using the `\textbackslash n' delimiter. In the single-needle setting, the needle is inserted at one of four normalized depths: 0, 0.33, 0.67, or 1.0. In the two-needle setting, the primary needle is fixed at depth 0.5, while the secondary needle is inserted at one of the same four depths.

\subsection{Multi-document QA Datasets}\label{sec:multi-document_qa_dataset}

To stress the retrieval pipeline on the dimensions it is meant to handle, we reformulate two existing corpora into evidence-grounded multi-document question answering (QA) benchmarks. Both are designed to make RAG's retrieval-vs.-reading split observable: gold evidence is labeled at the line (sentence or paragraph) level, so the same scoring code works for LC and any retriever-ordered context. 

\paragraph{\textsc{MultiHop-RAG}~\citep{tang2024multihop}.}
The official release contains \textbf{609 news articles} (4.6M characters, 1.2M tokens with the \texttt{cl100k\_base} tokenizer) and 2{,}556 queries in four templated types: \emph{inference}, \emph{comparison}, \emph{temporal}, and \texttt{null\_query} (unanswerable). Each non-null query carries an \texttt{evidence\_list} of \textbf{2--4 gold facts}, each consisting of a source article title and an exact textual span. Removing the 301 \texttt{null\_query} items leaves \textbf{2{,}255 usable queries}; for the pilot reported in this paper, we uniformly sample 50 queries (seed 0), naturally stratified across the three answerable types ($n_{\text{inf}}{=}13$, $n_{\text{comp}}{=}21$, $n_{\text{temp}}{=}16$).

For the LC condition, we construct an \emph{oracle-containing} per-query mini-haystack comprising every gold-evidence article plus 20 distractor articles sampled from the remainder of the corpus. The articles are concatenated in random order and split into sentence-per-line entries, producing contexts of approximately 241K characters on average. For the RAG condition, we perform \emph{open-corpus retrieval} over the full 609-article corpus: we divide the corpus into 256-token sliding windows (6{,}296 chunks), embed the chunks with \texttt{text-embedding-3-large}, and rank them by cosine similarity to produce top-$K$ contexts for $K\in\{4,10,25,50,100,200\}$.

The two conditions are therefore not pool-matched: LC evaluates reading and evidence utilization when all gold source articles are guaranteed present, whereas RAG must first retrieve the required evidence from the full corpus. Accordingly, this experiment diagnoses the contribution of open-corpus retrieval failures to end-to-end performance; it is not a direct LC--RAG comparison under identical candidate pools. In both conditions, gold facts are re-mapped to line numbers inside the assembled context; in RAG, a gold fact that is not retrieved has no corresponding gold line. Appendix~\ref{sec:app_multihop_rag} reports the corresponding retrieval ceilings (Hits@$k$ for \emph{any} vs.\ \emph{all} gold facts) and reproduces the answer-only RAG baselines from the original paper.

\paragraph{\textsc{QASPER-MP}.}
\textsc{QASPER}~\citep{dasigi-etal-2021-qasper} contains 5{,}049 information-seeking questions over 1{,}585 NLP papers, each annotated by domain experts with paragraph-level evidence and a short free-form or extractive answer (including \texttt{Unanswerable}). The original task is single-document and, therefore, does not exercise the retrieval pipeline.

We construct \textbf{QASPER-MP} (multi-paper) by pairing each gold paper with \textbf{ten distractor papers} sampled from the QASPER pool, yielding an 11-paper haystack per query, and rewrite every question to include a realistic paper cue~-~in $\sim$50\% of cases a first-author surname (e.g.,~``Smith et al.''), and otherwise a 2--3 word verbatim title fragment~-~split by a hash of the question id. This forces the system to resolve document identity from a realistic cue rather than the full paper title. The task jointly tests document identification, evidence localization, and answer generation. Since distractors come from the same NLP-paper collection, they provide domain-matched context. A successful system must identify the relevant paper and supporting paragraph(s) before answering, while retaining QASPER's expert-annotated answers and evidence.

For LC, papers are serialized in an XML-like format that preserves titles, abstracts, and section structure (\texttt{<paper>}, \texttt{<title>}, \texttt{<abstract>}, \texttt{<section>}); paragraphs are line-numbered, and the same paragraph-level gold labels are used for evidence scoring. For RAG, we evaluate both a \emph{flat} variant (concatenate the top-$K$ retrieved paragraphs in rank order) and a \emph{structured} variant that re-emits the surrounding \texttt{<paper>}/\texttt{<title>}/\texttt{<abstract>} scaffolding and inserts ``\texttt{...}'' markers for discontinuities, so that retrieved fragments retain their document provenance.

We evaluate \textsc{QASPER-MP} at two scales. For proprietary models, we conduct a pilot on $N{=}50$ queries, comprising 38 answerable and 12 unanswerable examples. For open-weight models, we expand the evaluation to $N{=}200$ queries, providing broader benchmark coverage and a more stable estimate of aggregate performance. Because the proprietary-model results are based on the smaller pilot, comparisons across the two model groups should be interpreted with the corresponding difference in sampling uncertainty in mind. Construction details, sampling procedures, prompts, and the LLM-judge protocol are provided in Appendix~\ref{sec:app_qasper_mp}.

\section{Experiments}
\label{sec:experiments}

% We evaluate EBG under two prompt conditions: an \textit{answer-only} 
% prompt (baseline, similar to NoLiMa~\citep{modarressi2025nolima}) 
% and an \textit{answer~+~evidence} prompt that additionally requires 
% the model to cite the supporting line numbers. 
% Comparing the two isolates the effect of grounding on accuracy and 
% reveals evidence-retrieval failures invisible to answer-only evaluation. 
We organize our experiments around four research questions:

\begin{itemize}
[leftmargin=1em, itemsep=0pt, topsep=0pt, parsep=0pt, partopsep=0pt]
\item\textbf{RQ1. }\textit{What are the practical limitations of RAG, and can long-context EBG address them?}

\item\textbf{RQ2. }\textit{Does requiring models to explicitly cite evidence improve answer accuracy?}

% \item\textbf{RQ2.1 }\textit{Are the models answering from parametric knowledge or making random guesses?}

% \item\textbf{RQ2.1. } Do LLMs respect the evidence?

\item\textbf{RQ3. }\textit{Can models correctly cite evidence lines even in a two-needle setting?}

\item\textbf{RQ4. }\textit{Can models remain context-faithful even when evidence contradicts their built-in knowledge?}

% \item\textbf{RQ5. }\textit{Do LLMs perform better with randomly ordered haystacks or with retriever-ordered chunks?}

\end{itemize}

\subsection{Experimental Setup}
\label{sec:experimental_setup}
We evaluate 10 state-of-the-art proprietary models from three major providers across different experiments: \texttt{GPT-4o}, \texttt{GPT-4.1}, \texttt{o3-mini}, \texttt{GPT-5}, 
\texttt{GPT-5.5},
\texttt{Sonnet-4}, \texttt{Sonnet-4.5},
\texttt{Opus-4.6},
\texttt{Gemini-2.5-Fl}, and \texttt{Gemini-3-Fl}.  Additionally we evaluate three open-weight models served through Fireworks.ai\footnote{\url{https://fireworks.ai/}}~~-~~\texttt{GPT-OSS-120B}, \texttt{GLM-5.2}, and \texttt{Nemotron-3.5-Lightning-30B-A3B} (or \texttt{Nemo-3.5-L} in short)~-~to test whether our findings hold beyond proprietary systems. We evaluate evidence-based generation (\textit{EBG}) under both long-context (\textit{LC}) and \textit{RAG} settings. Prompt templates for the \emph{answer-only} and \emph{EBG} conditions are provided in Appendix~\ref{app:prompt_templates}. 
% Under \textit{answer-only} prompting, we report answer accuracy ($\texttt{Ans}^*$); under \textit{EBG}, we report answer (\texttt{Ans}), evidence (\texttt{Evid}), and joint answer-plus-evidence (\texttt{Ans
% +Evid}) accuracies.

\subsubsection{Multi-document QA setup.}
\label{sec:multidoc_setup}
For \textbf{RQ1}, we use two multi-document QA datasets~-~\textsc{MultiHop-RAG} and \textsc{QASPER-MP}.
% in which each query is
% evaluated under matched LC EBG and RAG pipelines.
% \textsc{MultiHop-RAG}~\citep{tang2024multihop} contains 2{,}255 news
% queries each requiring 2--4 evidence facts across distinct articles.
% \textsc{QASPER-MP} is a multi-paper adaptation of
% QASPER~\citep{dasigi-etal-2021-qasper} in which each gold paper is placed
% alongside ten distractors and the question is rewritten with a short,
% realistic paper cue (a first-author surname or 2--3 word title fragment).
Construction details, retrieval variants, and the LLM-judge protocol are
in Appendices~\ref{sec:app_multihop_rag} and~\ref{sec:app_qasper_mp}.

\subsubsection{EBG Setup}
\paragraph{Synthetic Dataset.}
We sweep context lengths from 10K to 1M characters. For each length, we run $N{=}100$ trials: each trial draws a fresh distractor haystack and samples one needle uniformly from the 10 curated two-hop needles. Here we apply the \textit{answer-only} and \textit{EBG} prompts and report \texttt{Ans*}, \texttt{Ans}, \texttt{Evid}, and \texttt{Ans+Evid} accuracies. A sanity check at $L{=}10\text{K}$ confirms all models achieve ${\geq}0.98$ joint accuracy, establishing that degradation at longer lengths reflects context handling rather than needle ambiguity. This setup addresses \textbf{RQ2} and \textbf{RQ3}; there are no contradictory needles, so RQ4 is not evaluated here.

\paragraph{NoLiMa-based Dataset.}
Under the NoLiMa setting, we use $B^c_1$ as the haystack source across all context lengths to maintain a controlled evaluation setup. We use the \texttt{T01} test variant throughout, as it provides exactly one input argument for the one-hop question and one for the two-hop question associated with each needle. This enables a consistent comparison between direct retrieval and commonsense composition while holding the underlying needle fixed. For example, for \texttt{0405}, the one-hop question is instantiated with \textit{fish-based meals}, whereas the two-hop question uses \textit{Brandade}; similarly, \texttt{0402} uses \textit{milk} and \textit{a cappuccino}, respectively. We test two characters, \textit{Yuki} and \textit{Stuart}, at the insertion depths defined in \S~\ref{sec:benchmark_nolima_data}, yielding 16 needle-placement conditions per context length per model (4 needles $\times$ 4 depths).

We evaluate three settings: \textit{single-needle} (one-hop and two-hop, addressing \textbf{RQ2}), \textit{two-needle} (two-hop with explicit secondary evidence, addressing \textbf{RQ3}), and \textit{contradictory two-needle} (CTN, addressing \textbf{RQ4}). Results are reported separately for one-hop and two-hop questions and averaged across the corresponding conditions.

\subsubsection{NoLiMa-based RAG Setup}\label{sec:nolima_rag_setup}

We compare long-context \textit{LC} and \textit{RAG} on the NoLiMa-based dataset under \textbf{RQ1} (Appendix \ref{app:nolima_rag}). We first analyze RAG retrieval performance for the single-needle and two-needle settings using one-hop and two-hop questions, examining how the retrieval depth required for full evidence coverage changes with haystack length and reasoning complexity. After needle insertion, each haystack line is embedded using OpenAI's \texttt{text-embedding-3-small} model and ranked by cosine similarity to the query.

We then compare \textit{RAG} and \textit{LC} under the same \textit{EBG} prompting, reporting \texttt{Ans} accuracy. Since RAG contexts can be smaller than their source haystacks, we use matched character budgets. For each haystack length, we identify the minimum top-$k$ providing full evidence coverage, compute its character count, and match it to the closest available LC context length.
Table~\ref{tab:context_matching} summarizes this mapping. For example, full-coverage retrieval from the 1M-character haystack in the two-needle setting requires 920 vectors (217,152 characters) and is matched to the 200K LC condition. This approximately controls for the amount of text presented to the model and isolates retrieval-ranked versus long-context inputs under comparable budgets.

\subsection{Results}
\label{sec:results}

We report results organized by the research questions introduced in
\S~\ref{sec:experiments}.

% \paragraph{RQ1: LC-EBG addresses two RAG limitations.}
% On \textsc{MultiHop-RAG}, queries require 2--4 evidence facts spread
% across distinct articles, so the retriever must surface every fact in
% its top-$k$ budget and missing one collapses the answer. We first
% reproduce the answer-only RAG baselines from the original benchmark
% under our pipeline (\texttt{text-embedding-3-large} + 256-token windows;
% Appendix~\ref{sec:app_multihop_rag}), and measure the retrieval ceiling
% directly: at $k{=}10$, the retriever covers \emph{any} gold fact for
% $82\%$ of queries but covers \emph{all} gold facts for only $19\%$ (and
% just $8\%$ on inference queries). \textbf{This all-vs-any gap is the structural
% limitation of similarity retrieval on multi-fact queries}, and it
% upper-bounds any RAG reader regardless of model strength. The reader
% numbers confirm it: at $k{=}10$ \textit{RAG} answer accuracy is
% $60/48/44\%$ (\texttt{GPT-5}/\texttt{Sonnet-4.5}/\texttt{Gemini-3-Fl}), while \textit{LC-EBG}
% on the same queries reaches $86/82/78\%$ ($+26$ to $+34$\,pp); even at
% $k{=}200$, where the retrieved context already approaches the size of
% the \textit{LC} haystack, \textit{RAG} only ties \textit{LC-EBG} for one of the three
% models. \textbf{The bottleneck is the retrieval ceiling, not reading}.

\paragraph{RQ1: LC-EBG addresses two RAG limitations.}
On \textsc{MultiHop-RAG}, queries require 2--4 evidence facts spread
across distinct articles, so faithful answering requires the retriever to
surface all required facts within its top-$k$ budget. For this benchmark,
the two conditions use different evidence-access regimes:
\emph{open-corpus RAG} searches 6{,}296 chunks constructed from all
\textbf{609 articles}, whereas the \emph{oracle-containing LC} condition
receives every gold-evidence article plus 20 distractor articles. The
conditions are therefore not pool-matched; this comparison diagnoses the
cost of open-corpus retrieval relative to reading when the gold source
articles are guaranteed to be available.

We first reproduce the answer-only RAG baselines from the original
benchmark under our pipeline
(\texttt{text-embedding-3-large} + 256-token windows;
Appendix~\ref{sec:app_multihop_rag}) and measure annotated-gold retrieval
coverage directly. At $k{=}10$, the retriever covers \emph{any} gold fact
for $82\%$ of queries but covers \emph{all} gold facts for only $19\%$
(and just $8\%$ on inference queries).
\textbf{This all-vs-any gap exposes a limitation of the evaluated
similarity-retrieval setup: it often retrieves some relevant evidence but
fails to recover the complete evidence set required for multi-fact queries
within a practical top-$k$ budget.}
When the gold annotations are exhaustive and every annotated fact is
necessary, all-gold coverage bounds the faithfully grounded performance
of the downstream RAG reader. The reader results corroborate this
limitation (Appendix~\ref{sec:app_multihop_c},
Table~\ref{tab:lc_vs_rag}).

Under \textit{EBG} prompting at $k{=}10$, open-corpus \textit{RAG}
achieves \texttt{Ans} accuracies of $60/48/44\%$ for \texttt{GPT-5},
\texttt{Sonnet-4.5}, and \texttt{Gemini-3-Fl}, respectively, whereas the
oracle-containing \textit{LC-EBG} condition achieves $86/82/78\%$ on the
same queries. The resulting $26$--$34$\,pp gap includes the cost of
retrieving the required evidence from the full corpus and should not be
interpreted as a pool-matched estimate of an intrinsic LC--RAG
difference. The gap remains at $k{=}200$, where the open-corpus RAG context approaches the size of the LC haystack despite comparable context sizes, but its \texttt{Ans} accuracy remains lower for all three models. Under \textit{answer-only}
prompting, open-corpus RAG at $k{=}200$ reaches parity with the
oracle-containing LC condition only for \texttt{Sonnet-4.5}
($68\%$ vs.\ $68\%$).

Consistent with this, our
\textsc{NoLiMa}-based RAG analysis
(Table~\ref{tab:topk_budget_summary}) shows that \textbf{retrieval costs scale
sharply with reasoning complexity and haystack length}.  While one-hop full
coverage remains relatively efficient, two-hop retrieval grows from 1.9K
retrieved characters at 10K context to over 202K at 1M context,
effectively pushing RAG toward brute-force long-context processing
without guaranteeing complete evidence recovery.

% \textbf{The bottleneck is the retrieval ceiling, not reading}.

% On \textsc{QASPER-MP}, the
% query refers to a paper by an author or short title cue; answering
% requires document-level identity, not just topical fragments.
% Table~\ref{tab:qasper_mp_results_main} reports answerable-only judge
% scores under the apples-to-apples comparison (\textit{LC-EBG} and \textit{RAG} share the
% same 11-paper pool). \textit{LC-EBG} beats the best \textit{RAG} variant for every one of
% five frontier models, with gaps of $+0.05$ to $+0.13$ judge points.
% Aggregate scores look tied only because \textit{RAG} over-refuses on missed
% retrievals ($24$--$40\%$ refusal on answerable queries vs.\
% $13$--$18\%$ for \textit{LC-EBG}), which accidentally matches QASPER's
% \texttt{Unanswerable} gold label on the 24\% of pilot questions where
% Unanswerable is correct. With full-corpus retrieval
% (\texttt{flat\_full@10}), all-needles coverage drops to 48\% and \textit{LC-EBG}
% beats the best \textit{RAG} configuration by 5--13 judge points
% (Appendix~\ref{sec:app_qasper_mp}).
\begin{table}[t!]
\centering
\small
\setlength{\tabcolsep}{4pt}
\begin{tabular}{lcccc}
\toprule
\textbf{Model}        & \textbf{LC}             & \textbf{flat@3} & \textbf{struct@3} & \textbf{$\Delta$LC (best)} \\
\midrule
GPT-5        & \textbf{0.605} & 0.500 & 0.553 & $+0.05$ \\
GPT-5.5      & \textbf{0.671} & 0.500 & 0.539 & $+0.13$ \\
Opus-4.6     & \textbf{0.697} & 0.474 & 0.566 & $+0.13$ \\
Sonnet-4.5   & \textbf{0.592} & 0.513 & 0.474 & $+0.08$ \\
Gemini-3-Fl  & \textbf{0.658} & 0.487 & 0.539 & $+0.12$ \\
\bottomrule
\end{tabular}
\caption{\textbf{\textsc{QASPER-MP} answerable-subset LLM-judge scores
(\(n{=}38\), EBG)}. \textbf{LC} wins for all five models; the largest
gaps are for the strongest readers (\texttt{Opus-4.6} and \texttt{GPT-5.5}). Higher ($\uparrow$) values are better. Full
breakdown in Appendix~\ref{sec:app_qasper_mp}.}
\label{tab:qasper_mp_results_main}
\end{table}

On \textsc{QASPER-MP}, the query refers to a paper by an author or short
title cue; answering requires document-level identity, not just topical
fragments.
\textbf{This exposes a second limitation of RAG: chunk-level
similarity retrieval struggles with document identity and provenance
when queries refer indirectly to a source.}
Table~\ref{tab:qasper_mp_results_main} reports answerable-subset judge
scores for the pool-matched comparison, where \textit{LC-EBG} and
\textit{RAG} use the same 11-paper pool.
\textit{LC-EBG} exceeds the best pool-matched RAG variant for all five
frontier models by $+0.05$ to $+0.13$ judge points.
Aggregate scores look tied only because \textit{RAG} over-refuses on
missed retrievals ($24$--$40\%$ on answerable queries vs.
$13$--$18\%$ for \textit{LC-EBG}), accidentally matching QASPER's
\texttt{Unanswerable} gold label on the 24\% of pilot questions for
which it is correct. Under the open-corpus condition
(\texttt{flat\_full@10}), all-needles coverage drops to 48\%, and the
judge-score gap relative to LC is 5--13 points
(Appendix~\ref{sec:app_qasper_mp}).

Across both datasets, retrieval is a major error source despite different
evidence-access regimes: \textsc{MultiHop-RAG} compares open-corpus RAG
with oracle-containing LC, whereas \textsc{QASPER-MP} uses a pool-matched
11-paper comparison. This pattern persists for three open-weight readers
on $n{=}200$ queries, with LC exceeding RAG by $+13$ to $+36$\,pp in
EBG \texttt{Ans} accuracy on \textsc{MultiHop-RAG} and by $+0.07$ to
$+0.24$ answerable-subset judge points on \textsc{QASPER-MP}, alongside
a large residual grounding gap
(Appendices~\ref{sec:app_multihop_openweight}
and~\ref{sec:app_qasper_openweight}).

\begin{figure}[t]
    \centering
    \includegraphics[width=\linewidth]{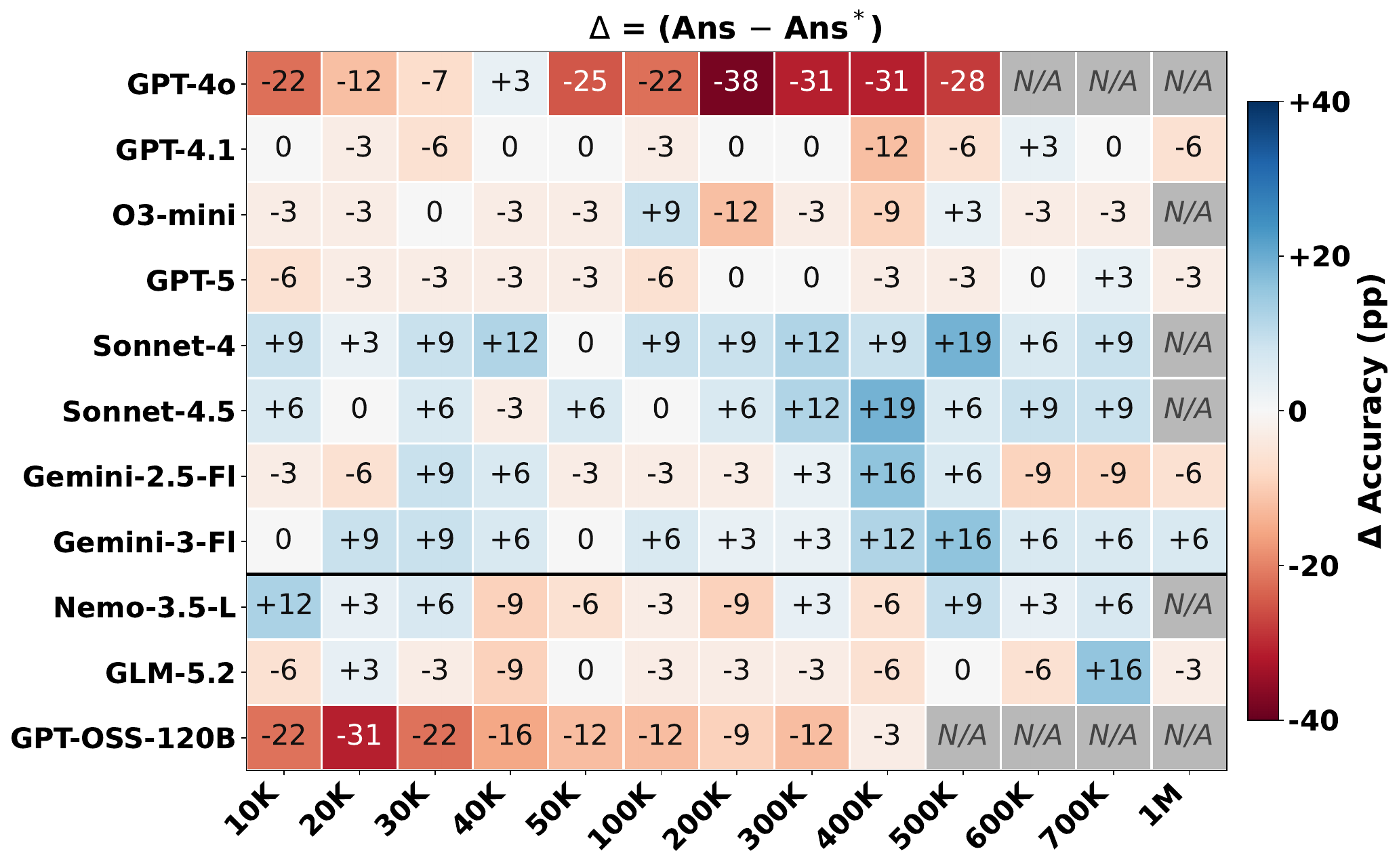}
    \caption{NoLiMa-based \textit{LC} benchmark comparing \textbf{EBG (\texttt{Ans})} and \textbf{\textit{answer-only} (Ans$^{*}$)} prompting across eight proprietary and three open-weight models over 10K--1M context lengths. Values report the \textbf{accuracy difference ($\Delta$)}, with larger positive values ($\uparrow$) indicating greater EBG gains; \textit{N/A} denotes inputs exceeding the model's context window.
    }
    \label{fig:answer_only_comparison}
\end{figure}

\paragraph{RQ2: Citation improves answer accuracy.}

On the synthetic benchmark (Fig.~\ref{fig:synthetic_ans_vs_ans_only}, Appendix~\ref{sec:app_lc_ebg_synthetic}), all three frontier models~-~GPT-5, Claude Sonnet-4.5, and Gemini-3-Flash~-~approach ceiling answer accuracy under both prompting conditions (\textit{answer-only} and \textit{EBG}). GPT-5 and Gemini-3-Flash maintain $\geq 98\%$ accuracy across all tested context lengths, while Sonnet-4.5 remains within the $93$--$100\%$ range up to 700K characters. 
Consequently, the synthetic dataset shows no measurable citation-as-scaffold effect on answer accuracy: \textit{Ans$^{*}$} and \textit{Ans} remain within sampling noise across all cells, \textbf{including at the longest contexts}. This does not refute RQ2 but reflects the controlled benchmark design: lexically dissimilar distractors make the two-hop facts relatively easy to isolate, leaving little headroom for EBG's benefits.

A different pattern emerges on the more challenging NoLiMa-based \textit{LC} benchmark (Fig.~\ref{fig:answer_only_comparison}), where the scaffolding effect becomes apparent \textbf{under long-context stress}. Sonnet and Gemini models gain between $+9$ and $+19$\,pp from \textit{EBG} prompting across context lengths from 300K to 900K. Among open-weight models, the smaller \texttt{Nemo-3.5-L} improves by $+3$ to $+12$\,pp, while \texttt{GLM-5.2} reaches a maximum gain of $+16$\,pp at 700K; in contrast, \texttt{GPT-OSS-120B} shows no benefit. Taken together, the contrast between the synthetic and NoLiMa settings suggests that \textbf{scaffolding is most useful when long-context retrieval is sufficiently challenging, while its effectiveness remains model-dependent and may not simply scale with model capacity}.

\paragraph{RQ3: Asymmetric Retrieval Behavior.}
% As observed earlier, frontier LLMs attain near-ceiling answer accuracy on the synthetic benchmark under two of our primary prompt settings, limiting its ability to expose retrieval asymmetries. The underlying trend, however, becomes apparent in the NoLiMa-based \textit{LC} benchmark. 
\begin{figure}[t]
    \centering
    \includegraphics[width=\linewidth]{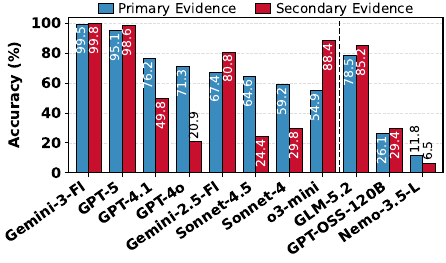}
    \caption{Two-needle citation breakdown: rate of correctly citing the \textbf{primary  needle vs.\ } the \textbf{secondary  needle} under NoLiMa-based \textit{LC} benchmark. Higher ($\uparrow$) values are better.}
    \label{fig:tn_primary_secondary_citation}
\end{figure}

As shown in Fig.~\ref{fig:tn_primary_secondary_citation}, recovery of the two evidence types varies substantially across models. \texttt{Gemini-3-Fl} and \texttt{GPT-5} achieve near-perfect and balanced citation of both the primary and secondary needles (99.5/99.8\% and 95.1/98.6\%, respectively). \texttt{Gemini-2.5-Fl} also performs relatively well, but exhibits an inversion, with secondary-needle citation exceeding primary-needle citation (80.8\% vs.\ 67.4\%). A similar, more pronounced inversion appears for \texttt{o3-mini} (88.4\% vs.\ 54.9\%). These cases suggest that successful recovery of one evidence type does not necessarily imply comparable recovery of the other.

This asymmetry reverses for several other models, where primary-needle recovery substantially exceeds secondary-needle recovery. \texttt{GPT-4.1}, \texttt{GPT-4o}, and the \texttt{Sonnet} models exhibit this pattern, with the largest disparities observed for \texttt{GPT-4o} (71.3\% vs.\ 20.9\%) and \texttt{Sonnet-4.5} (64.6\% vs.\ 21.4\%). This disparity indicates that retrieving the primary evidence does not guarantee successful attribution of the complementary evidence required for the task. Moreover, primary-needle citation itself remains imperfect, showing that retrieval failures extend beyond the secondary signal alone.

The open-weight models show substantial variation in retrieval. \texttt{GLM-5.2} achieves balanced recovery (78.5\% primary and 85.2\% secondary), whereas \texttt{GPT-OSS-120B} records lower rates for both needles (26.1\% and 29.4\%). \texttt{Nemo-3.5-L} performs poorly on both (11.8\% and 6.5\%), indicating broader retrieval limitation rather than directional asymmetry. \textbf{Overall, multi-evidence retrieval varies across models and evidence types, reinforcing model-dependent patterns as in RQ1.}

% \begin{figure*}[t]
%     \centering
%     \begin{subfigure}[t]{0.49\linewidth}
%         \centering
%         \includegraphics[width=\linewidth]{images/contradictory_two_needle_comparison.pdf}
%         \caption{TN vs.\ CTN overall (answer vs.\ \texttt{both\_correct}).}
%         \label{fig:ctn_comparison}
%     \end{subfigure}
%     \hfill
%     \begin{subfigure}[t]{0.49\linewidth}
%         \centering
%         \includegraphics[width=\linewidth]{images/contradictory_two_needle_both_accuracy.pdf}
%         \caption{TN vs.\ CTN \texttt{both\_correct} across settings.}
%         \label{fig:ctn_both_breakdown}
%     \end{subfigure}
%     \caption{CTN two-hop evaluation. TN uses a factual secondary needle; CTN uses a contradictory one conflicting with world knowledge, requiring context-faithful reasoning and citation.}
%     \label{fig:ctn_two_needle}
% \end{figure*}

\begin{figure}[t]
\centering
\includegraphics[width=\linewidth]{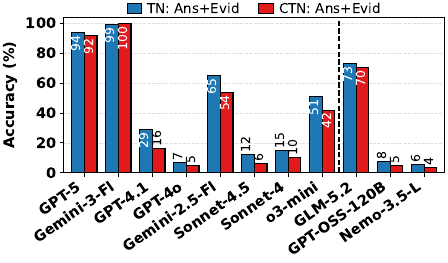}
\caption{\textbf{TN vs.\ CTN}, \texttt{Ans+Evid}. TN uses a factual secondary needle; CTN uses a contradictory secondary needle, conflicting with world knowledge, requiring context-faithful reasoning and citation. Models to the right of the dashed line are open-weight. Higher ($\uparrow$) values are better. \texttt{Ans} accuracy for the same runs is reported in Figure~\ref{fig:ctn_two_needle_ans}.}
\label{fig:ctn_two_needle}
\end{figure}

\paragraph{RQ4: CTN amplifies failure.}
Performance generally declines under contradiction, with substantial variation across models (Fig.~\ref{fig:ctn_two_needle}); answer accuracy is reported separately in Figure~\ref{fig:ctn_two_needle_ans}. \texttt{GPT-5} and \texttt{Gemini-3-Fl} remain relatively context-faithful, preserving high answer accuracy, though their \texttt{Ans+Evid} still drops, indicating greater sensitivity of evidence grounding to contradiction. In contrast, \texttt{GPT-4.1}, \texttt{GPT-4o}, and the \texttt{Sonnet} variants show sharper declines, particularly in \texttt{Ans+Evid}, suggesting they can recover the expected answer while failing to reconcile contradictory evidence.

CTN robustness is also uneven among open-weight models. \texttt{GLM-5.2} is notably resilient: \texttt{Ans} decreases only from 79\% to 72\%, while \texttt{Ans+Evid} changes from 73\% to 70\%, largely preserving the small gap between the two metrics. By contrast, \texttt{GPT-OSS-120B} and \texttt{Nemo-3.5-L} start from substantially weaker TN baselines: the former falls from 28/8\% (\texttt{Ans}/\texttt{Ans+Evid}) to 8/5\% under CTN, and the latter from 12/6\% to 5/4\%. These models therefore already struggle to extract and ground answers under TN, with CTN further compounding this weakness.

Overall, \texttt{Ans+Evid} is generally more vulnerable to CTN than \texttt{Ans}, showing that \textbf{resolving contradiction while grounding the answer in supplied evidence is harder than producing the correct answer itself.} The widening \texttt{Ans}--\texttt{Ans+Evid} gap highlights a common failure mode: models reach the expected answer without faithfully using the context that should justify it, even when answers remain correct. See Appendix~\ref{app:tn_vs_ctn} for more details.

\section{Conclusion and Future Work}
We introduced \textbf{evidence-based generation (EBG)}, a framework that requires models to produce both an answer and line-level citations, together with a joint \texttt{Ans+Evid} metric that measures correctness and grounding simultaneously. By separating retrieval from evidence-grounded reasoning, EBG exposes failures obscured by answer-only evaluation. We evaluate these failures using evidence-grounded versions of \textsc{MultiHop-RAG} and \textsc{QASPER-MP}, a controlled synthetic two-hop haystack, and an evidence-augmented NoLiMa spanning 10K--1M characters. \textbf{This evaluation suite spans both realistic and controlled settings, allowing retrieval, reasoning, and grounding errors to be examined under complementary conditions.}

Across ten proprietary frontier and three open-weight LLMs, EBG reveals two key RAG limitations: \emph{(i)}~similarity retrieval finds \emph{any} relevant chunk more reliably than \emph{all} evidence required for multi-fact queries, imposing a retrieval ceiling at realistic top-$k$ budgets; and \emph{(ii)}~chunking can discard document-level meta-information needed by identity-bearing queries, causing RAG to over-refuse on retrieval misses. EBG further exposes grounding failures hidden by answer-only metrics: models often cite incomplete evidence, and the gap between answer and joint accuracy widens when in-context evidence conflicts with parametric knowledge. These findings show that correct answers alone can substantially overstate a model's ability to faithfully use long-context evidence. They also highlight evidence attribution as a useful diagnostic for distinguishing retrieval limitations from downstream reasoning failures. \textbf{Consequently, EBG provides a more informative basis for comparing RAG and long-context systems than evaluations that collapse retrieval, reasoning, and attribution into answer accuracy alone.}

These results do not imply that ``LC replaces RAG''; rather, LC and RAG exhibit distinct failure modes that EBG helps localize, motivating evaluations that measure both answer accuracy and evidence fidelity. \textbf{The appropriate architecture may therefore depend not only on answer quality, but also on which sources of failure are acceptable for a given task and context regime.} Future work includes scaling EBG to the full benchmarks and testing document-aware chunking. Richer evidence attribution and the observed reasoning benefits of evidence generation also motivate investigating grounding objectives as part of model training.

\section*{Limitations}

Our evaluation focuses on synthetic and semi-synthetic long-context benchmarks designed to isolate grounding and evidence attribution behavior under controlled conditions. While these settings enable precise analysis, they do not capture the full complexity of real-world domains such as legal, scientific, or medical reasoning. In addition, our evidence metric operates at the line level and therefore measures coarse-grained provenance rather than fine-grained span-level attribution. The benchmarks also assume that the reference evidence is sufficiently complete and unambiguous, whereas real-world questions may admit multiple valid evidence sets or interpretations.

Our RAG analysis estimates end-to-end performance through a decomposition of retrieval and long-context accuracy rather than exhaustive evaluation of full production RAG pipelines across all configurations. We therefore do not systematically explore retrieval variants such as query rewriting, hybrid search, or adaptive context selection, which could alter the observed retrieval ceilings. Finally, because we evaluate proprietary frontier and open-weight models through API access, results may vary as deployed systems evolve over time.

\section*{Ethics and Broader Impact}

This work studies evidence grounding and provenance in long-context language models. Our benchmarks are constructed from synthetic or publicly available text and do not contain sensitive personal information. We do not train new foundation models; all experiments are conducted through inference-only evaluation of existing systems.

We believe evidence-based evaluation can improve transparency and verifiability in LLM-based systems by encouraging models to explicitly identify the context used to support their answers. Such capabilities may be particularly important in high-stakes applications such as legal analysis, scientific search, and medical document understanding, where unsupported or hallucinated answers can have significant consequences.

At the same time, citation generation should not be interpreted as a guarantee of faithful reasoning. A model may cite irrelevant or incomplete evidence while producing a plausible answer. We therefore view evidence-based generation as a complementary framework rather than a complete solution to factuality or interpretability in language models.

% \newpage

\bibliography{custom}

\appendix

\section{Prompt Templates}
\label{app:prompt_templates}

The haystack is line-numbered (\texttt{[1]}, \texttt{[2]}, \ldots, \texttt{[N]}) and substituted into the \texttt{{haystack}} placeholder of each task template. We use four prompt variants: an \emph{answer-only} variant (no system prompt), a \emph{Chain-of-Thought (CoT)} variant (no system prompt), an \emph{evidence-based generation (EBG)} variant that asks the model to cite supporting line numbers in JSON, and a \emph{CoT-based EBG} variant (no system prompt) that requests both brief reasoning and the EBG JSON answer-and-evidence output.

\begin{promptbox}[Answer-only Prompt (User Message)]
\footnotesize\ttfamily
You will answer a question based on the following book snippet:\\[2pt]
\{haystack\}\\[4pt]
Use the information provided in the book snippet to answer the question. Your answer should be short and based on either explicitly stated facts or strong, logical inferences.\\[4pt]
Question: \{question\}\\[4pt]
Return only the final answer with no additional explanation or reasoning.
\end{promptbox}

\begin{promptbox}[CoT Prompt (User Message)]
\footnotesize\ttfamily
You will answer a question based on the following book snippet:\\[2pt]\{haystack\}\\[4pt]Use the information provided in the book snippet to answer the question. Be aware that some details may not be stated directly, and you may need to INFER the answer based on the given information. Begin with a brief explanation of your reasoning in NO MORE THAN THREE (3) sentences. Then, return the final answer on a new line.\\[4pt]Question: \{question\}
\end{promptbox}

\begin{promptbox}[EBG System Prompt]
\footnotesize\ttfamily
Your job is to answer the question entirely from the context and provide a reference. Your answer should cite all lines the answer is based on.
\end{promptbox}

\begin{promptbox}[EBG User Prompt]
\footnotesize\ttfamily
<context>\{haystack\}</context>\\[4pt]
Answer the question based on information only from the context. If the question is not answerable from the context, answer NA. Your response should comprise only the answer and all lines the answer is based on in json format. For example:\\[2pt]
\textasciigrave\textasciigrave\textasciigrave json\\
\{\,``lines'': [25, 412], ``answer'': ``John''\,\}\\[2pt]
\textasciigrave\textasciigrave\textasciigrave\\[4pt]
Question: \{question\}
\end{promptbox}

\begin{promptbox}[CoT-based EBG Prompt (User Message)]
\footnotesize\ttfamily
You will answer a question based on the following book snippet:\\[2pt]
<context>\{haystack\}</context>\\[4pt]
Use the information provided in the book snippet to answer the question. Be aware that some details may not be stated directly, and you may need to INFER the answer based on the given information. Begin with a brief explanation of your reasoning in NO MORE THAN THREE (3) sentences. Then, answer the question based on information only from the context. If the question is not answerable from the context, answer NA. Provide the answer (just the character name) and all lines the answer is based on in json format. For example:\\[2pt]
\textasciigrave\textasciigrave\textasciigrave json\\
\{\,``lines'': [25, 412], ``answer'': ``John''\,\}\\[2pt]
\textasciigrave\textasciigrave\textasciigrave\\[4pt]
Question: \{question\}
\end{promptbox}

\section{Additional Details of Synthetic Dataset}\label{app:synth_dataset_details}

We provide additional details on the construction and structure of our synthetic dataset and present a representative example to illustrate the resulting task format and evidence organization.

\subsection{Example}\label{app:synth_dataset_example}
Consider facts \textit{``Himalaya Inc.\ launched FitBand''} and \textit{``Himalaya Inc.\ is based in Mumbai''}, with question \textit{``Which smart wearable was launched by a company based in Mumbai?''} At $L{=}10\text{K}$ the two facts land at lines 56 and 67 among random distractors:

\begin{promptbox}[Haystack excerpt and question (synthetic dataset)]
\footnotesize\ttfamily
\ldots\\
54: Tech research group DataSynergy optimized machine learning algorithms.\\
55: A renowned sculptor in Rome crafted a public monument.\\
56: Himalaya Inc. launched a new smart wearable called FitBand.\\
57: A notable charity, WarmHearts, distributed winter supplies to families in need.\\
58: GreenWave unveiled a new system for harnessing tidal energy.\\
\ldots\\
65: Fitness startup VitalMotion developed smart workout equipment.\\
66: GlobalVoyager published a feature on little-known cultural festivals.\\
67: Himalaya Inc. is based in Mumbai.\\
68: Director James Park released a documentary on endangered rainforests.\\
69: Education reformist Tara Singh promoted progressive teaching methods.\\
\ldots\\[4pt]
Question: Which smart wearable was launched by a company based in Mumbai?
\end{promptbox}

Neither line alone is sufficient: line 56 identifies the product but not the company location, while line 67 identifies the location but not the product. The model must therefore compose evidence across both lines. Under the EBG prompt template, the expected output is a JSON object containing both the answer and the cited line numbers:

\begin{responsebox}[Expected EBG response]
\footnotesize\ttfamily
\{\,``lines'': [56, 67], ``answer'': ``FitBand''\,\}
\end{responsebox}

\begin{table*}[t!]
\centering
\small
\resizebox{\linewidth}{!}{
\begin{tabularx}{\linewidth}{lX}
\toprule
\textbf{Variant} & \textbf{Needle Template} \\
\midrule
\rowcolor{gray!15}\multicolumn{2}{l}{\textit{Original NoLiMa primary needles (under SN)}} \\
\texttt{0402}     & \textit{``A message came in from \{CHAR\} saying, `I'm lactose intolerant,' and nothing more.''} \\
\texttt{0402Inv}  & \textit{``A message came in saying, `I'm lactose intolerant,' from \{CHAR\}.''} \\
\texttt{0405}     & \textit{``Then \{CHAR\} mentioned that he has been vegan for years.''} \\

\texttt{0405Inv}  & \textit{``There was a vegan guest, named \{CHAR\}.''} \\
\midrule
\rowcolor{gray!15}\multicolumn{2}{l}{\textit{Gender neutral rephrased primary needle (under SN)}}\\
\texttt{0405-R}     & \textit{``Then \{CHAR\} mentioned being vegan for years.''} \\
\midrule
\rowcolor{gray!15}\multicolumn{2}{l}{\textit{Explicit secondary needles (under TN)}} \\
\texttt{TN-0402-S} & \textit{``\{CHAR\} said, `Cappuccino is a classic Italian coffee made with espresso and steamed milk.'\,''} \\
\texttt{TN-0405-S} & \textit{``Then \{CHAR\} mentioned that Brandade is a fish-based dish made with salted cod.''} \\
\midrule
\rowcolor{gray!15}\multicolumn{2}{l}{\textit{Explicit secondary needle (under CTN)}} \\
% \texttt{CTN\_0402-P} & \textit{``A message came in from \{CHAR\} saying, `I'm lactose intolerant,' and nothing more.''} \\
\texttt{CTN-0402-S} & \textit{``Black coffee is a rich and creamy beverage made by brewing coffee beans in steamed whole milk.''} \\
% \texttt{CTN\_0405-P} & \textit{``Then \{CHAR\} mentioned that he has been vegan for years.''} \\
\texttt{CTN-0405-S} & \textit{``Falafel is a popular Middle Eastern dish made from ground chickpeas mixed with minced lamb.''} \\
\bottomrule
\end{tabularx}
}
\caption{\textbf{Needle templates} used in the evaluation. Original NoLiMa needles under the \textbf{single-needle} (\textit{SN}) setting evaluate both one-hop and two-hop reasoning. We add \texttt{0405-R}, a gender-neutral rephrased version of the \texttt{0405} primary needle template under SN, to examine the influence of gendered wording in the needle template. Explicit secondary needles under the \textbf{two-needle} (\textit{TN}) setting directly provide the required world-knowledge fact and evaluate two-hop reasoning under fully contextualized evidence. The \textbf{contradictory two-needle} (\textit{CTN}) setting instead introduces secondary needles that conflict with world knowledge, enabling evaluation of whether models remain strictly faithful to the provided context during two-hop reasoning.}
\label{tab:needle_templates}
\end{table*}

Such a response is scored as correct on \texttt{Ans}, \texttt{Evid}, and \texttt{Ans+Evid}. The remaining nine needles follow the same structure with varied entities, relations, and answer forms.

\section{NoLiMa Datatset Adaptation Details}\label{app:nolima_dataset_adaptation_details}

\subsection{Needle settings}\label{app:nolima_needle_setting_details}

Table \ref{tab:needle_templates} presents the different needle templates used across different settings.
\subsection{Single Needle (SN) setting}
In this setting, \texttt{0402} and \texttt{0402Inv} encode the same \textit{dietary} fact about a character (\texttt{\{CHAR\}}) using different surface forms with altered word order. We refer to the first variant as the \emph{default} form and the second as the \emph{inverse} form. Similarly, needles \texttt{0405} and \texttt{0405Inv} state that a character (\texttt{\{CHAR\}}) is \textit{vegan}. These needles support both \textbf{one-hop} questions answerable directly from the needle (e.g., ``\textit{Which character is lactose intolerant?''}) and \textbf{two-hop} questions requiring an additional world-knowledge inference (e.g., ``\textit{Which character cannot drink a cappuccino?''}, which requires knowing that \textit{cappuccino} contains \textit{milk}). The
SN context supplies the entity-specific primary
needle but not the complete inference chain required by the
question. For example, a needle stating that a character is vegan
must be combined with parametric commonsense knowledge that
veganism excludes eating fish.

We therefore refer to the cited primary needle in SN as \emph{anchor evidence}. For SN, evidence accuracy measures whether the model cites the correct contextual anchor, and joint accuracy measures answer correctness together with anchor citation; these metrics should not be interpreted as establishing a complete context-only evidence chain. This distinction is specific to SN because part of the information required to answer the query may reside in the model's parametric knowledge. Accordingly, SN citations cannot directly attribute the contribution of such internally stored information.

\subsection{Two-needle (TN) setting}
As shown in Table~\ref{tab:needle_templates}, we introduce two explicit secondary needles under the two-needle (\textit{TN}) setting: \texttt{TN-0402-S} and \texttt{TN-0405-S}. These needles explicitly provide the world-knowledge fact required for two-hop reasoning (e.g., \textit{``{CHAR} said, `Cappuccino is a classic Italian coffee made with espresso and steamed milk.',''}). Each secondary needle pairs with the corresponding primary needles. For example, \texttt{TN-0402-S} pairs with \texttt{0402} and \texttt{0402Inv}, while \texttt{TN-0405-S} pairs with \texttt{0405} and \texttt{0405Inv}. Answering a two-hop question, therefore, requires retrieving and jointly reasoning over both the primary and secondary needles.

\subsection{Contradictory Two-needle (CTN) setting}
This setting introduces secondary needles that conflict with world knowledge. For example, \texttt{CTN-0402-S} claims that \textit{black coffee} contains \textit{steamed milk}, while \texttt{CTN-0405-S} claims that \textit{falafel} contains \textit{minced lamb}. As in \textit{TN}, each secondary needle pairs with its primary needles (e.g., \texttt{CTN-0402-S} with \texttt{0402} and \texttt{0402Inv}), testing whether models follow provided context or override it with parametric knowledge.
For both question types, the primary needle provides an entity-specific anchor while inference requires unstated bridge knowledge. One-hop questions require a commonsense bridge (e.g., veganism implies avoiding fish), while two-hop questions require an item-composition bridge (e.g., Brandade is fish-based).

\section{Long-context EBG vs.\ RAG on MultiHop-RAG}
\label{sec:app_multihop_rag}

% This appendix documents an additional set of experiments that compare
% long-context evidence-based generation (\textit{LC-EBG}) against a retrieval-augmented
% generation (\textit{RAG}) pipeline on the \textsc{MultiHop-RAG}
% benchmark~\citep{tang2024multihop}. The motivation is direct: our synthetic
% two-hop benchmark and the NoLiMa-adapted benchmark are constructed haystacks,
% which leaves open the question of whether the trends we observe carry over to
% realistic, web-derived multi-document QA where \textit{RAG} is the natural baseline.
% The filtered \textsc{MultiHop-RAG} subset (\S\ref{sec:multi-document_qa_dataset}) of 2{,}255 multi-hop queries is a clean testbed for
% both \textit{EBG} (requiring line-level evidence) and \textit{RAG} (requiring chunked, embeddable
% corpus).

% We run the following experiments on a 50-query pilot set (randomly sampled with seed 0 and naturally stratified across the three non-null question types in \textsc{MultiHop-RAG}: $n_{\text{inference}}{=}13$, $n_{\text{comparison}}{=}21$, and $n_{\text{temporal}}{=}16$). We use the same 50 queries throughout the appendix, making all reported results directly comparable.

This appendix analyzes an \emph{oracle-containing long-context
(LC)} reading condition and an \emph{open-corpus retrieval-augmented
generation (RAG)} condition on the \textsc{MultiHop-RAG} benchmark
\citep{tang2024multihop}. The LC condition guarantees that every
gold-evidence article is available to the reader together with 20
distractor articles, whereas the RAG condition must retrieve the
required evidence from the full 609-article corpus. The purpose of
this comparison is therefore to diagnose the contribution of
open-corpus retrieval failures to end-to-end performance; it is not
a pool-matched comparison of LC and RAG under identical candidate
document sets. The motivation is direct: our synthetic
two-hop benchmark and the NoLiMa-adapted benchmark are constructed haystacks,
which leaves open the question of whether the trends we observe carry over to
realistic, web-derived multi-document QA where \textit{RAG} is the natural baseline.

The filtered \textsc{MultiHop-RAG} subset (\S\ref{sec:multi-document_qa_dataset}) of 2{,}255 queries tests both \textit{EBG} (line-level evidence) and \textit{RAG} (chunked corpus). We run experiments on a 50-query pilot set, sampled with seed 0 and stratified across three types: $n_{\mathrm{inf}}{=}13$, $n_{\mathrm{comp}}{=}21$, and $n_{\mathrm{temp}}{=}16$. We use the same queries, prompts, and scoring throughout, while LC and RAG differ in candidate-document pools as described below.

\paragraph{LC haystacks.} For each query, we assemble a mini-haystack of (i)~all gold-evidence documents and (ii)~20 distractor documents sampled without replacement from the remaining corpus. Documents are randomly concatenated, split into sentence-level lines, and rendered in the \texttt{{lineno}: {text}} format used by our synthetic benchmark. Gold evidence facts are re-mapped to line numbers so EBG labels match the prompt. Mini-haystacks average \(\sim241\)K characters (\(\sim58\)K tokens), within every model's context window. This is the \emph{LC (full)} condition.

\paragraph{RAG haystacks.}
For each query, we construct RAG contexts at $K\in{4,10,25,50,100,200}$. The retriever searches the full 609-document corpus, divided into 256-token sliding windows (6{,}296 chunks). We embed chunks and queries with \texttt{text-embedding-3-large}, rank by cosine similarity, and use the top-$K$ chunks as the RAG context. Each chunk is prefixed with \texttt{[CHUNK <rank> | <title>]}, and rendered with the same sentence-per-line numbering as LC. Gold evidence facts are re-mapped to lines in the retrieved context; unretrieved facts have no corresponding gold line.

The conditions are not pool-matched: LC has all gold source articles plus 20 distractors, whereas RAG must locate the required evidence across all 609 articles. This design deliberately preserves the native evidence-access setting of each approach in practice, enabling a realistic comparison between them under their respective retrieval assumptions.

\subsection{Setup}
\label{sec:app_multihop_setup}

\paragraph{Prompt modes.} Each (query, context) pair is evaluated under two prompts differing in one sentence:
\begin{itemize}[leftmargin=*]
\item \emph{Answer-only ($\texttt{Ans}^{*}$).} The model returns a JSON object with only an \texttt{answer} field. We report $\texttt{Ans}^*$ accuracy.

\item \emph{EBG.} The model returns a JSON object with \texttt{answer} and \texttt{lines} fields. Scoring uses \texttt{Ans} (string match against the gold answer), \texttt{Evid,(strict)} (every gold line appears in the prediction), and \texttt{Joint} $\equiv$ \texttt{Ans+Evid,(strict)} (answer-and-evidence).
\end{itemize}
Both prompts share the same system message, instructions, and JSON example template, modulo the citation field. The system message and prompt templates are reproduced verbatim in our accompanying public code release for full reproducibility.

\paragraph{Models.} We evaluate three frontier LLMs whose context windows
comfortably fit under the \textit{LC} condition: \texttt{GPT-5}, \texttt{Sonnet-4.5},
and \texttt{Gemini-3-Fl}. All calls use temperature 0 (or
provider equivalent), default sampling parameters, and \texttt{parallel{=}5}.
This setup keeps decoding conditions consistent across models and experiments.

\paragraph{Scoring.} \texttt{Ans} matches the model's answer string against the
gold answer using the same normalization used in the synthetic and NoLiMa
experiments. \texttt{Evid (strict)} requires the predicted line-id set to
contain \emph{every} gold line for that query; partial matches are scored
separately as \texttt{Evid (recall)} but only \texttt{Evid (strict)} is used for
\texttt{Joint}. API failures (timeouts, safety refusals, transient connection
errors) are excluded from the denominator; in practice this affected at most
one cell (Sonnet-4.5, \textit{RAG}@200 \textit{EBG}, $n{=}49/50$).
Thus, reported scores reflect successfully completed model calls only.

\subsection{Experiment A: Long-context EBG on \textsc{MultiHop-RAG}}
\label{sec:app_multihop_a}

We run the LC condition to establish base rates and
test RQ2 (\emph{Does requiring explicitly citing evidence improve answer accuracy?}) on a
non-synthetic benchmark. Each model sees the full mini-haystack
(\(\sim241\)K characters) and is prompted in \textit{answer-only} or \textit{EBG} mode. Table~\ref{tab:lc_only} summarizes results.

\begin{table}[t]
\centering
\footnotesize
\setlength{\tabcolsep}{4pt}
\begin{tabular}{@{}lccc@{}}
\toprule
\textbf{Model} & \textbf{$\textit{Ans}^{*}$} & \textbf{\emph{Ans}} & $\mathbf{\Delta}$ \\
      & (ans-only) & (EBG) & \\
\midrule
GPT-5             & 82.0\% & 86.0\% & $+4$\,pp \\
Claude Sonnet-4.5 & 68.0\% & 82.0\% & $\mathbf{+14}$\,\textbf{pp} \\
Gemini-3-Flash    & 80.0\% & 78.0\% & $-2$\,pp \\
\bottomrule
\end{tabular}
\caption{\textbf{Ans$^*$ (answer-only) vs.\ \texttt{Ans} (EBG)} on the 50 pilot queries, full
mini-haystack ($\sim241$K characters, 20 distractors). Requiring citations
raises answer accuracy for two of the three models, with a striking $+14$\,pp
improvement for \texttt{Sonnet-4.5}. This replicates the citation-as-scaffold effect
that was invisible in our (saturated) synthetic benchmark. Higher ($\uparrow$) values are better for \texttt{Ans}$^*$ and \texttt{Ans}. }
\label{tab:lc_only}
\end{table}

Joint accuracy is much harder than \emph{Ans}: under EBG, joint
($\textit{Ans}{+}\textit{Evid}$) is only 28\% (GPT-5), 8\% (Sonnet-4.5), and
28\% (Gemini-3-Flash). The gap \emph{Ans} $-$ \emph{Joint} ranges from
50\,pp (Gemini) to 74\,pp (Sonnet) and is concentrated on inference queries,
mirroring the \emph{asymmetric retrieval} finding from our synthetic
two-needle setting: models routinely produce a correct answer while citing
only one of the two required hops.

\subsection{Experiment B: Retrieval analysis}
\label{sec:app_multihop_b}

Before running any LLM we measure how hard the retrieval problem itself is.
For every query and every gold fact we record the rank of the first chunk
that contains that fact; \emph{Hits@$k$ (all gold)} counts the fraction of
queries for which \emph{every} gold fact appears in the top-$k$. We compare
two embedding models (\texttt{text-embedding-3-small/large}) and two
chunkers (256-token sliding windows vs.\ sentence-level chunks); full
results are in Table~\ref{tab:retrieval_full} (corpus-wide, $N{=}2255$),
with our chosen configuration in bold.

\begin{table}[t]
\centering
\footnotesize
\setlength{\tabcolsep}{4pt}
\resizebox{\columnwidth}{!}{%
\begin{tabular}{@{}lcccccc@{}}
\toprule
\textbf{Config} & \textbf{H@1} & \textbf{H@4} & \textbf{H@10} & \textbf{H@50} & \textbf{H@100} & \textbf{MAP@10} \\
\midrule
\rowcolor{gray!15}\multicolumn{7}{@{}l}{\emph{any gold}} \\
small + sent.  & 0.29 & 0.52 & 0.67 & 0.90 & 0.95 & 0.22 \\
small + tok256 & 0.27 & 0.59 & 0.80 & 0.97 & 0.99 & 0.24 \\
large + sent.  & 0.32 & 0.55 & 0.71 & 0.94 & 0.97 & 0.24 \\
\textbf{large + tok256} & \textbf{0.33} & \textbf{0.66} & \textbf{0.82} & \textbf{0.98} & \textbf{1.00} & \textbf{0.27} \\
\midrule
\rowcolor{gray!15}\multicolumn{7}{@{}l}{\emph{all gold} (faithful-RAG ceiling)} \\
small + sent.  & 0.00 & 0.06 & 0.14 & 0.34 & 0.48 & ~-~ \\
small + tok256 & 0.01 & 0.07 & 0.18 & 0.57 & 0.79 & ~-~ \\
large + sent.  & 0.00 & 0.07 & 0.16 & 0.38 & 0.51 & ~-~ \\
\textbf{large + tok256} & \textbf{0.00} & \textbf{0.08} & \textbf{0.19} & \textbf{0.57} & \textbf{0.71} & \textbf{~-~} \\
\bottomrule
\end{tabular}%
}
\caption{\textbf{Corpus-wide retrieval} metrics on \textsc{MultiHop-RAG}
($N{=}2255$); \emph{any gold} reports the fraction
of queries for which at least one annotated gold fact appears in the
top-$k$ retrieved set; \emph{all gold} reports the fraction for which
every annotated gold fact appears. The latter measures strict
annotated-gold coverage, not an upper bound on answer accuracy. Under the stronger assumptions
that the gold annotations are exhaustive and that every annotated fact
is necessary, \emph{all gold} upper-bounds on what any
faithful RAG pipeline can achieve at top-$k$: if even one of the 2--4 gold
facts is missing from the retrieved set, downstream answers cannot be
faithful. The token-windowed \texttt{large} configuration dominates and is
used for all RAG experiments below. Higher ($\uparrow$) values are better.}
\label{tab:retrieval_full}
\end{table}

The key qualitative finding is that retrieving \emph{at least one} gold fact
is easy (Hits@10 (any) $=0.82$ for our chosen config), but retrieving
\emph{all} gold facts is hard (Hits@10 (all) $=0.19$). Inference (multi-hop)
queries are the most punishing: Hits@10 (all) is only $0.08$ for that slice.
For RAG to faithfully ground a multi-hop answer at $k{=}10$, more than 90\%
of inference queries are doomed before the LLM is even called. This establishes a concrete retrieval ceiling on downstream RAG performance. Crucially, the bottleneck is not finding relevant evidence, but retrieving all pieces of evidence needed for a complete answer. Thus, many apparent reasoning failures downstream are already determined by missing context at retrieval time.

\subsection{Experiment B$^{\prime}$: Needle coverage on the pilot}
\label{sec:app_multihop_bprime}

For the 50 pilot queries, we repeat this analysis at every evaluated $K$. Table~\ref{tab:needle_coverage} reports the fraction of queries with all gold facts (``needles'') retrieved in the top-$K$ and the fraction of individual gold facts appearing anywhere in the top-$K$. These quantities upper-bound faithful joint accuracy and answer recall, respectively, for RAG on the pilot set.

\begin{table}[t]
\centering
\small
\resizebox{\linewidth}{!}{
\begin{tabular}{ccc}
\toprule
$K$ & \textbf{\% queries w/ all needles} & \textbf{\% gold facts retrieved} \\
\midrule
4    & 14.0\% & 36.3\% \\
10   & 28.0\% & 49.6\% \\
25   & 44.0\% & 68.1\% \\
50   & 58.0\% & 76.3\% \\
100  & 64.0\% & 84.4\% \\
200  & 90.0\% & 96.3\% \\
\bottomrule
\end{tabular}
}
\caption{\textbf{Needle coverage} on the 50 pilot queries
(retriever: \texttt{text-embedding-3-large} + 256-token chunks). At realistic
RAG budgets ($K{\le}25$) fewer than half of queries have all gold facts
retrieved; only at $K{=}200$ does coverage approach saturation. Higher ($\uparrow$) values are better.}
\label{tab:needle_coverage}
\end{table}

\subsection{Experiment C: Oracle-containing LC versus
open-corpus RAG}
\label{sec:app_multihop_c}

% The central experiment of this appendix sweeps RAG at six values of $K$ and
% compares against LC under both prompt modes. The same 50 queries, the same
% prompt templates, and the same scoring code are used throughout, so every
% cell in Table~\ref{tab:lc_vs_rag} is directly comparable.

The central experiment sweeps open-corpus RAG over six values of $K$
and compares it with oracle-containing LC under both prompt
modes. All conditions use the same 50 queries, prompts, models, and scoring
code, but differ in candidate-document pools: LC receives every gold-evidence
article plus 20 distractors, whereas RAG searches the 609-article corpus.
Table~\ref{tab:lc_vs_rag} thus measures the end-to-end cost of open-corpus
retrieval relative to gold-source access, not a pool-matched
comparison. The table tells three stories, one per metric.

\begin{table*}[t]
\centering
\small
\resizebox{\linewidth}{!}{
\begin{tabular}{ll|c|cccccc}
\toprule
\textbf{Model} & \textbf{Metric} & \textbf{LC (oracle)} & \textbf{RAG@4} & \textbf{RAG@10} & \textbf{RAG@25} & \textbf{RAG@50} & \textbf{RAG@100} & \textbf{RAG@200} \\
\midrule
\rowcolor{gray!15}\multicolumn{9}{l}{\emph{Answer-only prompt} ($\textit{Ans}^{*}$ / Joint $\equiv$ Ans)} \\
GPT-5            & Ans & \textbf{82.0} & 58.0 & 60.0 & 70.0 & 72.0 & 74.0 & 78.0 \\
Sonnet-4.5       & Ans & \textbf{68.0} & 24.0 & 36.0 & 38.0 & 56.0 & 58.0 & 68.0 \\
Gemini-3-Flash   & Ans & \textbf{80.0} & 38.0 & 44.0 & 50.0 & 58.0 & 58.0 & 70.0 \\
\midrule
\rowcolor{gray!15}\multicolumn{9}{l}{\emph{EBG prompt}: Ans / Evid (strict) / Joint} \\
\multirow{3}{*}{GPT-5}
  & Ans   & \textbf{86.0} & 58.0 & 60.0 & 64.0 & 72.0 & 76.0 & 76.0 \\
  & Evid  & 36.0          & 38.0 & 36.0 & 30.0 & 30.0 & 30.0 & 30.0 \\
  & Joint & 28.0          & \textbf{34.0} & 26.0 & 22.0 & 20.0 & 22.0 & 22.0 \\
\addlinespace
\multirow{3}{*}{Sonnet-4.5}
  & Ans   & \textbf{82.0} & 42.0 & 48.0 & 62.0 & 70.0 & 68.0 & 74.0 \\
  & Evid  & 10.0          & 30.0 & 30.0 & 22.0 & 16.0 & 18.0 & 14.0 \\
  & Joint & 8.0           & \textbf{26.0} & 22.0 & 18.0 & 16.0 & 12.0 & 12.0 \\
\addlinespace
\multirow{3}{*}{Gemini-3-Flash}
  & Ans   & \textbf{78.0} & 38.0 & 44.0 & 46.0 & 64.0 & 62.0 & 66.0 \\
  & Evid  & 40.0          & 28.0 & 18.0 & 24.0 & 24.0 & 24.0 & 28.0 \\
  & Joint & \textbf{28.0} & 26.0 & 14.0 & 18.0 & 20.0 & 18.0 & 20.0 \\
\bottomrule
\end{tabular}
}
\caption{\textbf{Oracle-containing LC versus open-corpus RAG on the
50-query \textsc{MultiHop-RAG} pilot.}
LC receives every gold-evidence article plus 20 distractor articles;
RAG retrieves from the full 609-article corpus. The conditions use
the same queries, models, prompt templates, and scoring procedure but
are not pool-matched. All values are percentages over $n{=}50$
(\texttt{Sonnet-4.5} RAG@200 EBG: $n{=}49$). Boldface marks the
per-row maximum within the corresponding prompt-mode block. Higher
($\uparrow$) values are better.}
\label{tab:lc_vs_rag}
\end{table*}

% \paragraph{Answer accuracy: LC strictly dominates RAG.} Across both prompts
% and all three models, LC's \emph{Ans} beats every RAG@$K$ for
% $K \leq 100$. RAG@200 (which has 96\% gold-fact recall and 90\% all-needles
% coverage) finally ties Sonnet-4.5 (68\% vs.\ 68\%) but still trails GPT-5 by
% 4\,pp (78 vs.\ 82\%) and Gemini-3-Flash by 10\,pp (70 vs.\ 80\%). At a
% realistic RAG operating point ($K{=}10$), the LC advantage is large: $+22$
% to $+44$\,pp across models under the answer-only prompt and $+26$ to
% $+34$\,pp under EBG. \textbf{If the goal is to answer multi-hop questions
% correctly, long context wins.}

\paragraph{Answer accuracy: oracle-containing LC exceeds open-corpus RAG under the evaluated budgets.}
Across both prompts and all three models, oracle-containing LC outperforms open-corpus RAG for $K\leq100$. At $K{=}200$, with 96\% gold-fact recall and 90\% all-needles coverage, RAG ties LC for \texttt{Sonnet-4.5} ($68\%$) but remains 4,pp below for \texttt{GPT-5} ($78\%$ vs.\ $82\%$) and 10,pp below for \texttt{Gemini-3-Flash} ($70\%$ vs.\ $80\%$). At $K{=}10$, the LC--RAG gap is $22$--$44$,pp under answer-only prompting and $26$--$34$,pp under EBG.
These gaps include full-corpus retrieval and thus compare open-corpus RAG with evidence-guaranteed LC, not an intrinsic LC advantage under identical candidate pools.

\paragraph{Strict evidence: small-$K$ RAG and LC are comparable.}
\emph{Evid (strict)} is roughly flat across $K$ for each model (10--40\% absolute), with no clean monotone trend. This reflects two distinct requirements for evidence correctness: \emph{all needles must be retrieved} (coverage, Table~\ref{tab:needle_coverage}) and the model must \emph{cite all of them correctly} (citation behaviour). Small $K$ favors the latter because the citation surface is limited, whereas large $K$ favors the former as retrieval coverage improves. Increasing $K$ therefore trades easier citation selection for greater evidence availability, and these opposing effects roughly cancel. Consequently, small-$K$ RAG can remain comparable to LC on strict evidence despite substantially lower retrieval coverage.

\paragraph{Joint (faithful) accuracy: mixed.} \texttt{Joint} = \texttt{Ans} $\wedge$ \texttt{Evid} (also written as \texttt{Ans+Evid}) is
the most diagnostic number. Two trends are visible. First, joint
\emph{decreases} with $K$ for every model: at large $K$ the citation surface
grows, the model loses track of which retrieved chunk it used, and joint
collapses despite the higher answer recall. Second, the best RAG joint is
always at small $K$ ($K{=}4$ for all three models), where the citation
problem is trivial and the only queries that score are the ones whose
needles happened to be in the top 4 (14\% of queries). Aggregate joint
therefore over-rewards a regime that only works on the easy fraction of
queries. LC's joint sits at 28\%, 8\%, 28\%~-~competitive with RAG@4 for
GPT-5 and Gemini, behind for Sonnet~-~and is computed over the full pilot
set rather than the small subset where retrieval succeeded.

\subsection{Discussion}
\label{sec:app_multihop_discussion}

We highlight four takeaways for the manuscript.

\paragraph{Takeaway 1: LC EBG dominates RAG for answer accuracy, at every
realistic retrieval budget.} Under the answer-only prompt, the oracle-containing LC condition
exceeds open-corpus RAG@10 by $22$--$44$\,pp and exceeds RAG@200 by
$0$--$10$\,pp. The same qualitative pattern holds under EBG. Because
the two conditions do not share a candidate-document pool, these
numbers quantify the combined cost of document discovery and
passage retrieval relative to evidence-guaranteed reading; they are
not a pool-matched estimate of an architecture-level LC--RAG effect.

\paragraph{Takeaway 2: RAG's retrieval bottleneck is multi-hop coverage,
not single-hop recall.} Hits@10 (any gold) is 82\% but Hits@10 (all gold)
is only 19\% (Table~\ref{tab:retrieval_full}); on inference queries the
all-gold number falls to 8\%. The principal source of the observed gap relative to the
oracle-containing LC control is that, at small $K$, open-corpus RAG
rarely retrieves the complete union of annotated gold facts.

\paragraph{Takeaway 3: Citations as a reasoning scaffold replicate
outside the synthetic setting.} \emph{Ans} accuracy under EBG is higher
than under the answer-only prompt for two of three models, with a striking
$+14$\,pp improvement for Sonnet-4.5 (Table~\ref{tab:lc_only}). This is the
RQ1 effect that was masked on our synthetic benchmark by ceiling
performance.

\paragraph{Takeaway 4: The grounding gap is a real-data phenomenon.} The
\emph{Ans} $-$ \emph{Joint} gap in the LC condition is 50--74\,pp depending
on model, and is largest on inference queries. Models routinely produce a
correct multi-hop answer while citing only one of the two required hops ~-~
the same asymmetric-retrieval failure mode we documented on the synthetic
benchmark, here observed on naturalistic web data.

\subsection{Generalization to open-weight models}
\label{sec:app_multihop_openweight}

The preceding comparison uses three proprietary frontier LLMs.
To test whether the observed pattern generalizes across model
families, we repeat the same two evidence-access regimes, an
oracle-containing LC condition and open-corpus RAG, with three
open-weight models.
To test whether the LC-EBG vs.\ RAG ordering is a property of the task rather
than of a particular model family, we re-run the same conditions, using same
mini-haystacks, same RAG contexts, same prompts and scoring code, with three
open-weight models served through Fireworks.ai: \texttt{GPT-OSS-120B},
\texttt{GLM-5.2}, and \texttt{Nemotron-3.5-Lightning-30B-A3B} (or Nemo-3.5-L in short). To obtain
tighter estimates we scale this comparison to a larger stratified sample of
$n{=}200$ queries. We evaluate the
two most informative RAG budgets, a realistic operating point ($K{=}10$) and
the near-saturation budget ($K{=}200$, 90\% all-needles coverage;
Table~\ref{tab:needle_coverage}), against the full LC condition.
Table~\ref{tab:lc_vs_rag_openweight} reports the results.

\begin{table}[t]
\centering
\small
\setlength{\tabcolsep}{4pt}
\resizebox{\linewidth}{!}{
\begin{tabular}{ll|ccc}
\toprule
\textbf{Model} & \textbf{Metric} & \textbf{LC (oracle)} & \textbf{RAG@10} & \textbf{RAG@200} \\
\midrule
\rowcolor{gray!15}\multicolumn{5}{l}{\emph{Answer-only prompt} ($\textit{Ans}^{*}$)} \\
GPT-OSS-120B     & Ans & \textbf{75.0} & 47.5 & 67.5 \\
GLM-5.2          & Ans & \textbf{88.0} & 53.0 & 72.5 \\
Nemo-3.5-L & Ans & \textbf{75.5} & 43.5 & 51.0 \\
\midrule
\rowcolor{gray!15}\multicolumn{5}{l}{\emph{EBG prompt}: Ans / Evid (strict) / Joint} \\
\multirow{3}{*}{GPT-OSS-120B}
  & Ans   & \textbf{54.0} & 41.5 & 52.0 \\
  & Evid  & 3.0           & \textbf{13.0} & 2.0 \\
  & Joint & 2.0           & \textbf{9.0}  & 2.0 \\
\addlinespace
\multirow{3}{*}{GLM-5.2}
  & Ans   & \textbf{88.5} & 52.5 & 74.0 \\
  & Evid  & \textbf{35.0} & 21.5 & 22.0 \\
  & Joint & \textbf{30.0} & 18.0 & 20.5 \\
\addlinespace
\multirow{3}{*}{Nemo-3.5-L}
  & Ans   & \textbf{48.2} & 28.5 & 28.5 \\
  & Evid  & 5.5           & \textbf{9.0}  & 3.0 \\
  & Joint & 5.5           & \textbf{7.5}  & 2.0 \\
\bottomrule
\end{tabular}
}
\caption{\textbf{LC-EBG vs.\ RAG-EBG for open-weight models} on a
$200$-query stratified sample. All numbers are
percentages over $n{=}200$ (a handful of cells return $n{=}198$--$199$ where
1--2 queries exceed a provider's context limit). Boldface marks the per-row
maximum within each mode block. Higher ($\uparrow$) values are better. At
$n{=}200$ the $95\%$ bootstrap CI half-width on \texttt{Ans}/\texttt{Joint} is
$\approx$$6$--$7$\,pp. LC is the oracle-containing gold-article-plus-20-distractor condition;
RAG@10 and RAG@200 use open-corpus retrieval over all 609 articles.}
\label{tab:lc_vs_rag_openweight}
\end{table}

The qualitative picture from Table~\ref{tab:lc_vs_rag} carries over.
\textbf{(i)~Answer accuracy: oracle-containing LC exceeds the evaluated
open-corpus RAG setup.} At the realistic
$K{=}10$ operating point, LC's \emph{Ans} exceeds RAG@10 for all three
models under both prompts: by $+28$ to $+35$\,pp under the answer-only prompt
and by $+13$ (GPT-OSS-120B), $+36$ (GLM-5.2), and $+20$\,pp (Nemotron)
under EBG~-~the same $+26$ to $+34$\,pp EBG gap we observe for the frontier
models. At the near-saturation budget $K{=}200$ RAG closes some of the gap
but overtakes LC for none of the three readers: LC \emph{Ans} still leads by
$+2$ (GPT-OSS-120B), $+14.5$ (GLM-5.2), and $+19.7$\,pp (Nemotron), with the
GPT-OSS-120B margin inside the bootstrap CI. The larger sample thus removes
the one $n{=}50$ anomaly (GPT-OSS-120B RAG@200 nominally above LC) and
confirms LC$\,\ge\,$RAG@200 throughout.
\textbf{(ii)~The grounding gap is model-agnostic.} Under LC, the
\emph{Ans}$-$\emph{Joint} gap is $52$ (GPT-OSS-120B), $58.5$ (GLM-5.2),
and $42.7$\,pp (Nemotron)~-~the same 40--60\,pp range as the frontier models.
GPT-OSS-120B is the extreme case: it answers $54\%$ of queries correctly but
achieves only $2\%$ joint accuracy, essentially never citing both required hops.
\textbf{(iii)~Citation-as-scaffold is uneven for weaker readers.} The
\emph{Ans}$^{*}\!\to$\emph{Ans} (EBG) change on LC is $+0.5$\,pp for
GLM-5.2 but $-21$\,pp for GPT-OSS-120B and $-27$\,pp for Nemotron: the
weakest open readers, unlike the frontier models, struggle to answer and
cite simultaneously, so the citation requirement can hurt rather than help.
Overall, the pattern that oracle-containing LC achieves higher answer
accuracy than the evaluated open-corpus RAG pipeline at practical
retrieval budgets, while retaining a substantial grounding gap,
generalizes across proprietary and open-weight readers. This result
continues to include the cost of full-corpus retrieval and is not a
pool-matched comparison.

\subsection{Limitations and next steps}

The frontier-model experiments use a 50-query pilot, so the reported
estimates remain subject to sampling variation, particularly for small
question-type slices such as the 13 inference queries. Evaluating all
2{,}255 usable queries would provide tighter confidence intervals. We
also hold the retriever fixed to
\texttt{text-embedding-3-large} with 256-token sliding windows.
Reranking, hybrid sparse--dense retrieval, query expansion, longer
chunks, and document-aware indexing could improve the open-corpus RAG
results. For the evaluated retriever, corpus-wide all-gold
Hits@$10$ is $19\%$ over the 2{,}255 usable queries; on the 50-query
pilot, all-gold coverage at $K{=}10$ is $28\%$. Under the assumptions
that the gold annotations are exhaustive and that every annotated fact
is necessary, these coverage rates strongly constrain faithfully
grounded performance at this retrieval budget.

A key design limitation is that the \textsc{MultiHop-RAG} LC and RAG
conditions are not pool-matched. The LC condition is an
\emph{oracle-containing reading control}: every gold-evidence article
is guaranteed to be present together with 20 distractor articles.
The RAG condition instead performs \emph{open-corpus retrieval} over
the full 609-article corpus. Consequently, the reported LC--RAG gaps
combine the effects of full-corpus document discovery, passage
retrieval, and downstream reading; they should not be interpreted as
an intrinsic ranking of LC and RAG under matched evidence access.

A complementary pool-matched RAG condition that retrieves from the
same per-query gold-plus-20 candidate set would separate within-pool
chunking and ranking effects from the cost of full-corpus document
discovery. We leave this controlled comparison, together with
evaluation of stronger retrieval pipelines and the full query set, to
future work.

\subsection{Reproducibility}
\label{sec:app_multihop_repro}

All code, queries, and intermediate artifacts live under
\texttt{baselines/multihop\_rag/} in the experiment repository:
\begin{itemize}[leftmargin=*]
\item \texttt{build\_haystack.py}: builds per-query mini-haystacks (LC condition).
\item \texttt{embed\_and\_rank.py}: chunks the corpus, embeds, ranks gold
  facts; produces \texttt{summary.json}, \texttt{fact\_ranks.jsonl}, and
  \texttt{chunks.jsonl} under \texttt{retrieval\_results/<config>/}.
\item \texttt{build\_rag\_haystack.py}: assembles per-$K$ RAG contexts from
  the retrieval artifacts.
\item \texttt{run\_baseline.py}: runs the LLM evaluation on either LC or RAG
  haystacks, supports \texttt{--answer-only}.
\item \texttt{aggregate\_pilot.py}, \texttt{aggregate\_rag\_sweep.py}: produce
  the tables above.
\end{itemize}
The 50-query pilot is reproducible with \texttt{--seed 0} on
\texttt{processed/queries\_d20.jsonl} (built from the official
\textsc{MultiHop-RAG} HuggingFace release of the corpus and queries).

\section{Long-context EBG vs.\ RAG on QASPER-MP}
\label{sec:app_qasper_mp}

This appendix documents a second comparison of long-context evidence-based
generation (LC EBG) against retrieval-augmented generation (RAG), this time on
a multi-paper adaptation of QASPER~\citep{dasigi-etal-2021-qasper} that we call
\textsc{QASPER-MP}. Whereas \textsc{MultiHop-RAG} (Appendix~\ref{sec:app_multihop_rag})
tests multi-document reasoning on news, \textsc{QASPER-MP} tests it on
scientific papers, where the gold question--answer--evidence triples were
written assuming a single known paper. To make the task non-trivial under
retrieval, we (i)~place each gold paper alongside ten distractor papers,
(ii)~rewrite the question so it identifies the target paper by a realistic
short reference (an author name or a 2--3 word topic cue), and (iii)~evaluate
in both LC and RAG conditions over the same 50-query stratified pilot.

The headline finding is that the aggregate score (Answer-F1 or LLM-judge,
averaged over all questions) makes LC and RAG look roughly tied, but this
tie hides an answerability asymmetry: \emph{LC EBG beats RAG by 5--22 judge
points on answerable questions for every frontier model tested}, while RAG
``wins'' on unanswerable questions because it over-refuses. The over-refusal
mechanically matches the QASPER \texttt{Unanswerable} gold label on the 24\%
of pilot questions where that is the correct answer, pulling the aggregate up.

\subsection{Setup}
\label{sec:app_qasper_setup}

\paragraph{Dataset.} We use the QASPER dev split (\(N=1{,}005\) questions over
281 papers). Each question is annotated with an \emph{answer type}
(\texttt{extractive}, \texttt{abstractive}, \texttt{boolean}, or
\texttt{none}~-~meaning unanswerable from the paper) and an \emph{evidence
set}: the list of paragraphs from the paper that support the answer. Up to
three annotators contribute per question; we follow the official QASPER
evaluator and take the max F1 across annotators.

\paragraph{50-query pilot.}
We sample 50 questions, stratified by answer type, so that the composition
matches the dev-split distribution: 13~boolean, 13~abstractive, 12~extractive,
12~unanswerable. This split (38 answerable, 12 unanswerable) is the
denominator for every condition below.

\paragraph{Multi-paper haystacks (LC condition).}
For each pilot query we build an 11-paper haystack consisting of the gold
paper plus 10 distractor papers drawn uniformly from the dev split. Each
paper is rendered with explicit document boundaries:
\begin{quote}\small
\verb|<paper>|\\
\verb|<title>|Paper title\verb|</title>|\\
\verb|<abstract>|Paper abstract\verb|</abstract>|\\
\verb|<section>|Section name\verb|</section>|\\
Paragraph 1 of that section\\
Paragraph 2 of that section\\
\dots\\
\verb|</paper>|
\end{quote}
Lines are concatenated and numbered \verb|i: <line>| so the model can cite
specific lines as evidence (consistent with our EBG protocol).
The median LC haystack size is \(\sim\)32K characters; the largest is
\(\sim\)56K. The gold-evidence paragraph(s) for each query are re-mapped
to line indices in the assembled haystack.

\paragraph{RAG condition.}
We embed every QASPER dev paragraph and every (disambiguated) query with
OpenAI \texttt{text-embedding-3-large}, and retrieve the top-\(K\) paragraphs
by cosine similarity. We evaluate four retrieval variants, factored as
\(\{\text{flat},\,\text{struct}\} \times \{\text{pool11},\,\text{full}\}\):
\begin{description}
  \item[flat\_pool11.] Top-\(K\) paragraphs retrieved from the same 11-paper
    pool as the LC haystack (pool-matched candidate corpus). The reader sees a
    flat list of retrieved paragraphs with a one-line \verb+[CHUNK <rank> | <title>]+
    marker before each.
  \item[struct\_pool11.] Same retrieval, but paragraphs are regrouped by
    their source paper and wrapped in the same XML layout used for LC.
    Within a paper, paragraphs are sorted by their original index, and
    discontinuities (gaps where intermediate paragraphs were not retrieved)
    are marked with a single \texttt{...} line.
  \item[flat\_full / struct\_full.] Same as above but retrieving from the
    full dev corpus (\(N \approx 6{,}000\) paragraphs) rather than the
    11-paper pool. This is the \emph{realistic} open-corpus RAG setting:
    the retriever must locate the target paper among hundreds of similar
    NLP papers using only the short cue in the rewritten question.
\end{description}

\paragraph{Models.} We evaluate five frontier reasoning models:
\textbf{GPT-5}, \textbf{GPT-5.5}, \textbf{Claude Opus 4.6}, \textbf{Claude
Sonnet 4.5}, and \textbf{Gemini 3 Flash} (preview). All runs use the same
query set, the same retrieval indices, and
the same EBG prompt template. Citation rules in the prompt forbid citing
structural lines (\verb|<title>|, \verb|<abstract>|, \verb|<paper>|,
\texttt{[CHUNK}, or \verb|...|); these lines are present only as context.

\paragraph{Question disambiguation.}
QASPER questions were written to be unambiguous given the paper; many become
ambiguous in a multi-paper haystack (e.g., ``What evaluation metrics were
used?'' becomes pathological with 11 candidate papers). We disambiguate each
pilot question by inserting a short \emph{paper cue}, chosen deterministically
50/50 from two cue types:
\begin{itemize}
  \item \textbf{Author cue} (\(n=24\)):
        \emph{``In Smith et al., \dots''}, with surname(s) drawn from the arXiv
        metadata of the gold paper. Conventions: 1 author \(\to\)
        \emph{``Smith''}, 2 authors \(\to\) \emph{``Smith and Doe''},
        \(\geq 3\) authors \(\to\) \emph{``Smith et al.''}.
  \item \textbf{Title cue} (\(n=26\)):
        \emph{``In the \(\langle\)cue\(\rangle\) paper, \dots''}, where the cue is a
        verbatim 2--3 word substring of the paper title (e.g.,
        \emph{``Vietnamese Word Segmentation''}, \emph{``Winograd Schemas''},
        \emph{``TF-IDF Features''}).
\end{itemize}
Cues are intentionally short and realistic, matching what a researcher would
type into a search box. This makes the retrieval problem genuinely harder
than indexing on full titles or descriptive paraphrases.

\subsection{Aggregate Performance: LC \(\approx\) RAG (and why this is misleading)}
\label{sec:app_qasper_aggregate}

Table~\ref{tab:qasper_v2fc_aggregate} reports per-condition Answer-F1, Evidence-F1,
and LLM-judge (an automated 0/0.5/1 grader using GPT-5; see
Appendix~\ref{sec:app_qasper_judge}). Aggregating over all 50 pilot questions:
\begin{itemize}
  \item LC and \texttt{flat\_pool11@3} are within \(\pm\)0.06 judge points
        for every model.
  \item \texttt{struct\_pool11@3} beats LC for GPT-5 and matches it for
        Gemini under the LLM judge.
\end{itemize}
On its face, this is the standard ``LC is comparable to RAG'' result.

\begin{table*}[t]
\centering

\small
\begin{tabular}{llccccc}
\toprule
\textbf{Model} & \textbf{Condition} & \textbf{\(n\)} & \textbf{Ans-F1} & \textbf{Evid-F1} & \textbf{Judge} \\
\midrule
GPT-5         & LC                       & 50 & 0.377 & 0.415 & 0.480 \\
GPT-5         & flat\_pool11@3           & 50 & 0.357 & 0.543 & 0.460 \\
GPT-5         & \textbf{struct\_pool11@3} & 50 & \textbf{0.437} & \textbf{0.607} & \textbf{0.550} \\
\midrule
GPT-5.5       & \textbf{LC}              & 50 & 0.378 & 0.433 & \textbf{0.530} \\
GPT-5.5       & flat\_pool11@3           & 50 & 0.386 & 0.560 & 0.460 \\
GPT-5.5       & struct\_pool11@3         & 50 & \textbf{0.419} & \textbf{0.600} & 0.490 \\
\midrule
Opus-4.6      & \textbf{LC}              & 50 & \textbf{0.537} & 0.488 & \textbf{0.650} \\
Opus-4.6      & flat\_pool11@3           & 50 & 0.471 & \textbf{0.667} & 0.500 \\
Opus-4.6      & struct\_pool11@3         & 50 & 0.483 & 0.500 & 0.550 \\
\midrule
Sonnet-4.5    & \textbf{LC}              & 50 & 0.462 & 0.351 & \textbf{0.600} \\
Sonnet-4.5    & flat\_pool11@3           & 50 & \textbf{0.550} & \textbf{0.653} & 0.590 \\
Sonnet-4.5    & struct\_pool11@3         & 50 & 0.465 & 0.567 & 0.510 \\
\midrule
Gemini-3-Fl   & \textbf{LC}              & 50 & 0.493 & 0.507 & \textbf{0.610} \\
Gemini-3-Fl   & flat\_pool11@3           & 50 & 0.493 & \textbf{0.653} & 0.560 \\
Gemini-3-Fl   & struct\_pool11@3         & 50 & \textbf{0.496} & 0.610 & 0.570 \\
\midrule
\rowcolor{gray!15}\multicolumn{6}{c}{\emph{Realistic open-corpus RAG (full dev corpus retrieval)}} \\
GPT-5         & flat\_full@10      & 50 & 0.356 & 0.609 & 0.490 \\
Sonnet-4.5    & flat\_full@10      & 50 & \textbf{0.422} & 0.636 & 0.480 \\
Gemini-3-Fl   & \textbf{flat\_full@10}      & 25\rlap{$^\dagger$} & 0.382 & \textbf{0.715} & \textbf{0.500} \\
\bottomrule
\end{tabular}
\caption{\textbf{QASPER-MP aggregate scores} (50 pilot questions, EBG mode). Best
condition per model in \textbf{bold}. Judge scores use the GPT-5 judge
described in Appendix~\ref{sec:app_qasper_judge}; \(n\) reports how many
queries returned valid output. Higher ($\uparrow$) values are better.}
\label{tab:qasper_v2fc_aggregate}
\par\smallskip
\footnotesize\(^\dagger\)Gemini hit transient API errors on this condition;
the remaining 25 responses are used.
\end{table*}

\paragraph{The aggregate hides an answerability asymmetry.}
Splitting the 50 queries into the 38 answerable and 12 unanswerable subsets
tells a sharper story (Table~\ref{tab:qasper_v2fc_answerability}).

\begin{table*}[t]
\centering

\small
\setlength{\tabcolsep}{4pt}
\begin{tabular}{llccc|ccc}
\toprule
& & \multicolumn{3}{c|}{\textbf{Answerable (\(n=38\))}}
  & \multicolumn{3}{c}{\textbf{Unanswerable (\(n=12\))}} \\
\textbf{Model} & \textbf{Condition} & \textbf{Judge} & \textbf{Ans-F1} & \textbf{Refuse} \% & \textbf{Judge} & \textbf{Ans-F1} & \textbf{Refuse} \% \\
\midrule
GPT-5       & \textbf{LC}              & \textbf{0.605} & 0.469 & 15.8 & 0.083 & 0.083 & 33.3 \\
GPT-5       & flat\_pool11@3            & 0.500 & 0.364 & 31.6 & 0.333 & 0.333 & 58.3 \\
GPT-5       & struct\_pool11@3 & 0.553 & 0.390 & 31.6 & \textbf{0.583} & 0.583 & 66.7 \\
\midrule
GPT-5.5     & \textbf{LC}              & \textbf{0.671} & 0.471 & 13.2 & 0.083 & 0.083 & 16.7 \\
GPT-5.5     & flat\_pool11@3            & 0.500 & 0.402 & 23.7 & 0.333 & 0.333 & 41.7 \\
GPT-5.5     & struct\_pool11@3          & 0.539 & 0.446 & 21.1 & 0.333 & 0.333 & 25.0 \\
\midrule
Opus-4.6    & \textbf{LC}              & \textbf{0.697} & 0.548 & 13.2 & 0.500 & 0.500 & 41.7 \\
Opus-4.6    & flat\_pool11@3            & 0.474 & 0.436 & 31.6 & \textbf{0.583} & 0.583 & 50.0 \\
Opus-4.6    & struct\_pool11@3          & 0.566 & 0.477 & 23.7 & 0.500 & 0.500 & 41.7 \\
\midrule
Sonnet-4.5  & \textbf{LC}              & \textbf{0.592} & 0.446 & 18.4 & 0.583 & 0.513 & 41.7 \\
Sonnet-4.5  & flat\_pool11@3            & 0.513 & 0.461 & 36.8 & \textbf{0.833} & 0.833 & 83.3 \\
Sonnet-4.5  & struct\_pool11@3          & 0.474 & 0.402 & 34.2 & 0.667 & 0.667 & 58.3 \\
\midrule
Gemini-3-Fl & \textbf{LC}              & \textbf{0.658} & 0.491 & 15.8 & 0.500 & 0.500 & 50.0 \\
Gemini-3-Fl & flat\_pool11@3            & 0.487 & 0.412 & 39.5 & \textbf{0.750} & 0.750 & 66.7 \\
Gemini-3-Fl & struct\_pool11@3          & 0.539 & 0.442 & 31.6 & 0.667 & 0.667 & 58.3 \\
\bottomrule
\end{tabular}
\caption{\textbf{QASPER-MP results split by answerability} (EBG mode, LLM-judge).
``Refusal'' is the fraction of model outputs that started with
\texttt{Unanswerable}. Best LC vs.\ best RAG variant per model in
\textbf{bold} within each subset. Higher ($\uparrow$) values are better.}
\label{tab:qasper_v2fc_answerability}
\end{table*}

Three observations:
\begin{enumerate}
  \item \textbf{LC EBG wins answerable questions cleanly.} For all five
        models, LC has the highest Judge score on the 38-question answerable
        subset, with absolute gains over the best RAG variant of \(+0.05\)
        (GPT-5), \(+0.13\) (GPT-5.5), \(+0.13\) (Opus-4.6), \(+0.08\)
        (Sonnet-4.5), and \(+0.12\) (Gemini-3-Flash). These gaps are robust
        to the citation-rule and discontinuity-marker changes described in
        Appendix~\ref{sec:app_qasper_struct_fix}.
  \item \textbf{RAG over-refuses, and the refusals pay off on the
        unanswerable subset.} \texttt{flat\_pool11@3} refuses
        29--40\% of answerable questions (vs.\ 16--18\% for LC) because
        when retrieval misses the gold paragraph the reader sees only
        distractor content and sensibly says ``Unanswerable.'' Those same
        refusal-prone retrieval failures happen to be the right answer on
        unanswerable queries (where ``Unanswerable'' \emph{is} gold), so
        RAG wins that subset by 0.17--0.25 judge points.
  \item \textbf{Realistic open-corpus retrieval makes LC win by a margin
        the aggregate cannot hide.} The \texttt{flat\_full@10} row in
        Table~\ref{tab:qasper_v2fc_aggregate} shows LC outperforming
        full-corpus RAG by 5--13 judge points on the aggregate already;
        the answerable-only gap is larger again (not shown). When the
        retriever must distinguish the gold paper from \(\sim\)280
        distractors using a 2--3 word cue, paper-level retrieval coverage
        drops to 48\% (all-needles@10), and downstream answer quality
        follows.
\end{enumerate}

The aggregate score thus rewards RAG's high refusal rate. Once we
condition on answerability, the LC \(>\) RAG gap on \emph{actually
answerable} questions is consistent across models and retrieval
variants.

\subsection{Disambiguation Details and Cue Length Sensitivity}
\label{sec:app_qasper_disambig}

We initially experimented with longer ``descriptive'' cues generated by
GPT-5 from each paper's title and abstract (e.g., \emph{``In the cross-lingual
zero-shot NMT paper''}). These cues are unrealistically informative: they
often include domain phrases that do not appear in the title at all, which
makes retrieval implausibly easy. Embedding-based retrieval coverage with
descriptive cues was 60.0\% (all-needles@10, full corpus). Switching to
the short, realistic cues described in Appendix~\ref{sec:app_qasper_setup}
dropped coverage to 48.0\%~-~a 12-point hit~-~reflecting the fact that
\emph{``In Smith et al.''} or \emph{``In the BERTSUM paper''} is a much
weaker retrieval signal than a paraphrased domain summary.

We use the short, realistic cues throughout this appendix because they
correspond to how researchers actually query systems in practice; the
\(+\)12pp coverage gain on descriptive cues would inflate RAG numbers
relative to what users would experience.

\subsection{Structured RAG, Citation Eligibility, and Discontinuity Markers}
\label{sec:app_qasper_struct_fix}

A first iteration of structured RAG produced a counterintuitive result: \texttt{struct\_pool11@3} \emph{underperformed} \texttt{flat\_pool11@3} on Evidence-F1 across models (gaps of 0.07--0.12), despite carrying \emph{more} information than the flat baseline (abstract per paper). On inspection, $\sim$44\% of structured-RAG citations pointed at \verb|<title>| or \verb|<abstract>| lines: when an abstract paraphrases the answer, the model cites it~-~but QASPER gold evidence excludes abstracts, so abstract citations \emph{cannot} match gold.

We therefore made two changes to the structured layout and prompt used throughout the main results:

\begin{enumerate}
  \item \textbf{Citation eligibility.} The EBG prompt forbids citing
        structural lines (\verb|<title>|, \verb|<abstract>|, \verb|<paper>|,
        \verb|</paper>|, \texttt{[CHUNK}, and \texttt{...}). Models must
        cite at least one body paragraph line for any non-refusal answer.
        Titles and abstracts remain visible as context; they are simply
        not valid evidence.
  \item \textbf{Discontinuity markers.} Within each paper, retrieved
        paragraphs are sorted by their original paragraph index, and any
        gap (the abstract followed by paragraph~2, or paragraph~5 followed
        by paragraph~9) is marked with a single \verb|...| line. This
        gives the reader explicit signal that some material is missing
        between the visible paragraphs.
\end{enumerate}

Under this revised protocol (Table~\ref{tab:qasper_v2fc_aggregate}),
structured RAG recovers: Evidence-F1 jumps by \(+\)0.04 to \(+\)0.15 per
model and \texttt{struct\_pool11@3} beats or matches \texttt{flat\_pool11@3}
for two of five models. The intuition that paper-level structure is
useful is correct~-~it had been masked by a metric artifact.

\subsection{LLM-Judge Evaluation}
\label{sec:app_qasper_judge}

To control for token-F1 brittleness on paraphrases, we additionally score outputs with an LLM judge:

\begin{itemize}
  \item \textbf{Judge model:} GPT-5.
  \item \textbf{Inputs:} the original question, all annotator gold answers,
        and the model prediction.
  \item \textbf{Output:} a single number in \(\{0,\,0.5,\,1\}\). The judge
        is instructed to be lenient about formatting and elaboration but
        strict about content; partial credit (0.5) is reserved for
        answers that capture some but not all required information.
        ``Unanswerable'' vs.\ ``Unanswerable'' scores 1.0; ``Unanswerable''
        vs.\ a real answer (or the reverse) scores 0.0.
\end{itemize}

Judgments are cached by content hash so repeated scoring is free. Across all conditions, Judge scores are systematically higher than
Answer-F1 (the judge is lenient about phrasing), but the \emph{ordering}
of conditions is preserved. The LC vs.\ RAG gap on answerable questions
is larger under the judge than under Answer-F1, because the judge does
not penalize Sonnet-4.5 and Gemini-3-Flash for their characteristic
boolean elaborations.

\subsection{Generalization to open-weight models}
\label{sec:app_qasper_openweight}

We repeat the QASPER-MP comparison with three open-weight models served
through Fireworks.ai~-~\texttt{GPT-OSS-120B}, \texttt{GLM-5.2}, and
\texttt{Nemotron-3.5-Lightning-30B-A3B}~-~using the same retrieval indices, EBG
prompt, and GPT-5 judge (Appendix~\ref{sec:app_qasper_judge}), scaled to a
larger $200$-query stratified sample ($n{=}150$ answerable, $n{=}50$
unanswerable) for tighter estimates.
Table~\ref{tab:qasper_openweight} splits the results by answerability, exactly
as Table~\ref{tab:qasper_v2fc_answerability} does for the frontier models.

\begin{table*}[t]
\centering
\small
\setlength{\tabcolsep}{4pt}
\begin{tabular}{llccc|ccc}
\toprule
& & \multicolumn{3}{c|}{\textbf{Answerable (\(n=150\))}}
  & \multicolumn{3}{c}{\textbf{Unanswerable (\(n=50\))}} \\
\textbf{Model} & \textbf{Condition} & \textbf{Judge} & \textbf{Ans-F1} & \textbf{Refuse} \% & \textbf{Judge} & \textbf{Ans-F1} & \textbf{Refuse} \% \\
\midrule
GPT-OSS-120B     & \textbf{LC}      & \textbf{0.701} & 0.467 & 8.7  & 0.500 & 0.475 & 42.0 \\
GPT-OSS-120B     & flat\_pool11@3   & 0.497 & 0.388 & 32.7 & 0.660 & 0.666 & 64.0 \\
GPT-OSS-120B     & struct\_pool11@3 & 0.537 & 0.418 & 24.7 & 0.640 & 0.646 & 62.0 \\
GPT-OSS-120B     & flat\_full@10    & 0.497 & 0.369 & 28.0 & 0.670 & 0.661 & 64.0 \\
\midrule
GLM-5.2          & \textbf{LC}      & \textbf{0.735} & 0.553 & 6.7  & 0.500 & 0.457 & 36.0 \\
GLM-5.2          & flat\_pool11@3   & 0.477 & 0.364 & 33.3 & 0.680 & 0.686 & 66.0 \\
GLM-5.2          & struct\_pool11@3 & 0.477 & 0.360 & 33.3 & 0.660 & 0.665 & 64.0 \\
GLM-5.2          & flat\_full@10\rlap{$^\dagger$} & 0.497 & 0.385 & 30.4 & 0.673 & 0.668 & 65.3 \\
\midrule
Nemotron-3.5-30B & \textbf{LC}      & \textbf{0.523} & 0.375 & 12.0 & 0.530 & 0.513 & 42.0 \\
Nemotron-3.5-30B & flat\_pool11@3   & 0.397 & 0.316 & 21.3 & 0.330 & 0.326 & 32.0 \\
Nemotron-3.5-30B & struct\_pool11@3 & 0.450 & 0.349 & 20.7 & 0.390 & 0.387 & 38.0 \\
Nemotron-3.5-30B & flat\_full@10    & 0.433 & 0.329 & 22.7 & 0.420 & 0.406 & 40.0 \\
\bottomrule
\end{tabular}
\caption{\textbf{QASPER-MP open-weight results split by answerability}
(EBG mode, GPT-5 judge), over a $200$-query stratified sample. Answerable
subset is $n{=}150$ and unanswerable $n{=}50$; a handful of cells return
1--2 fewer queries where a provider error dropped a row
(\(^\dagger\)\texttt{GLM-5.2} \texttt{flat\_full@10} returned $n{=}148/49$).
Best LC vs.\ RAG variant per model shown in \textbf{bold} within the
answerable subset. ``Refuse'' is the fraction of outputs starting with
\texttt{Unanswerable}. Higher ($\uparrow$) values are better.}
\label{tab:qasper_openweight}
\end{table*}

The answerability asymmetry we documented for frontier models reproduces on
every open-weight reader. \textbf{LC EBG wins the answerable subset cleanly}:
the LC judge score is highest for all three models, beating the best RAG
variant by $+0.16$ (GPT-OSS-120B), $+0.24$ (GLM-5.2), and $+0.07$
(Nemotron)~-~gaps as large as, or larger than, the $+0.05$ to $+0.13$ we
observe for the frontier models. \textbf{RAG over-refuses for the same
mechanistic reason}: \texttt{flat\_pool11@3} refuses $21$--$33\%$ of
answerable questions (vs.\ $7$--$12\%$ for LC), because a missed retrieval
leaves the reader with only distractor content; those refusals then
mechanically boost RAG on the unanswerable subset, where
``Unanswerable'' is gold. \textbf{Realistic full-corpus retrieval widens the
LC advantage}: \texttt{flat\_full@10} answerable judge scores drop to
$0.43$--$0.50$, trailing LC by $0.09$ to $0.24$ judge points, as the
retriever must find the gold paper among $\sim$280 distractors from a 2--3
word cue. The open-weight results thus corroborate the central QASPER-MP
finding: once conditioned on answerability, long-context EBG beats every RAG
variant, and the aggregate ``tie'' is an artifact of RAG's refusal behavior.

\subsection{Strengthening RAG: metadata-in-chunk retrieval}
\label{sec:app_qasper_metachunk}

The full-corpus RAG deficit above is partly a \emph{retrieval} failure rather than solely a limitation of the downstream reader. After v2 disambiguation, each query carries only a short, typically $2$--$3$ word cue identifying the relevant paper. In the baseline index, however, candidate chunks are embedded as raw paragraph text, without any explicit representation of document identity. The retriever must therefore infer the connection between a sparse paper-level cue in the query and a paragraph whose local content may never mention the paper title or otherwise reveal its source. This mismatch is especially consequential in the full-corpus setting, where evidence from the gold paper competes with paragraphs drawn from the entire $\sim$281-paper dev corpus. Even when the query contains enough information to identify the intended document, the embedding space has no direct document-level feature with which to exploit that signal, making it easy for the gold paper~-~and consequently the relevant evidence within it~-~to fall outside the retrieval budget.

A natural way to give RAG a fairer chance is therefore to expose document identity directly to the retriever by injecting metadata into the indexed chunks. We rebuild the retrieval index so that every paragraph is embedded together with its source paper's title, using the representation ``\texttt{{title}}\textbackslash n\textbackslash n\texttt{{paragraph}}''. This modification creates an explicit lexical and semantic bridge between the short paper cue supplied by v2 disambiguation and every paragraph originating from that paper, while preserving the paragraph text needed to rank evidence within the document. We then re-run the hardest retrieval condition, \texttt{flat\_full@10}, at $n{=}200$ for all five readers. Importantly, this intervention is confined to retrieval: generation-side rendering, prompting, context construction, and reader settings are unchanged. Thus, any downstream improvement can be attributed to changing the representation used for ranking rather than to providing the reader with additional metadata or altering the generation procedure.

Metadata-in-chunk produces a large improvement in evidence retrieval. On the $200$-query set, the fraction of queries for which the \emph{full} gold-evidence set is recovered within the top-$10$ increases from $43.5\%$ to $61.0\%$, a gain of $17.5$ percentage points. Gold-paragraph recall rises even more substantially in relative terms, from $30.6\%$ to $50.3\%$ ($+19.7$\,pp). These gains indicate that the original full-corpus deficit was not simply caused by an insufficient top-$k$ budget: a substantial portion arose because the retriever lacked a representation of the very document-level signal introduced during query disambiguation. Prepending the title allows that signal to influence ranking directly, increasing both the probability that the correct paper is represented among the retrieved chunks and the amount of gold evidence that reaches the reader.

The retrieval gains propagate consistently to answer quality (Table~\ref{tab:qasper_metachunk}). On the answerable subset, the judge score improves for \emph{every} model. The four strongest readers~-~GPT-5, Sonnet-4.5, GPT-OSS-120B, and GLM-5.2~-~gain between $+0.10$ and $+0.13$, while Nemotron improves by $+0.03$. The answerable-question refusal rate also falls by $5$--$8$,pp, consistent with fewer queries reaching the reader with distractor or insufficient context. That the same retrieval-only intervention benefits readers with different capabilities strengthens the interpretation that evidence availability is a shared bottleneck across otherwise diverse reader architectures and levels of baseline answer quality. Part of the reader-level degradation under full-corpus RAG is therefore upstream: when document identity is represented during retrieval, more evidence survives the top-$10$ bottleneck, giving all five readers a better opportunity to answer. This shows that chunk representation can materially affect end-to-end RAG performance over a large, heterogeneous corpus.

\begin{table}[t]
\centering
\small
\setlength{\tabcolsep}{4pt}
\resizebox{\linewidth}{!}{
\begin{tabular}{l|c|cc|cc}
\toprule
& & \multicolumn{2}{c|}{\textbf{full@10 plain}}
  & \multicolumn{2}{c}{\textbf{full@10 meta}} \\
\textbf{Model} & \textbf{LC} & \textbf{Judge} & \textbf{Ref\%}
  & \textbf{Judge} & \textbf{Ref\%} \\
\midrule
GPT-5            & 0.757 & 0.473 & 27 & \textbf{0.607} & 20 \\
Sonnet-4.5       & 0.623 & 0.490 & 29 & \textbf{0.587} & 24 \\
GPT-OSS-120B     & 0.701 & 0.497 & 28 & \textbf{0.623} & 19 \\
GLM-5.2          & 0.735 & 0.497 & 30 & \textbf{0.597} & 23 \\
Nemotron-3.5-30B & 0.523 & 0.433 & 23 & \textbf{0.467} & 17 \\
\bottomrule
\end{tabular}
}
\caption{\textbf{Metadata-in-chunk retrieval on \texttt{flat\_full@10}}
(QASPER-MP, EBG, GPT-5 judge, answerable subset, $n{=}200$). ``meta''
prepends the paper title to each chunk before embedding. Judge improves over
plain retrieval for every model (\textbf{bold}); ``Ref\%'' is the
answerable-question refusal rate. LC (full document neighborhood) is shown for
reference. The $95\%$ bootstrap CI half-width on the judge scores is
$\approx$$0.07$.}
\label{tab:qasper_metachunk}
\end{table}

Crucially, \textbf{a stronger retriever narrows but does not close the LC gap}. With metadata-in-chunk, \texttt{flat\_full@10} trails LC on the answerable subset for all models~-~by $0.04$ (Sonnet-4.5), $0.06$ (Nemotron), $0.08$ (GPT-OSS-120B), $0.14$ (GLM-5.2), and $0.15$ (GPT-5) judge points. Roughly half the full-corpus LC advantage is recoverable by fixing retrieval, but a residual remains: even when the right paragraphs are retrieved, removing document context costs answer quality relative to reading the full neighborhood. This points beyond evidence discovery: retrieval preserves passages but discards surrounding structure. That context supplies definitions, antecedents, methodological details, and cross-paragraph relationships needed to interpret evidence. The central finding remains robust to better RAG~-~LC EBG still wins.
\begin{figure*}[h!]
    \centering
    \begin{subfigure}[t]{\textwidth}
        \centering
        \includegraphics[width=\linewidth]{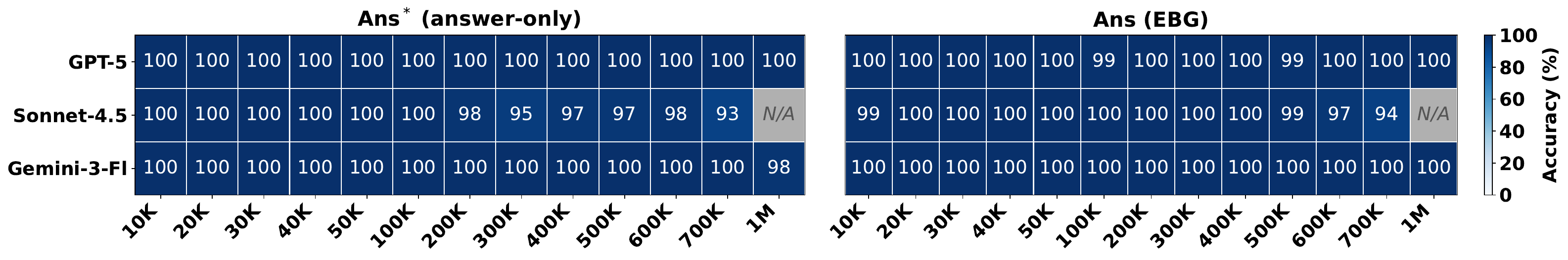}
        \caption{}
        \label{fig:synthetic_ans_vs_ans_only}
    \end{subfigure}
    \\
    \begin{subfigure}[t]{0.6\textwidth}
        \centering
        \includegraphics[width=\linewidth]{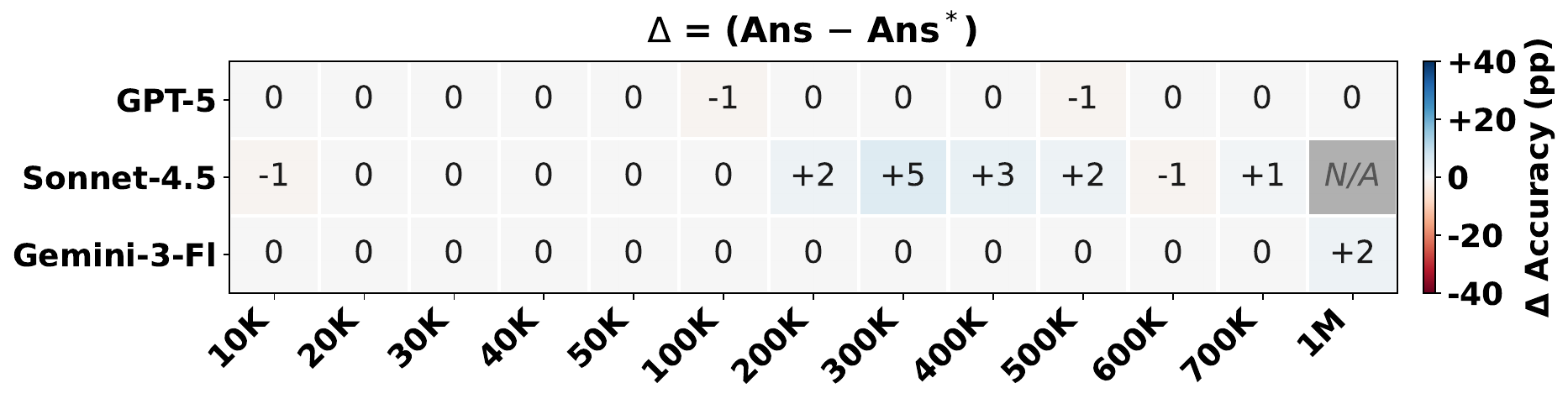}
        \caption{}
        \label{fig:synthetic_delta_ans_star_and_ans}
    \end{subfigure}

    \caption{\textbf{Synthetic two-hop benchmark comparing \texttt{Ans$^{*}$} (answer-only prompting) and \texttt{Ans} (under \textit{EBG} prompting)} for three frontier models across 10K--1M characters. (a) Absolute \texttt{Ans$^{*}$} and \texttt{Ans} accuracies. Higher ($\uparrow$) values are better. (b) Accuracy difference between the two prompting settings. Cells report percent accuracy averaged over $\geq 50$ trials per context length. Sonnet-4.5's 1M entry is marked \textit{N/A} because the input exceeds its 200K-token context window. The two prompting settings produce nearly identical performance on this controlled synthetic benchmark, indicating that the citation requirement neither substantially helps nor hurts answer accuracy. Larger ($\uparrow$) positive difference indicates effectiveness of \textit{EBG} over \textit{answer-only} prompting.}
    \label{fig:synthetic_ans_comparison}
\end{figure*}
\subsection{Summary}

\textsc{QASPER-MP} corroborates the picture in
Appendix~\ref{sec:app_multihop_rag}: under the pool-matched \texttt{pool11} condition,
long-context EBG and RAG end up at comparable aggregate scores, but the
mechanism is asymmetric. RAG's higher refusal rate inflates its score on
unanswerable questions; on the answerable subset~-~where retrieval
quality actually matters~-~LC EBG beats every RAG variant for every one
of the five frontier models we tested. The gap is largest for the
strongest readers (Opus-4.6 and GPT-5.5 each show $+0.13$ judge points
LC over best RAG on answerable questions), suggesting that LC EBG
benefits disproportionately from stronger long-context comprehension.
The gap widens further when the retrieval corpus expands from the
apples-to-apples 11-paper pool to the realistic full-corpus setting,
where LC outperforms the best RAG configuration by 5--13 judge points.

\section{Long-context EBG: Synthetic and NoLiMa-based Datasets}
\label{sec:app_lc_ebg}

This appendix collects the controlled long-context EBG results that support RQ2--RQ4: the synthetic two-hop benchmark (\S~\ref{sec:app_lc_ebg_synthetic}) and the NoLiMa-based benchmark (\S~\ref{sec:app_lc_ebg_nolima}).

\subsection{Ans-only vs EBG on Synthetic Dataset}
\label{sec:app_lc_ebg_synthetic}

\paragraph{Observations.}
Fig.~\ref{fig:synthetic_ans_comparison} compares evidence-based generation (\textbf{EBG}) against \text{answer-only} prompting on the synthetic two-hop benchmark, highlighting both the advantages and limitations of requiring explicit evidence attribution.
\begin{itemize}[leftmargin=1em, itemsep=0pt, topsep=0pt, parsep=0pt, partopsep=0pt]
    \item \textbf{Performance rapidly saturates for all frontier models.}  
    Across nearly the entire 10K--1M character range, all models operate close to ceiling performance, with accuracies clustering near 100\% under both prompt settings. This indicates that the synthetic two-hop benchmark is largely solved by current frontier systems once the relevant evidence is successfully retrieved, and that increasing context length alone does not substantially reduce answer accuracy for the strongest models.

    \item \textbf{Answer-only and EBG prompting achieve nearly identical performance across all context lengths.}  
    For \texttt{GPT-5} and \texttt{Gemini-3-Fl}, \texttt{Ans}$^{*}$ (answer-only prompting) and \texttt{Ans} (answer with evidence prompting) remain effectively indistinguishable from 10K to 1M characters, with almost all cells at or near 100\% accuracy. The difference heatmap confirms that the vast majority of entries have zero difference, indicating that adding citation requirements does not materially affect answer correctness on this synthetic benchmark. 

\end{itemize}

\begin{table*}[h!]
\centering

\begin{subtable}{\textwidth}
\centering
\resizebox{0.9\linewidth}{!}{
\begin{tabular}{l*{6}{ccc}}
\toprule
& \multicolumn{3}{c}{10K} & \multicolumn{3}{c}{20K} & \multicolumn{3}{c}{30K}
& \multicolumn{3}{c}{40K} & \multicolumn{3}{c}{50K} & \multicolumn{3}{c}{100K} \\
\cmidrule(lr){2-4}\cmidrule(lr){5-7}\cmidrule(lr){8-10}
\cmidrule(lr){11-13}\cmidrule(lr){14-16}\cmidrule(lr){17-19}
Model
& Ans* & Ans & $\Delta$
& Ans* & Ans & $\Delta$
& Ans* & Ans & $\Delta$
& Ans* & Ans & $\Delta$
& Ans* & Ans & $\Delta$
& Ans* & Ans & $\Delta$ \\
\midrule

\rowcolor{gray!25}\multicolumn{19}{l}{\textit{One-hop}} \\
GPT-4o & 100.00 & 75.00 & \dneg{-25.00} & 100.00 & 87.50 & \dneg{-12.50} & 93.75 & 81.25 & \dneg{-12.50} & 81.25 & 87.50 & \dpos{6.25} & 100.00 & 68.75 & \dneg{-31.25} & 100.00 & 87.50 & \dneg{-12.50} \\
GPT-4.1 & 100.00 & 100.00 & \dzero & 100.00 & 100.00 & \dzero & 100.00 & 93.75 & \dneg{-6.25} & 100.00 & 100.00 & \dzero & 100.00 & 100.00 & \dzero & 100.00 & 100.00 & \dzero \\
O3-mini & 87.50 & 81.25 & \dneg{-6.25} & 87.50 & 81.25 & \dneg{-6.25} & 87.50 & 87.50 & \dzero & 87.50 & 87.50 & \dzero & 87.50 & 81.25 & \dneg{-6.25} & 68.75 & 75.00 & \dpos{6.25} \\
GPT-5 & 100.00 & 100.00 & \dzero & 100.00 & 100.00 & \dzero & 100.00 & 100.00 & \dzero & 100.00 & 100.00 & \dzero & 100.00 & 100.00 & \dzero & 100.00 & 100.00 & \dzero \\
Sonnet-4 & 100.00 & 100.00 & \dzero & 100.00 & 100.00 & \dzero & 93.75 & 100.00 & \dpos{6.25} & 100.00 & 100.00 & \dzero & 100.00 & 100.00 & \dzero & 100.00 & 93.75 & \dneg{-6.25} \\
Sonnet-4.5 & 100.00 & 100.00 & \dzero & 100.00 & 100.00 & \dzero & 87.50 & 100.00 & \dpos{12.50} & 100.00 & 100.00 & \dzero & 93.75 & 100.00 & \dpos{6.25} & 100.00 & 93.75 & \dneg{-6.25} \\
Gemini-2.5-Fl & 100.00 & 100.00 & \dzero & 100.00 & 100.00 & \dzero & 100.00 & 100.00 & \dzero & 100.00 & 100.00 & \dzero & 100.00 & 100.00 & \dzero & 100.00 & 100.00 & \dzero \\
Gemini-3-Fl & 100.00 & 100.00 & \dzero & 100.00 & 100.00 & \dzero & 100.00 & 100.00 & \dzero & 100.00 & 100.00 & \dzero & 100.00 & 100.00 & \dzero & 100.00 & 100.00 & \dzero \\
Nemo-3.5-L & 75.00 & 93.75 & \dpos{18.75} & 56.25 & 56.25 & \dzero & 62.50 & 62.50 & \dzero & 62.50 & 50.00 & \dneg{-12.50} & 56.25 & 50.00 & \dneg{-6.25} & 56.25 & 43.75 & \dneg{-12.50} \\
GLM-5.2 & 100.00 & 100.00 & \dzero & 100.00 & 100.00 & \dzero & 100.00 & 100.00 & \dzero & 100.00 & 93.75 & \dneg{-6.25} & 100.00 & 100.00 & \dzero & 87.50 & 100.00 & \dpos{12.50} \\
GPT-OSS-120B & 75.00 & 75.00 & \dzero & 75.00 & 37.50 & \dneg{-37.50} & 81.25 & 68.75 & \dneg{-12.50} & 62.50 & 62.50 & \dzero & 75.00 & 68.75 & \dneg{-6.25} & 93.75 & 81.25 & \dneg{-12.50} \\

\midrule
\rowcolor{gray!25}\multicolumn{19}{l}{\textit{Two-hop}} \\
GPT-4o & 87.50 & 68.75 & \dneg{-18.75} & 93.75 & 81.25 & \dneg{-12.50} & 87.50 & 87.50 & \dzero & 75.00 & 75.00 & \dzero & 87.50 & 68.75 & \dneg{-18.75} & 87.50 & 56.25 & \dneg{-31.25} \\
GPT-4.1 & 100.00 & 100.00 & \dzero & 100.00 & 93.75 & \dneg{-6.25} & 100.00 & 93.75 & \dneg{-6.25} & 100.00 & 100.00 & \dzero & 100.00 & 100.00 & \dzero & 100.00 & 93.75 & \dneg{-6.25} \\
O3-mini & 68.75 & 68.75 & \dzero & 68.75 & 68.75 & \dzero & 62.50 & 62.50 & \dzero & 62.50 & 56.25 & \dneg{-6.25} & 68.75 & 68.75 & \dzero & 37.50 & 50.00 & \dpos{12.50} \\
GPT-5 & 100.00 & 87.50 & \dneg{-12.50} & 100.00 & 93.75 & \dneg{-6.25} & 100.00 & 93.75 & \dneg{-6.25} & 100.00 & 93.75 & \dneg{-6.25} & 100.00 & 93.75 & \dneg{-6.25} & 100.00 & 87.50 & \dneg{-12.50} \\
Sonnet-4 & 68.75 & 87.50 & \dpos{18.75} & 81.25 & 87.50 & \dpos{6.25} & 68.75 & 81.25 & \dpos{12.50} & 68.75 & 93.75 & \dpos{25.00} & 81.25 & 81.25 & \dzero & 62.50 & 87.50 & \dpos{25.00} \\
Sonnet-4.5 & 87.50 & 100.00 & \dpos{12.50} & 100.00 & 100.00 & \dzero & 93.75 & 93.75 & \dzero & 100.00 & 93.75 & \dneg{-6.25} & 93.75 & 100.00 & \dpos{6.25} & 81.25 & 87.50 & \dpos{6.25} \\
Gemini-2.5-Fl & 100.00 & 93.75 & \dneg{-6.25} & 93.75 & 81.25 & \dneg{-12.50} & 68.75 & 87.50 & \dpos{18.75} & 62.50 & 75.00 & \dpos{12.50} & 68.75 & 62.50 & \dneg{-6.25} & 56.25 & 50.00 & \dneg{-6.25} \\
Gemini-3-Fl & 68.75 & 68.75 & \dzero & 56.25 & 75.00 & \dpos{18.75} & 62.50 & 81.25 & \dpos{18.75} & 81.25 & 93.75 & \dpos{12.50} & 93.75 & 93.75 & \dzero & 81.25 & 93.75 & \dpos{12.50} \\
Nemo-3.5-L & 37.50 & 43.75 & \dpos{6.25} & 25.00 & 31.25 & \dpos{6.25} & 18.75 & 31.25 & \dpos{12.50} & 25.00 & 18.75 & \dneg{-6.25} & 25.00 & 18.75 & \dneg{-6.25} & 12.50 & 18.75 & \dpos{6.25} \\
GLM-5.2 & 62.50 & 50.00 & \dneg{-12.50} & 75.00 & 81.25 & \dpos{6.25} & 68.75 & 62.50 & \dneg{-6.25} & 62.50 & 50.00 & \dneg{-12.50} & 50.00 & 50.00 & \dzero & 50.00 & 31.25 & \dneg{-18.75} \\
GPT-OSS-120B & 50.00 & 6.25 & \dneg{-43.75} & 43.75 & 18.75 & \dneg{-25.00} & 31.25 & 0.00 & \dneg{-31.25} & 37.50 & 6.25 & \dneg{-31.25} & 31.25 & 12.50 & \dneg{-18.75} & 25.00 & 12.50 & \dneg{-12.50} \\
\bottomrule
\end{tabular}
}
\caption{Performance from 10K to 100K.}
\end{subtable}

% \vspace{0.75em}

\begin{subtable}{\textwidth}
\centering
\resizebox{0.9\linewidth}{!}{
\begin{tabular}{l*{6}{ccc}}
\toprule
& \multicolumn{3}{c}{200K} & \multicolumn{3}{c}{300K} & \multicolumn{3}{c}{400K}
& \multicolumn{3}{c}{500K} & \multicolumn{3}{c}{600K} & \multicolumn{3}{c}{700K} \\
\cmidrule(lr){2-4}\cmidrule(lr){5-7}\cmidrule(lr){8-10}
\cmidrule(lr){11-13}\cmidrule(lr){14-16}\cmidrule(lr){17-19}
Model
& Ans* & Ans & $\Delta$
& Ans* & Ans & $\Delta$
& Ans* & Ans & $\Delta$
& Ans* & Ans & $\Delta$
& Ans* & Ans & $\Delta$
& Ans* & Ans & $\Delta$ \\
\midrule

\rowcolor{gray!25}\multicolumn{19}{l}{\textit{One-hop}} \\
GPT-4o & 93.75 & 62.50 & \dneg{-31.25} & 87.50 & 68.75 & \dneg{-18.75} & 87.50 & 56.25 & \dneg{-31.25} & 93.75 & 62.50 & \dneg{-31.25} & \nacell & \nacell & \nacell & \nacell & \nacell & \nacell \\
GPT-4.1 & 100.00 & 100.00 & \dzero & 100.00 & 100.00 & \dzero & 100.00 & 87.50 & \dneg{-12.50} & 100.00 & 93.75 & \dneg{-6.25} & 87.50 & 93.75 & \dpos{6.25} & 93.75 & 93.75 & \dzero \\
O3-mini & 87.50 & 75.00 & \dneg{-12.50} & 56.25 & 43.75 & \dneg{-12.50} & 62.50 & 37.50 & \dneg{-25.00} & 43.75 & 31.25 & \dneg{-12.50} & 37.50 & 25.00 & \dneg{-12.50} & 31.25 & 18.75 & \dneg{-12.50} \\
GPT-5 & 100.00 & 100.00 & \dzero & 100.00 & 100.00 & \dzero & 100.00 & 100.00 & \dzero & 100.00 & 100.00 & \dzero & 100.00 & 100.00 & \dzero & 100.00 & 100.00 & \dzero \\
Sonnet-4 & 100.00 & 93.75 & \dneg{-6.25} & 87.50 & 93.75 & \dpos{6.25} & 81.25 & 93.75 & \dpos{12.50} & 68.75 & 87.50 & \dpos{18.75} & 87.50 & 87.50 & \dzero & 75.00 & 81.25 & \dpos{6.25} \\
Sonnet-4.5 & 100.00 & 100.00 & \dzero & 81.25 & 87.50 & \dpos{6.25} & 75.00 & 87.50 & \dpos{12.50} & 75.00 & 75.00 & \dzero & 75.00 & 87.50 & \dpos{12.50} & 56.25 & 75.00 & \dpos{18.75} \\
Gemini-2.5-Fl & 100.00 & 100.00 & \dzero & 100.00 & 100.00 & \dzero & 93.75 & 100.00 & \dpos{6.25} & 87.50 & 100.00 & \dpos{12.50} & 100.00 & 87.50 & \dneg{-12.50} & 100.00 & 87.50 & \dneg{-12.50} \\
Gemini-3-Fl & 100.00 & 100.00 & \dzero & 100.00 & 100.00 & \dzero & 100.00 & 100.00 & \dzero & 100.00 & 100.00 & \dzero & 100.00 & 100.00 & \dzero & 100.00 & 100.00 & \dzero \\
Nemo-3.5-L & 56.25 & 56.25 & \dzero & 50.00 & 50.00 & \dzero & 25.00 & 12.50 & \dneg{-12.50} & 25.00 & 43.75 & \dpos{18.75} & 31.25 & 37.50 & \dpos{6.25} & 12.50 & 25.00 & \dpos{12.50} \\
GLM-5.2 & 87.50 & 93.75 & \dpos{6.25} & 87.50 & 87.50 & \dzero & 81.25 & 75.00 & \dneg{-6.25} & 87.50 & 81.25 & \dneg{-6.25} & 68.75 & 62.50 & \dneg{-6.25} & 75.00 & 93.75 & \dpos{18.75} \\
GPT-OSS-120B & 75.00 & 68.75 & \dneg{-6.25} & 81.25 & 75.00 & \dneg{-6.25} & 56.25 & 50.00 & \dneg{-6.25} & \nacell & \nacell & \nacell & \nacell & \nacell & \nacell & \nacell & \nacell & \nacell \\

\midrule
\rowcolor{gray!25}\multicolumn{19}{l}{\textit{Two-hop}} \\
GPT-4o & 87.50 & 43.75 & \dneg{-43.75} & 75.00 & 31.25 & \dneg{-43.75} & 75.00 & 43.75 & \dneg{-31.25} & 87.50 & 62.50 & \dneg{-25.00} & \nacell & \nacell & \nacell & \nacell & \nacell & \nacell \\
GPT-4.1 & 93.75 & 100.00 & \dpos{6.25} & 100.00 & 100.00 & \dzero & 100.00 & 87.50 & \dneg{-12.50} & 87.50 & 81.25 & \dneg{-6.25} & 81.25 & 81.25 & \dzero & 87.50 & 87.50 & \dzero \\
O3-mini & 68.75 & 56.25 & \dneg{-12.50} & 31.25 & 37.50 & \dpos{6.25} & 37.50 & 43.75 & \dpos{6.25} & 18.75 & 37.50 & \dpos{18.75} & 18.75 & 25.00 & \dpos{6.25} & 12.50 & 18.75 & \dpos{6.25} \\
GPT-5 & 100.00 & 100.00 & \dzero & 100.00 & 100.00 & \dzero & 100.00 & 93.75 & \dneg{-6.25} & 100.00 & 93.75 & \dneg{-6.25} & 93.75 & 93.75 & \dzero & 93.75 & 100.00 & \dpos{6.25} \\
Sonnet-4 & 43.75 & 68.75 & \dpos{25.00} & 43.75 & 62.50 & \dpos{18.75} & 50.00 & 56.25 & \dpos{6.25} & 37.50 & 56.25 & \dpos{18.75} & 68.75 & 81.25 & \dpos{12.50} & 37.50 & 50.00 & \dpos{12.50} \\
Sonnet-4.5 & 50.00 & 62.50 & \dpos{12.50} & 43.75 & 62.50 & \dpos{18.75} & 50.00 & 75.00 & \dpos{25.00} & 37.50 & 50.00 & \dpos{12.50} & 68.75 & 75.00 & \dpos{6.25} & 50.00 & 50.00 & \dzero \\
Gemini-2.5-Fl & 62.50 & 56.25 & \dneg{-6.25} & 50.00 & 56.25 & \dpos{6.25} & 50.00 & 75.00 & \dpos{25.00} & 37.50 & 37.50 & \dzero & 31.25 & 25.00 & \dneg{-6.25} & 31.25 & 25.00 & \dneg{-6.25} \\
Gemini-3-Fl & 81.25 & 87.50 & \dpos{6.25} & 87.50 & 93.75 & \dpos{6.25} & 62.50 & 87.50 & \dpos{25.00} & 68.75 & 100.00 & \dpos{31.25} & 81.25 & 93.75 & \dpos{12.50} & 75.00 & 87.50 & \dpos{12.50} \\
Nemo-3.5-L & 37.50 & 18.75 & \dneg{-18.75} & 0.00 & 6.25 & \dpos{6.25} & 6.25 & 6.25 & \dzero & 6.25 & 6.25 & \dzero & 0.00 & 0.00 & \dzero & 0.00 & 0.00 & \dzero \\
GLM-5.2 & 68.75 & 56.25 & \dneg{-12.50} & 56.25 & 50.00 & \dneg{-6.25} & 56.25 & 50.00 & \dneg{-6.25} & 43.75 & 50.00 & \dpos{6.25} & 50.00 & 43.75 & \dneg{-6.25} & 43.75 & 56.25 & \dpos{12.50} \\
GPT-OSS-120B & 37.50 & 25.00 & \dneg{-12.50} & 43.75 & 25.00 & \dneg{-18.75} & 18.75 & 18.75 & \dzero & \nacell & \nacell & \nacell & \nacell & \nacell & \nacell & \nacell & \nacell & \nacell \\
\bottomrule
\end{tabular}
}
\caption{Performance from 200K to 700K.}
\end{subtable}

% \vspace{0.75em}

\begin{subtable}{\textwidth}
\centering
\resizebox{0.52\linewidth}{!}{
\begin{tabular}{l*{3}{ccc}}
\toprule
& \multicolumn{3}{c}{800K} & \multicolumn{3}{c}{900K} & \multicolumn{3}{c}{1M} \\
\cmidrule(lr){2-4}\cmidrule(lr){5-7}\cmidrule(lr){8-10}
Model
& Ans* & Ans & $\Delta$
& Ans* & Ans & $\Delta$
& Ans* & Ans & $\Delta$ \\
\midrule

\rowcolor{gray!25}\multicolumn{10}{l}{\textit{One-hop}} \\
GPT-4o & \nacell & \nacell & \nacell & \nacell & \nacell & \nacell & \nacell & \nacell & \nacell \\
GPT-4.1 & 100.00 & 81.25 & \dneg{-18.75} & 87.50 & 75.00 & \dneg{-12.50} & 87.50 & 81.25 & \dneg{-6.25} \\
O3-mini & \nacell & \nacell & \nacell & \nacell & \nacell & \nacell & \nacell & \nacell & \nacell \\
GPT-5 & 100.00 & 100.00 & \dzero & 100.00 & 100.00 & \dzero & 100.00 & 100.00 & \dzero \\
Sonnet-4 & \nacell & \nacell & \nacell & \nacell & \nacell & \nacell & \nacell & \nacell & \nacell \\
Sonnet-4.5 & \nacell & \nacell & \nacell & \nacell & \nacell & \nacell & \nacell & \nacell & \nacell \\
Gemini-2.5-Fl & 87.50 & 93.75 & \dpos{6.25} & 93.75 & 75.00 & \dneg{-18.75} & 100.00 & 93.75 & \dneg{-6.25} \\
Gemini-3-Fl & 100.00 & 100.00 & \dzero & 100.00 & 100.00 & \dzero & 100.00 & 100.00 & \dzero \\
Nemo-3.5-L & 12.50 & 12.50 & \dzero & 25.00 & 25.00 & \dzero & \nacell & \nacell & \nacell \\
GLM-5.2 & 81.25 & 81.25 & \dzero & 87.50 & 81.25 & \dneg{-6.25} & 50.00 & 43.75 & \dneg{-6.25} \\
GPT-OSS-120B & \nacell & \nacell & \nacell & \nacell & \nacell & \nacell & \nacell & \nacell & \nacell \\

\midrule
\rowcolor{gray!25}\multicolumn{10}{l}{\textit{Two-hop}} \\
GPT-4o & \nacell & \nacell & \nacell & \nacell & \nacell & \nacell & \nacell & \nacell & \nacell \\
GPT-4.1 & 93.75 & 81.25 & \dneg{-12.50} & 81.25 & 62.50 & \dneg{-18.75} & 75.00 & 68.75 & \dneg{-6.25} \\
O3-mini & \nacell & \nacell & \nacell & \nacell & \nacell & \nacell & \nacell & \nacell & \nacell \\
GPT-5 & 100.00 & 87.50 & \dneg{-12.50} & 87.50 & 81.25 & \dneg{-6.25} & 93.75 & 87.50 & \dneg{-6.25} \\
Sonnet-4 & \nacell & \nacell & \nacell & \nacell & \nacell & \nacell & \nacell & \nacell & \nacell \\
Sonnet-4.5 & \nacell & \nacell & \nacell & \nacell & \nacell & \nacell & \nacell & \nacell & \nacell \\
Gemini-2.5-Fl & 43.75 & 37.50 & \dneg{-6.25} & 37.50 & 18.75 & \dneg{-18.75} & 37.50 & 31.25 & \dneg{-6.25} \\
Gemini-3-Fl & 56.25 & 81.25 & \dpos{25.00} & 75.00 & 93.75 & \dpos{18.75} & 81.25 & 93.75 & \dpos{12.50} \\
Nemo-3.5-L & 0.00 & 0.00 & \dzero & 12.50 & 12.50 & \dzero & \nacell & \nacell & \nacell \\
GLM-5.2 & 43.75 & 62.50 & \dpos{18.75} & 43.75 & 31.25 & \dneg{-12.50} & 25.00 & 25.00 & \dzero \\
GPT-OSS-120B & \nacell & \nacell & \nacell & \nacell & \nacell & \nacell & \nacell & \nacell & \nacell \\
\bottomrule
\end{tabular}
}
\caption{Performance from 800K to 1M.}
\end{subtable}

\caption{Performance of models across different context lengths, for one-hop and two-hop questions, averaged across $n=16$ (4 needles $\times$ 4 depths) cases, \textbf{comparing \texttt{Ans}$^*$ 
(answer-only) with \texttt{Ans} (EBG)}. $\Delta$ denotes the difference $\texttt{Ans} - \texttt{Ans}^*$; positive values are shown in bold \textbf{\textcolor{blue}{blue}}, negative values in bold \textbf{\textcolor{red}{red}}. Higher ($\uparrow$) values are better.
Positive values indicate that \textit{EBG} prompting outperforms 
\textit{answer-only} prompting, while negative values indicate that \textit{answer-only} 
prompting improves over \textit{EBG} prompting. Zero indicates no 
difference between the two prompting strategies. ``N/A'' denotes unavailable 
entries due to unsupported context lengths. Here we present scores correct upto two decimal places.}
\label{tab:scales}
\end{table*}

  \begin{figure*}[t]
    \centering
    \includegraphics[width=\textwidth]{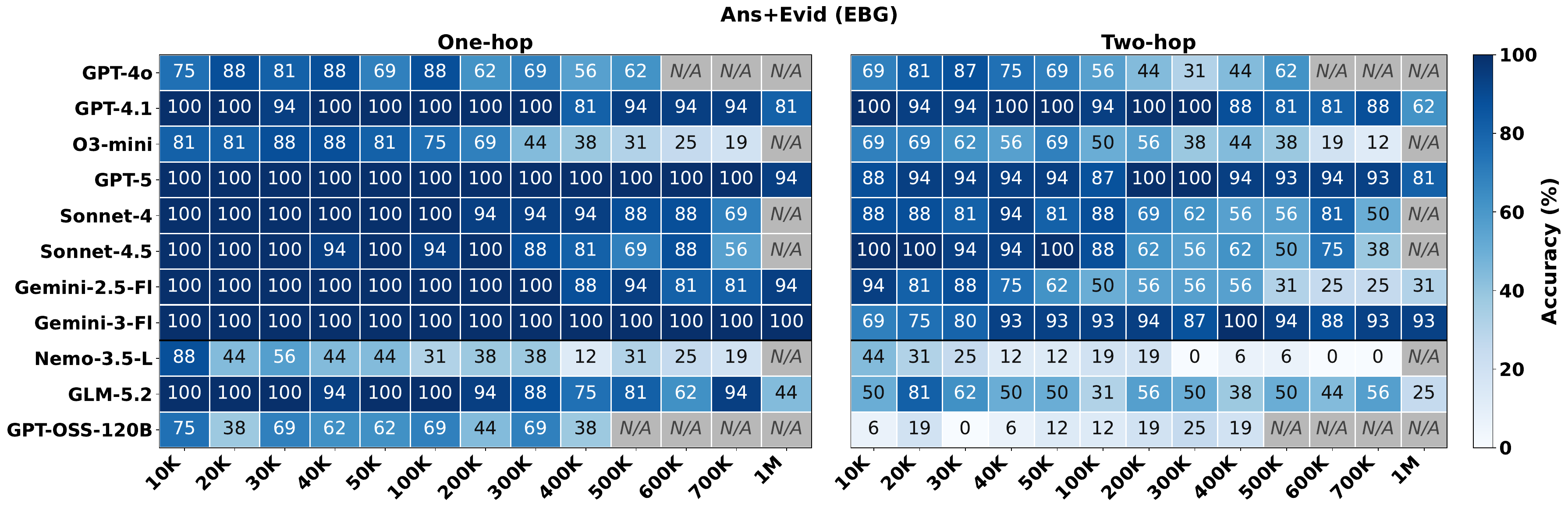}
    \caption{Performance of different models on \textbf{Ans+Evid (EBG)} accuracy for \textbf{one-hop and two-hop} questions across varying context lengths, averaged over needle depths. Cells marked with ``N/A'' indicate no available results for that model-context combination. Higher ($\uparrow$) values are better).
    }
    \label{fig:ans-evid-acc_one-hop_vs_two-hop}
\end{figure*}

\subsection{Ans-only vs EBG on NoLiMa-based Dataset}
\label{sec:app_lc_ebg_nolima}

\paragraph{Observations.}As shown in Table~\ref{tab:scales}, \textit{EBG} prompting is more beneficial for multi-hop queries than for simple one-hop queries. In one-hop settings, many strong models already perform near ceiling under answer-only prompting, leaving little room for explicit evidence organization to help. In some cases, the additional structure introduced by \textit{EBG} even hurts performance, suggesting that it may interfere with models that already perform strong implicit retrieval or that are sensitive to extra reasoning overhead.
The benefits of \textit{EBG} become clearer in the two-hop setting, where models must retrieve and combine evidence across multiple pieces of context. The best long-context performers are \texttt{GPT-5} and \texttt{Gemini-3-Fl}, but for different reasons. \texttt{GPT-5} is highly robust regardless of prompting strategy, while \texttt{Gemini-3-Fl} gains substantially from \textit{EBG} in two-hop settings.

The strongest evidence in favor of \textit{EBG} comes from the \texttt{Gemini-3-Fl} two-hop results, where \textit{EBG} improves performance consistently across almost all context lengths, including the 1M setting. The \texttt{Sonnet} models also show frequent gains from \textit{EBG} in two-hop tasks, further supporting the view that explicit evidence organization is most useful when reasoning requires multiple retrieval steps. In contrast, the strongest evidence against universal \textit{EBG} use comes from \texttt{GPT-4o}, where \textit{EBG} substantially lowers performance in both one-hop and two-hop tasks.

Overall, the table suggests that \textit{EBG} is especially valuable for long-context multi-hop reasoning in models, but it may be unnecessary for simple one-hop retrieval.

\subsection{Strict EBG on NoLiMa-based Dataset}

\paragraph{Observations.}
Fig. \ref{fig:ans-evid-acc_one-hop_vs_two-hop} reports the combined \texttt{Ans+Evid} accuracy across models, context lengths, and reasoning types. Since this metric requires both the final answer and the supporting evidence to be correct, it is a \textbf{stricter} measure than answer accuracy alone. The results show clear differences in how well models remain both accurate and grounded as context length increases.

\paragraph{One-hop performance.}
In the one-hop setting, the reported joint \texttt{Ans+Evid} metric measures answer correctness
together with citation of the primary contextual anchor, rather than
a complete context-only reasoning chain, because for one-hop, part of the inference relies on \emph{bridge
knowledge} not stated in the haystack (such as \emph{veganism} implying avoidance
of \emph{fish} or \emph{lactose} intolerance implying avoidance of \emph{milk}. Here, we observe that most frontier models perform very strongly at short and medium context lengths. \texttt{GPT-5}, \texttt{Sonnet-4}, \texttt{Sonnet-4.5}, \texttt{Gemini-2.5-Fl}, and \texttt{Gemini-3-Fl} frequently achieve near-perfect or perfect accuracy across many context lengths. This suggests that when only one relevant piece of evidence needs to be retrieved, strong long-context models can usually identify the evidence and produce the correct answer reliably.

\texttt{Gemini-3-Fl} is the strongest one-hop model overall, maintaining 100\% accuracy across all evaluated context lengths, including 1M characters. \texttt{GPT-5} is also highly robust, achieving 100\% accuracy across most context lengths and only dropping slightly to 94\% at 1M. \texttt{Gemini-2.5-Fl} also remains strong, although it declines at longer contexts, reaching 81\% at 600K and 700K before recovering to 94\% at 1M.

\texttt{GPT-4.1} performs well in one-hop reasoning, but its performance decreases at the largest context length, reaching 81\% at 1M. \texttt{GPT-4o} and \texttt{o3-mini} show much weaker robustness. \texttt{GPT-4o} drops substantially after 100K characters, while \texttt{o3-mini} degrades sharply beyond 200K characters, falling to 31\% at 500K and 19\% at 700K.

\paragraph{Two-hop performance.}
The two-hop setting is significantly harder. Even models that perform very well in one-hop reasoning show larger performance drops when they must retrieve and combine multiple pieces of evidence. This indicates that long-context multi-hop reasoning requires more than simply locating one relevant span; the model must maintain multiple pieces of information and reason over their relationship.

\texttt{Gemini-3-Fl} is the strongest model in the two-hop setting. It maintains high accuracy across the context range and achieves 93\% at 1M characters. \texttt{GPT-5} is strong from 10K to 700K, but drops to 81\% at 1M. \texttt{GPT-4.1} performs well through mid-range contexts but declines at 1M, reaching 62\%.

\subsection{TN vs.\ CTN on NoLiMa-based Dataset: Adherence to Contextual Evidence}
\label{app:tn_vs_ctn}

We conduct an experiment on the NoLiMa-based dataset to evaluate whether LLMs remain faithful to contextual evidence in two-hop reasoning questions when presented with either factual secondary evidence that aligns with world knowledge (\textit{TN}) or counterfactual secondary evidence that conflicts with world knowledge (\textit{CTN}; e.g., ``black coffee has milk'').

\paragraph{Experiment.}
Under TN and CTN, each query is associated with two needles. We insert the primary needle at the middle of the haystack (depth $0.5$) and the secondary needle at one of the sampled depths $\{0.0, 0.33, 0.67, 1.0\}$.
Since questions must be answered strictly from the provided context, success in CTN requires relying on the contradictory contextual evidence rather than parametric memory. We evaluate the performance of TN and CTN using \texttt{Ans} and \texttt{Ans+Evid} accuracy metrics.

\begin{figure}[t]
\centering
\includegraphics[width=\linewidth]{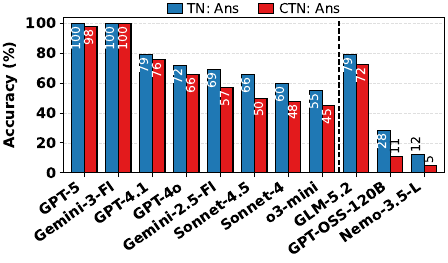}
\caption{\textbf{TN vs.\ CTN} \texttt{Ans accuracy under EBG prompting}. We report an accuracy for the same runs whose \texttt{Ans+Evid} accuracy appears in Figure~\ref{fig:ctn_two_needle}. Higher ($\uparrow$) values indicate better performance.}
\label{fig:ctn_two_needle_ans}
\end{figure}

\paragraph{Observations.}
\begin{itemize}[leftmargin=1em, itemsep=0pt, topsep=0pt, parsep=0pt, partopsep=0pt]
    \item \textbf{Contradiction reliably hurts performance.} Averaged across models, CTN reduces \texttt{Ans} accuracy by $\approx$\textbf{7.0\,pp} (76.2\% $\to$ 69.2\%) and \texttt{Ans+Evid} by $\approx$\textbf{5.9\,pp} (49.9\% $\to$ 44.0\%).
    \item \textbf{The Ans+Evid accuracy metric is much harder than Ans.} \texttt{Ans+Evid} is far below \texttt{Ans} in both \textit{TN} and \textit{CTN}, indicating many correct-answer cases still fail to produce a complete evidence trail.
    \item \textbf{Robustness varies sharply by model.} \texttt{Gemini-3-Fl} and \texttt{GPT-5} remain most stable; the largest \texttt{Ans+Evid} drops under \textit{CTN} occur for \texttt{GPT-4.1} ($\approx$\textbf{-13\,pp}), \texttt{Gemini-2.5-Fl} ($\approx$\textbf{-10\,pp}), and \texttt{o3-mini} ($\approx$\textbf{-8.6\,pp}).
    \item \textbf{Depth sensitivity persists under contradiction,} with the CTN penalty tending to be larger when the secondary needle is deeper, consistent with degraded multi-evidence retrieval under harder placements, i.e., mostly at depths 0.33 and 0.67.
    \item \textbf{CTN remains below TN across context lengths}, indicating contradiction amplifies long-context brittleness as both conditions degrade with length.
    \item \textbf{Contradiction weakens evidence grounding, not just answers.} CTN lowers citation rates for both needles (primary $\approx$\textbf{-7.4\,pp} and  secondary $\approx$\textbf{-4.6\,pp}), pointing to provenance failure as a primary failure mode under conflicting information.
    \item \textbf{The answer dimension degrades too, but less than the joint metric.} Figure~\ref{fig:ctn_two_needle_ans} reports \texttt{Ans} accuracy for the same runs. Every model except \texttt{Gemini-3-Fl} loses answer accuracy under contradiction, with the largest drops for \texttt{GPT-OSS-120B} ($\approx$\textbf{-17\,pp}), \texttt{Sonnet-4.5} ($\approx$\textbf{-16\,pp}), and \texttt{Gemini-2.5-Fl} and \texttt{Sonnet-4} ($\approx$\textbf{-12\,pp} each).
\end{itemize}

% \section{Effect of Chain-of-Thought Prompting}
% \label{app:cot_prompting}

% We conduct two prompting ablations to probe whether the gains associated with
% evidence-based generation (\textit{EBG}) arise from evidence citation alone or
% from the reasoning structure induced by the prompt. Both experiments use the
% NoLiMa-based \textit{LC} benchmark with Book~1, the \texttt{T01} test variant,
% four needle families, four insertion depths, and context lengths ranging from
% 10K to 1M characters. For each comparison, results are paired by model,
% context length, test, insertion depth, and character before aggregation.
% Each model--context--hop cell therefore contains up to 16 matched instances
% (4 tests $\times$ 4 depths); API errors are excluded.

% The first condition, \textit{CoT}, requests a brief explanation of no more than
% three sentences followed by the final answer on a separate line. It does not
% request evidence citations, so only answer accuracy is comparable with EBG.
% The second condition, \textit{CoT-EBG}, adds the same brief reasoning
% instruction while preserving the EBG JSON schema containing the answer and
% supporting line numbers. This condition can therefore be evaluated using
% answer accuracy (\texttt{Ans}), evidence accuracy (\texttt{Evid}), and their
% instance-level conjunction (\texttt{Ans+Evid}).

\subsection{EBG vs.\ CoT Prompting on NoLiMa-based Dataset}
\label{app:ebg_vs_cot}

\paragraph{Experiment.}
We compare answer accuracy under standard EBG and CoT prompting separately for one-hop and two-hop questions. We define $\Delta = \texttt{Ans} - \texttt{Ans}^{c}$, where \texttt{Ans} denotes answer accuracy under EBG prompting and $\texttt{Ans}^{c}$ denotes answer accuracy under CoT prompting. Positive values of $\Delta$ therefore indicate an advantage for EBG, whereas negative values indicate an advantage for CoT.

\begin{figure*}[t]
\centering
\subfloat[]{
\includegraphics[width=\linewidth]
{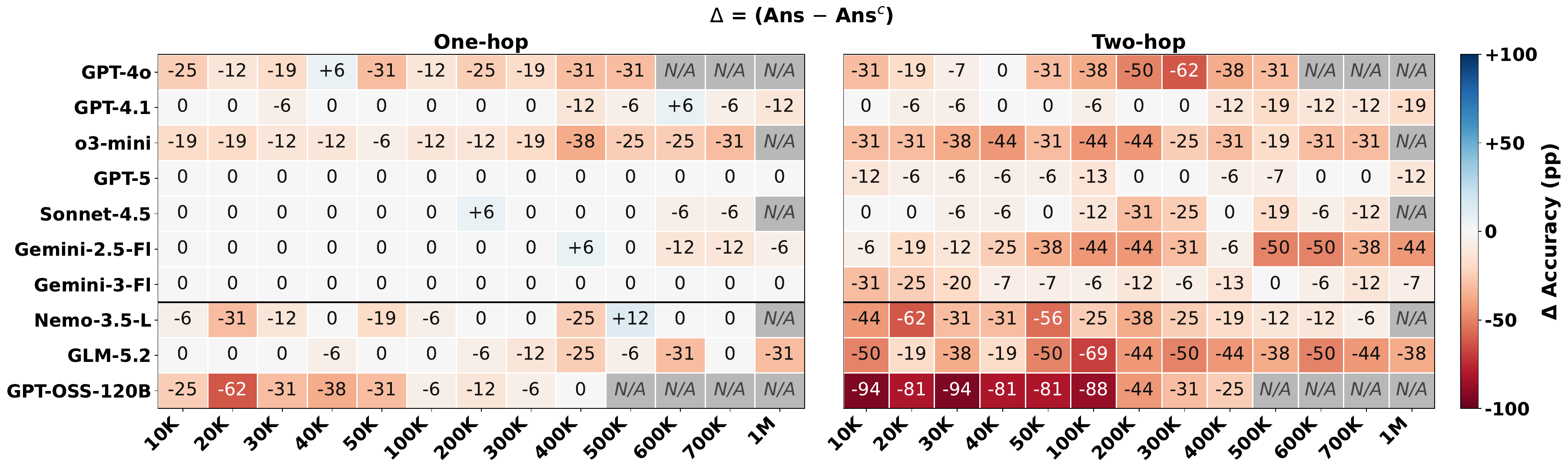}
\label{fig:ebg_vs_cot_delta_ans}
}
\hfill
\subfloat[]{
\includegraphics[width=\linewidth]
{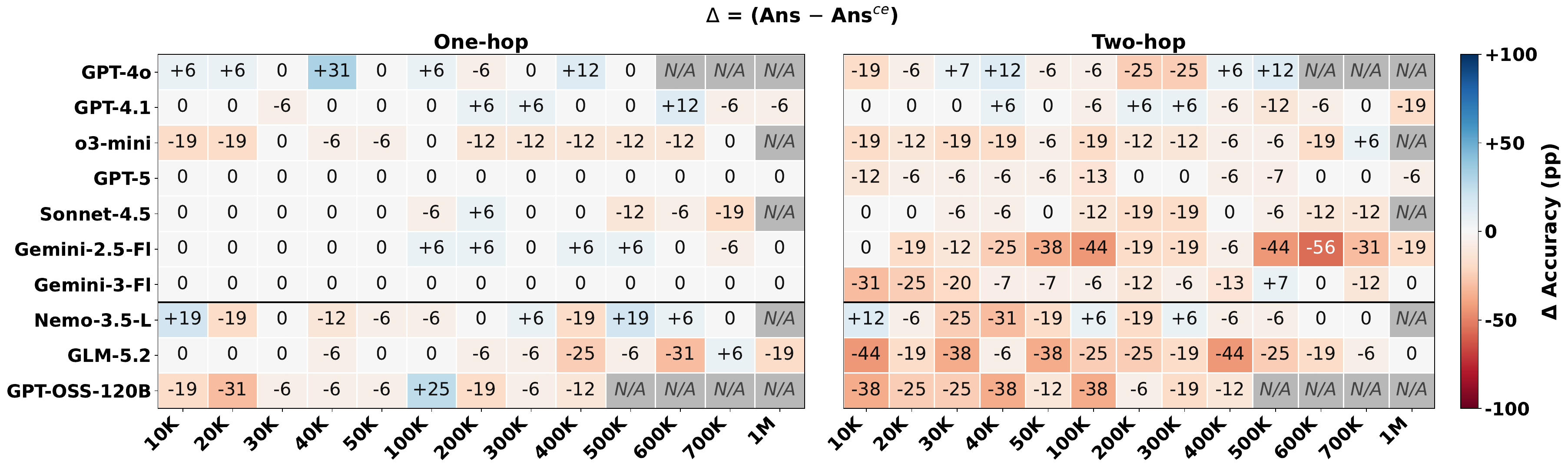}
\label{fig:ebg_vs_cot_ebg_delta_ans}
}
\caption{\textbf{Comparison of answer-accuracy differences between EBG vs.\ CoT and EBG vs.\ CoT-EBG prompting} on the NoLiMa-based dataset, with results reported separately for one-hop and two-hop questions.
 \textbf{(a)} \textbf{EBG vs.\ CoT}, where each cell reports \textbf{$\Delta=\texttt{Ans}-\texttt{Ans}^{c}$} in percentage points; positive values favor EBG and negative values favor CoT. \textbf{(b)} \textbf{EBG vs.\ CoT-EBG,} where each cell reports \textbf{$\Delta=\texttt{Ans}-\texttt{Ans}^{ce}$}; positive values favor EBG and negative values favor CoT-EBG. Cells marked \textit{N/A} indicate that no results are available for the corresponding model--context combination.}
\label{fig:ebg_vs_cot_delta_and_ebg_vs_cot_ebg_delta}
\end{figure*}

% \begin{figure*}[t]
%     \centering
%     \includegraphics[width=\linewidth]
%         {images/ebg_vs_cot_delta.pdf}
%     \caption{\textbf{Matched answer-accuracy difference between EBG and CoT
%     prompting} on the NoLiMa-based \textit{LC} benchmark, reported separately
%     for one-hop and two-hop questions. Each cell reports
%     $\Delta=\texttt{Ans}-\texttt{Ans}^{c}$ in percentage points, where the
%     unmarked score denotes EBG and $^{c}$ denotes CoT. Positive values favor
%     EBG; negative values favor CoT. Cells marked \textit{N/A} indicate that no available results for that model-context combination}
%     \label{fig:ebg_vs_cot}
% \end{figure*}

\paragraph{Observations.}
As shown in Fig.~\ref{fig:ebg_vs_cot_delta_ans}, the effect of CoT is substantially
larger for two-hop questions than for one-hop questions. On a model-macro
average, CoT improves answer accuracy by approximately $8.5$\,pp for one-hop
questions and $27.2$\,pp for two-hop questions. At the model level, CoT never
reduces aggregate one-hop accuracy: it improves accuracy for eight models and
matches EBG for \texttt{GPT-5} and \texttt{Gemini-3-Fl}. The largest one-hop
gains occur for \texttt{GPT-OSS-120B} ($+23.6$\,pp), \texttt{GPT-4o}
($+20.0$\,pp), and \texttt{o3-mini} ($+19.3$\,pp).

For two-hop questions, CoT achieves higher aggregate answer accuracy for every
evaluated model. The improvements range from $5.9$\,pp for \texttt{GPT-5} to
$68.8$\,pp for \texttt{GPT-OSS-120B}, with large gains also observed for
\texttt{GLM-5.2} ($+42.3$\,pp), \texttt{o3-mini} ($+33.3$\,pp),
\texttt{Gemini-2.5-Fl} ($+31.3$\,pp), and \texttt{GPT-4o} ($+30.8$\,pp).
This larger two-hop effect suggests that explicitly eliciting a short reasoning
step is particularly useful when answering requires an additional inference
beyond direct needle retrieval. Because vanilla CoT also removes the citation
and JSON-output requirements, however, this comparison should be interpreted
as a comparison between complete prompting strategies rather than as an
isolated estimate of the effect of reasoning alone.

\subsection{EBG vs.\ CoT-EBG Prompting on NoLiMa-based Dataset}
\label{app:ebg_vs_cot_ebg}

\paragraph{Setup.}
To examine whether explicit reasoning remains beneficial when evidence citation
is retained, we compare standard EBG with CoT-EBG. We report \texttt{Ans} so
that this comparison is directly commensurable with the vanilla-CoT comparison
in Appendix~\ref{app:ebg_vs_cot}, which can only be scored on answer accuracy
because the plain CoT prompt requests no citations. Holding the metric fixed
across the two experiments isolates the effect of the citation requirement
itself. We define $\Delta = \texttt{Ans} - \texttt{Ans}^{ce}$, where $ce$
denotes CoT-EBG; negative values therefore favor CoT-EBG.

% \begin{figure*}[t]
% \centering
% \includegraphics[width=\linewidth]{images/ebg_vs_cot_ebg_delta_ans.pdf}
% \caption{\textbf{Matched answer-accuracy difference between EBG and CoT-EBG prompting} on the NoLiMa-based \textit{LC} benchmark, reported separately for one-hop and two-hop questions. Each cell reports $\Delta = \texttt{Ans} - \texttt{Ans}^{ce}$ in percentage points, where the unmarked score denotes EBG and $ce$ denotes CoT-EBG. Positive values favor EBG; negative values favor CoT-EBG. Cells marked N/A indicate that no available results for that
% model--context combination.}
% \label{fig:ebg_vs_cot_ebg}
% \end{figure*}

\paragraph{Observations.}
Fig.~\ref{fig:ebg_vs_cot_ebg_delta_ans} shows a smaller and more model-dependent effect
for one-hop questions. CoT-EBG improves aggregate answer accuracy for five models, matches EBG for two, and lowers it for three, with a model-macro average of $2.2$~pp in favor of CoT-EBG. This mixed result is consistent with
the near-ceiling performance of several models on direct one-hop retrieval, where an additional reasoning instruction has limited room to help and may introduce unnecessary output complexity. The two-hop result is more consistent: CoT-EBG improves aggregate \texttt{Ans} accuracy for every evaluated model, by $12.3$~pp on a model-macro average. The gains range from $2.4$~pp for \texttt{GPT-4.1} to $25.5$~pp for \texttt{Gemini-2.5-Fl}. Substantial improvements are also observed for \texttt{GPT-OSS-120B} and \texttt{GLM-5.2} ($+23.6$~pp each), \texttt{o3-mini} ($+12.0$~pp), and \texttt{Gemini-3-Fl} ($+10.4$~pp). Thus, explicitly eliciting a brief reasoning step remains beneficial when the evidence-citation requirement and output schema are retained, particularly for multi-hop
questions.

Taken together, the two experiments suggest that the gains from structured prompting are not attributable to citation formatting alone. \textbf{Explicit reasoning improves answer accuracy both under vanilla CoT and when the EBG citation
requirement is retained, with the largest and most consistent effects appearing
in the two-hop setting}.

\section{RAG on NoLiMa-based Dataset}\label{app:nolima_rag}
We use \texttt{text-embedding-3-small} to generate embeddings for both the NoLiMa haystack lines (with inserted needles) and the corresponding queries, and perform retrieval using cosine similarity.

\begin{table*}[h!]
\centering
\resizebox{\textwidth}{!}{
\begin{tabular}{lccccccccccccccccccc}
\toprule
& \multicolumn{19}{c}{\textbf{Context Length}}\\
% \cmidrule{2-20}
% \textbf{Metric} 
& \textbf{10K} & \textbf{20K} & \textbf{30K} & \textbf{40K}
& \textbf{50K} & \textbf{60K} & \textbf{70K} & \textbf{80K}
& \textbf{90K} & \textbf{100K} & \textbf{200K} & \textbf{300K}
& \textbf{400K} & \textbf{500K} & \textbf{600K} & \textbf{700K}
& \textbf{800K} & \textbf{900K} & \textbf{1M} \\
\midrule

\# Vectors
& 66 & 124 & 173 & 247 & 297 & 381 & 395 & 470 & 513 & 584
& 1,179 & 1,739 & 2,275 & 2,852 & 3,468 & 4,034 & 4,658 & 5,247 & 5,730 \\
\midrule

\rowcolor{gray!15}\multicolumn{20}{l}{\textit{Full-coverage (one-hop)}}\\
\# Vectors ($\downarrow$)
& 2 & 4 & 4 & 4 & 4 & 9 & 11 & 8 & 9 & 10
& 20 & 35 & 43 & 48 & 59 & 62 & 83 & 92 & 96 
\\
Chars ($\downarrow$)
& 350 & 823 & 818 & 607 & 584 & 1,902 & 2,343 & 1,876 & 1,985 & 2,452
& 4,725 & 7,126 & 8,244 & 11,548 & 12,410 & 13,981 & 17,771 & 20,234 & 22,910 \\
\midrule
\rowcolor{gray!15}\multicolumn{20}{l}{\textit{Full-coverage (two-hop)}}\\
\# Vectors ($\downarrow$)
& 10 & 21 & 25 & 36 & 42 & 63 & 71 & 82 & 78 & 92
& 169 & 276 & 348 & 444 & 528 & 635 & 750 & 821 & 890 \\
Chars ($\downarrow$)
& 1,956 & 4,756 & 5,875 & 6,771 & 9,906 & 12,600 & 15,902 & 16,841 & 17,236 & 20,567
& 39,052 & 63,009 & 77,792 & 99,519 & 117,251 & 136,473 & 164,002 & 177,879 & 202,412 \\

\bottomrule
\end{tabular}
}
\caption{\textbf{Retrieval analysis of \textit{RAG} under NoLiMa}, showing average total vector count; minimum full-coverage retrieval depth for one-hop and two-hop queries (i.e., the smallest top-$k$ such that all required needles/evidence facts are contained within the retrieved set for every sample) and corresponding retrieved character budget across varying haystack context lengths, averaged over three books. Lower ($\downarrow$) values are better.}
\label{tab:topk_budget_summary}
\end{table*}

\subsection{Retrieval Analysis}
Table~\ref{tab:topk_budget_summary} summarizes how the retrieval
requirements of a RAG system changes as the haystack context length grows
from $10\mathrm{K}$ to $1\mathrm{M}$. The first row reports the average
number of vectors, or indexed chunks, produced from the full context. As
expected, this number increases with context length: for example, the
retriever searches over only $66$ vectors at $10\mathrm{K}$ context, but
over $5{,}730$ vectors at $1\mathrm{M}$ context.

The remaining rows measure the retrieval depth required for
\emph{full-coverage}. Here, full coverage means that the retrieved
top-$k$ set contains all evidence facts needed to answer every query.
Thus, the reported number of vectors is the smallest value of $k$ such
that no required evidence is missed.

For one-hop queries, where the answer depends on a single evidence fact,
the required retrieval depth grows moderately with context length. At
$10\mathrm{K}$ context, retrieving only $2$ vectors, corresponding to
$350$ characters, is sufficient for full coverage. However, at
$1\mathrm{M}$ context, full coverage requires retrieving $96$ vectors,
corresponding to $22{,}910$ characters. This shows that even relatively
simple queries require a larger retrieval budget as the haystack becomes
longer.

For two-hop queries, where the answer depends on combining multiple
evidence facts, the retrieval requirement grows much more sharply. At
$10\mathrm{K}$ context, full coverage requires retrieving $10$ vectors,
or $1{,}956$ characters. At $1\mathrm{M}$ context, however, it requires
retrieving $890$ vectors, or $202{,}412$ characters. This indicates that
multi-hop evidence is substantially harder to retrieve in long-context
settings, since the relevant pieces of information may be distributed
across many more chunks.

Several observations follow from these results. First, the total number
of indexed vectors grows approximately linearly with context length,
increasing from $66$ vectors at $10\mathrm{K}$ to $5{,}730$ vectors at
$1\mathrm{M}$. Second, the retrieval depth required for full coverage is
much smaller than the total number of indexed vectors, but it still grows
substantially as the context becomes longer. Third, two-hop queries are
consistently more demanding than one-hop queries: at $1\mathrm{M}$
context, two-hop full coverage requires $890$ vectors, compared with only
$96$ vectors for one-hop full coverage. This is more than a $9\times$
increase in the number of retrieved vectors:
\[
\frac{890}{96} \approx 9.27.
\]

\begin{figure}[t!]
    \centering
    \includegraphics[width=\linewidth]{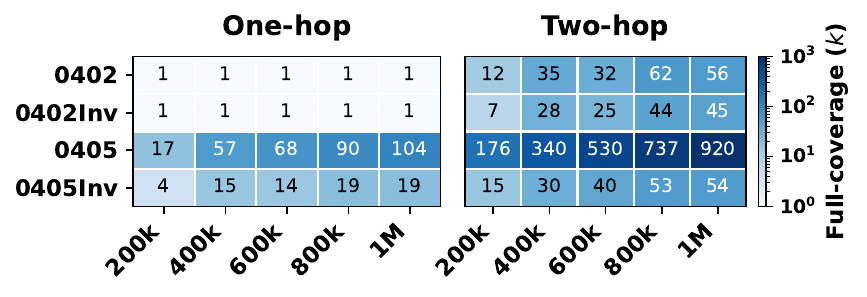}
    \caption{Minimum \textbf{full-coverage retrieval depth ($k$)} (for book $B^c_1$ under \textsc{NoLiMa}). Lower ($\downarrow$) values are better.}
    \label{fig:rag_topk_requirement_grouped}
\end{figure}
Fourth, the retrieved character budget also grows significantly. At
$1\mathrm{M}$ context, one-hop full coverage requires $22{,}910$
characters, whereas two-hop full coverage requires $202{,}412$
characters. Thus, two-hop retrieval requires nearly nine times more
retrieved text:
\[
\frac{202{,}412}{22{,}910} \approx 8.83.
\]

Another important observation is that standard small retrieval budgets,
such as top-$5$ or top-$10$, are unlikely to provide full coverage in
long-context settings. For instance, at $100\mathrm{K}$ context,
one-hop full coverage already requires top-$10$, while two-hop full
coverage requires top-$92$. At $1\mathrm{M}$ context, the required
two-hop retrieval depth increases to top-$890$, which is far beyond
typical RAG configurations.

Overall, the table shows that standard small top-$k$ retrieval settings
may be insufficient for long-context RAG, especially for multi-hop
queries. While one-hop retrieval remains relatively manageable, two-hop
retrieval at $1\mathrm{M}$ context requires retrieving a large fraction
of the indexed chunks:
\[
\frac{890}{5730} \approx 15.5\%.
\]
Therefore, as context length increases, guaranteeing full evidence
coverage can require a very large retrieved character budget, reducing
the efficiency advantage usually associated with RAG.

% \paragraph{Needle-specific Analysis.}

% Fig.~\ref{fig:rag_per_needle_by_hop} summarizes per-needle combined RAG accuracy by hop, highlighting plateau performance once retrieval is sufficiently broad.

% \paragraph{Observations.}
% \begin{itemize}[leftmargin=1em, itemsep=0pt, topsep=0pt, parsep=0pt, partopsep=0pt]
%     \item \textbf{One-hop plateaus high across needles,} with \texttt{0402}-family above the \texttt{0405}-family, reflecting easier evidence localization and higher downstream ceiling.
%     \item \textbf{Two-hop exhibits both lower ceilings and larger needle gaps.} The \texttt{0405} needle under two-hop reaches a substantially lower plateau than \texttt{0402}-family needles, consistent with its higher retrieval difficulty and downstream reasoning/citation sensitivity.
% \end{itemize}

\subsection{Character and Pronoun Sensitivity in Needle Retrieval}
\label{sec:rag_character_pronoun}

\paragraph{Motivation.}
During retrieval analysis, we observed that the \texttt{0405} needle family requires substantially larger top-$k$ budgets to achieve reliable retrieval compared to the \texttt{0402} family (Figure~\ref{fig:rag_topk_requirement_grouped}). This suggests that retrieval difficulty may be influenced not only by reasoning complexity but also by the \emph{surface form of the needle itself}. The default \texttt{0405} template contains a gendered pronoun:

\begin{quote}
\textit{``Then \{CHAR\} mentioned that he has been vegan for years.''}
\end{quote}

Because the pronoun \textit{he} is independent of the inserted character name, it may introduce additional semantic signals in the embedding representation that are not directly relevant to the query.

\begin{figure}[t]
\centering
\includegraphics[width=\linewidth]{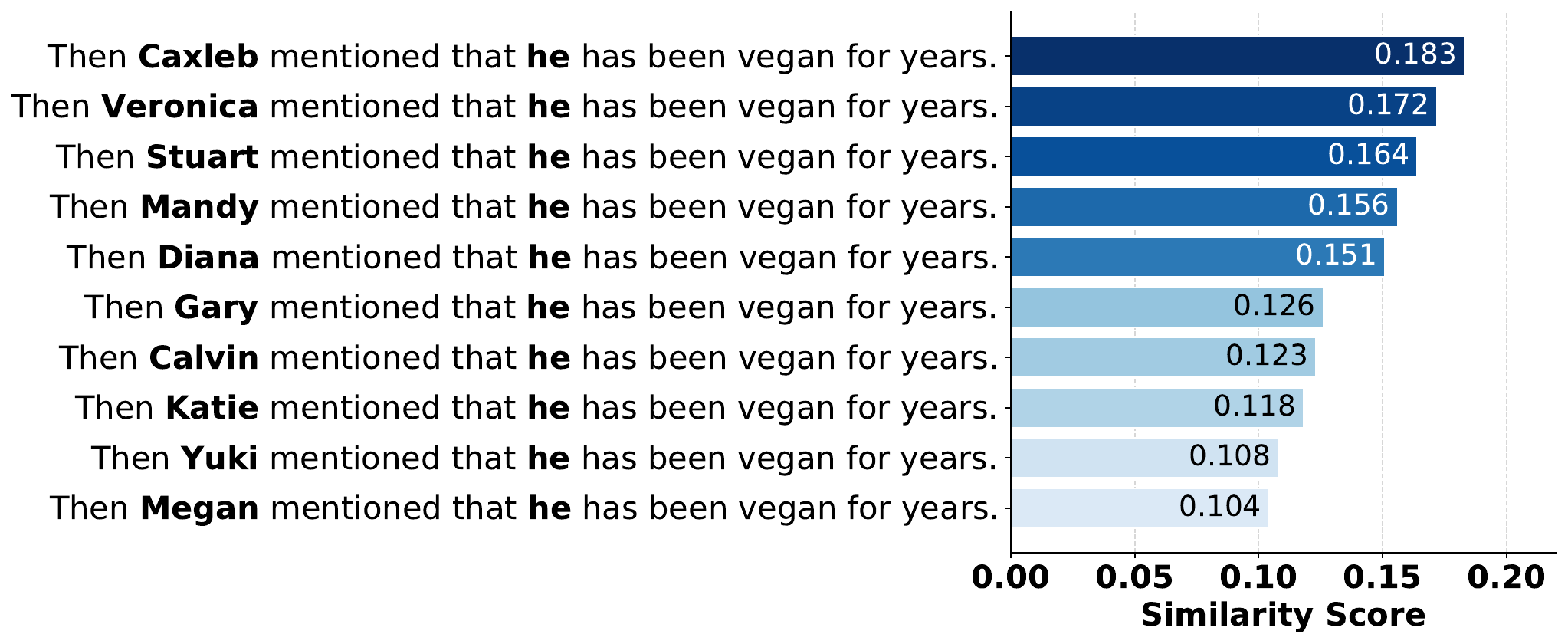}
\caption{\textbf{Ranking of the \texttt{0405} needle instantiated with 10 different character names} based on cosine similarity to a one-hop test query. Although the sentence semantics remain unchanged, similarity scores vary across character tokens. Higher ($\uparrow$) values are better.}
\label{fig:0405_character_similarity}
\end{figure}

\paragraph{Character sensitivity analysis.}
To investigate this effect, we compute cosine similarity between the query embedding and needle sentences instantiated with different character names drawn from the NoLiMa corpus. Figure~\ref{fig:0405_character_similarity} shows the top retrieved variants of the needle and their similarity scores for the query \textit{``Which character cannot eat Brandade?''}.

% \begin{figure}[t]
% \centering
% \includegraphics[width=0.95\linewidth]{rag_plots/character_analysis/0405.png}
% \caption{Top-10 cosine similarities between the query embedding and the 0405 needle instantiated with different character names. Even though the semantic meaning of the sentence remains identical, similarity scores vary depending on the character token.}
% \label{fig:0405_character_similarity}
% \end{figure}

Although the sentence semantics remain unchanged, we observe noticeable variation in similarity scores across character substitutions. This indicates that character tokens themselves influence the embedding representation and can affect retrieval ranking.

\paragraph{Effect of pronoun removal.}
To isolate the role of the gendered pronoun, we construct a rephrased variant of the needle (\texttt{0405-R}, Table \ref{tab:needle_templates}) that removes the pronoun while preserving the meaning:

\begin{quote}
\textit{``Then \{CHAR\} mentioned being vegan for years.''}
\end{quote}

\begin{figure}[t]
\centering
\includegraphics[width=\linewidth]{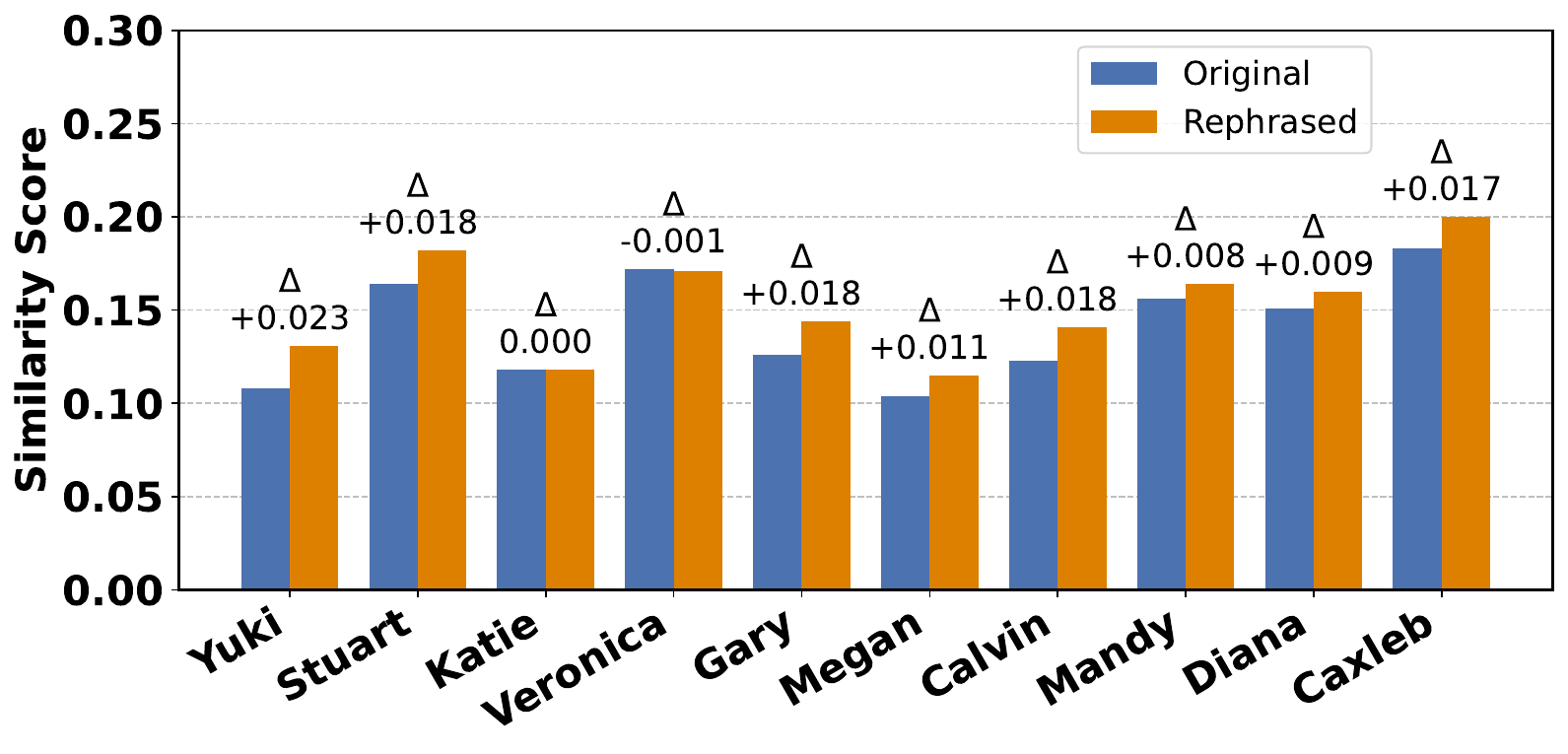}
\caption{Cosine similarity \textbf{comparison between the original \texttt{0405} needle and the rephrased \texttt{0405-R} variant across 10 characters}. Removing the pronoun consistently increases similarity scores across different characters. Higher positive difference ($\uparrow$) indicates greater effectiveness of the rephrased version compared to the original.}
\label{fig:0405_rephrased_similarity}
\end{figure}

We recompute similarity scores for both the original and rephrased needles across multiple character substitutions. Figure~\ref{fig:0405_rephrased_similarity} compares cosine similarity between the two formulations.

% \begin{figure}[t]
% \centering
% \includegraphics[width=0.95\linewidth]{rag_plots/character_analysis/CharacterWOGender.png}
% \caption{Cosine similarity comparison between the original 0405 needle and the rephrased variant without the gendered pronoun. Removing the pronoun consistently increases similarity scores across most character substitutions.}
% \label{fig:0405_rephrased_similarity}
% \end{figure}
% \begin{figure}[t]
% \centering
% \includegraphics[width=0.95\linewidth]{rag_pdf_plots/characters_rephrased_similarity_comparison.pdf}
% \caption{Cosine similarity comparison between the original 0405 needle and the rephrased variant without the gendered pronoun. Removing the pronoun consistently increases similarity scores across most character substitutions.}
% \label{fig:0405_rephrased_similarity}
% \end{figure}

Across most characters, the rephrased template yields slightly higher similarity scores. While the improvement is modest for individual instances, it is systematic across characters, suggesting that removing the pronoun reduces semantic noise in the embedding space.

\paragraph{Implications for RAG retrieval.}
These results help explain the higher retrieval budget required for the \texttt{0405} needle family observed earlier. When the needle contains additional syntactic elements that are not directly aligned with the query semantics, embedding similarity becomes less discriminative. In large haystacks, even small similarity differences can significantly affect ranking, pushing the correct needle lower in the retrieval list and increasing the required top-$k$ budget.

\begin{table*}[t!]
\centering
\resizebox{\textwidth}{!}{
\begin{tabular}{lccccccccccccccccccc}
\toprule
& \multicolumn{19}{c}{\textbf{Context Length}}\\
& \textbf{10K} & \textbf{20K} & \textbf{30K} & \textbf{40K}
& \textbf{50K} & \textbf{60K} & \textbf{70K} & \textbf{80K}
& \textbf{90K} & \textbf{100K} & \textbf{200K} & \textbf{300K}
& \textbf{400K} & \textbf{500K} & \textbf{600K} & \textbf{700K}
& \textbf{800K} & \textbf{900K} & \textbf{1M} \\
\midrule

\# Vectors
& 71 & 135 & 169 & 226 & 290 & 392 & 407 & 453 & 515 & 542 & 1{,}161 & 1{,}775 & 2{,}319 & 2{,}858 & 3{,}502 & 4{,}086 & 4{,}641 & 5{,}229 & 5{,}671 \\
\midrule

\rowcolor{gray!15}\multicolumn{20}{l}{\textit{Full-coverage (one-hop)}}\\

\multicolumn{20}{l}{\textbf{Embedding-RAG}}\\
\# Vectors ($\downarrow$)
& 2 & 4 & 4 & 3 & 1 & 11 & 13 & 4 & 7 & 14 & 17 & 41 & 57 & 48 & 68 & 45 & 90 & 94 & 104 \\
Chars ($\downarrow$)
& 118 & 850 & 758 & 363 & 62 & 1{,}556 & 2{,}258 & 1{,}415 & 1{,}721 & 3{,}284 & 4{,}985 & 7{,}879 & 10{,}185 & 11{,}106 & 12{,}182 & 9{,}589 & 18{,}310 & 22{,}216 & 23{,}589 \\
\multicolumn{20}{l}{\textbf{BM25-RAG}}\\
\# Vectors ($\downarrow$)
& 5 & 8 & 6 & 9 & 13 & 18 & 31 & 20 & 37 & 33 & 59 & 110 & 147 & 168 & 214 & 195 & 251 & 285 & 307 \\
Chars ($\downarrow$)
& 1{,}607 & 2{,}919 & 1{,}238 & 2{,}438 & 4{,}858 & 3{,}603 & 9{,}067 & 6{,}816 & 12{,}364 & 10{,}194 & 21{,}656 & 35{,}511 & 49{,}102 & 56{,}184 & 65{,}122 & 70{,}084 & 80{,}559 & 94{,}943 & 102{,}475 \\
\multicolumn{20}{l}{\textbf{Reranker-RAG}}\\
\# Vectors ($\downarrow$)
& 1 & 1 & 1 & 1 & 1 & 1 & 1 & 1 & 1 & 1 & 1 & 1 & 1 & 1 & 1 & 1 & 1 & 1 & 1 \\
Chars ($\downarrow$)
& 62 & 62 & 62 & 62 & 62 & 62 & 62 & 62 & 62 & 62 & 62 & 62 & 62 & 62 & 62 & 62 & 62 & 62 & 62 \\
\midrule

\rowcolor{gray!15}\multicolumn{20}{l}{\textit{Full-coverage (two-hop)}}\\
\multicolumn{20}{l}{\textbf{Embedding-RAG}}\\
\# Vectors ($\downarrow$)
& 11 & 24 & 18 & 45 & 41 & 72 & 69 & 65 & 70 & 101 & 176 & 287 & 340 & 422 & 530 & 630 & 737 & 823 & 920 \\
Chars ($\downarrow$)
& 1{,}681 & 3{,}706 & 3{,}518 & 9{,}442 & 10{,}675 & 12{,}552 & 13{,}877 & 15{,}742 & 18{,}841 & 24{,}965 & 45{,}742 & 66{,}739 & 72{,}493 & 93{,}694 & 113{,}052 & 136{,}498 & 167{,}983 & 180{,}487 & 217{,}152 \\
\multicolumn{20}{l}{\textbf{BM25-RAG}}\\
\# Vectors ($\downarrow$)
& 13 & 33 & 24 & 49 & 50 & 78 & 82 & 74 & 90 & 118 & 202 & 337 & 407 & 510 & 640 & 742 & 855 & 971 & 1{,}061 \\
Chars ($\downarrow$)
& 2{,}877 & 6{,}085 & 5{,}582 & 11{,}192 & 13{,}144 & 14{,}521 & 18{,}093 & 17{,}787 & 26{,}198 & 30{,}901 & 55{,}545 & 84{,}908 & 98{,}295 & 123{,}608 & 150{,}872 & 178{,}140 & 210{,}240 & 235{,}357 & 265{,}785 \\

\multicolumn{20}{l}{\textbf{Reranker-RAG}}\\
\# Vectors ($\downarrow$)
& 2 & 2 & 5 & 11 & 3 & 18 & 15 & 9 & 9 & 17 & 30 & 36 & 60 & 67 & 72 & 91 & 117 & 115 & 149 \\
Chars ($\downarrow$)
& 118 & 129 & 602 & 1{,}156 & 476 & 2{,}418 & 1{,}957 & 2{,}312 & 1{,}868 & 2{,}162 & 4{,}679 & 6{,}364 & 9{,}074 & 10{,}919 & 9{,}497 & 13{,}532 & 17{,}431 & 16{,}042 & 21{,}372 \\

\bottomrule
\end{tabular}
}
\caption{\textbf{Retrieval analysis of three RAG pipelines under NoLiMa}, showing total vector count; minimum full-coverage retrieval depth for one-hop and two-hop queries (i.e., the smallest top-$k$ such that all required needles/evidence facts are contained within the retrieved set for every sample) and corresponding retrieved character budget across varying haystack context lengths for BM25-RAG, Embedding-RAG, and Reranker-RAG. Here all the results are computed only on Book 1 ($B^c_1$). Lower ($\downarrow$) values are better.}
\label{tab:topk_budget_by_pipeline_full}
\end{table*}

\subsection{Retrieval Analysis under Embedding, BM25, and Reranker-RAG Pipelines}
\label{sec:embed_bm25_reranker}
Motivated by the results and observations in Table~\ref{tab:topk_budget_summary}, we further investigate whether the observed retrieval bottleneck is specific to embedding-based semantic retrieval or persists across alternative retrieval and ranking strategies.
\paragraph{Setup of Embedding-RAG.}
We first consider the embedding-based RAG pipeline from the preceding experiment (\S~\ref{sec:nolima_rag_setup}), restricting the evaluation to Book~1 ($B^c_1$). The pipeline follows the same retrieval setup as before: the haystack is divided into chunks, represented using dense vector embeddings, and ranked by semantic similarity to the query. We retain the original chunking, retrieval, and full-coverage evaluation protocol, and report the minimum top-$k$ required to retrieve all evidence for one-hop and two-hop queries, together with the corresponding retrieved character budget. Thus, the Embedding-RAG results in Table~\ref{tab:topk_budget_by_pipeline_full} serve as the semantic-retrieval baseline against which Okapi BM25 and Qwen3-Reranker-8B are compared. 
% Restricting all three pipelines to the same book ensures that differences in retrieval depth reflect the retrieval and ranking strategies rather than variation across source documents.
\paragraph{Setup of BM25-RAG. }To examine whether the retrieval limitations observed with embedding-based RAG are specific to semantic similarity search, we replace the vector embedding-based retriever with Okapi BM25\footnote{https://pypi.org/project/rank-bm25/}, a lexical retrieval method based on term matching and term-frequency statistics. We keep the evaluation protocol unchanged and measure the minimum retrieval depth required to achieve full evidence coverage for both one-hop and two-hop queries across increasing NoLiMa context lengths. As in our primary analysis, full coverage is defined as the smallest top-$k$ for which all required evidence facts are present in the retrieved set for every sample. We report both this retrieval depth and its corresponding character budget in Table~\ref{tab:topk_budget_by_pipeline_full}.

\paragraph{Setup of Reranker-based RAG.}
We further evaluate whether a stronger learned ranking stage can reduce the retrieval depth required for complete evidence coverage. Specifically, we use \texttt{Qwen3-Reranker-8B}, a fine-tuned reranking model, to reorder candidate passages (retrieved by the embedding-based RAG retriever) according to their relevance to the query. We then evaluate the resulting ranking using the same full-coverage criterion as the BM25-RAG and Embedding-RAG settings. This experiment tests the extent to which a learned reranker can move required evidence toward the top of the ranking and consequently reduce the context passed to the downstream model.

\paragraph{Observations.} We observe that BM25-RAG consistently requires substantially larger retrieval budgets than Embedding-RAG, with the difference becoming more pronounced as context length increases. At 1M characters, full coverage of one-hop queries requires 307 vectors (102,475 characters) with BM25-RAG, compared with 104 vectors (23,589 characters) for Embedding-RAG. The same pattern holds for two-hop queries, where BM25-RAG requires 1,061 vectors (265,785 characters), compared with 920 vectors (217,152 characters) for Embedding-RAG. Thus, replacing semantic retrieval with Okapi BM25 does not remove the full-coverage bottleneck; instead, lexical retrieval generally requires retrieving more of the haystack to guarantee that all necessary evidence is available. This gap suggests that semantic similarity provides a stronger ranking signal for isolating relevant evidence as the distractor pool expands. Moreover, the persistent two-hop overhead across both retrievers indicates that jointly recovering multiple evidence pieces remains intrinsically more demanding than single-evidence retrieval.

As shown in Table~\ref{tab:topk_budget_by_pipeline_full}, Qwen3-Reranker-8B improves retrieval efficiency over BM25-RAG and Embedding-RAG. For one-hop queries, evidence ranks first across all context lengths, yielding full coverage at top-$1$ with 62 characters. For two-hop queries, depth grows with context length: at 1M characters, Reranker-RAG reaches full coverage with 149 vectors (21,372 characters), compared with 920 vectors (217,152 characters) for Embedding-RAG and 1,061 vectors (265,785 characters) for BM25-RAG. Thus, a fine-tuned reranker dramatically reduces the retrieval budget for full evidence coverage. Nevertheless, increasing top-$k$ for two-hop queries shows that stronger reranking does not eliminate the difficulty of jointly surfacing evidence as the haystack grows. The one-hop--two-hop contrast suggests that evidence multiplicity, not context length alone, drives the remaining burden: joint coverage becomes harder when multiple passages must survive the cutoff.

\subsection{Randomly Sampled Haystacks vs. Top-$k$ Ranked Retrieval}\label{app:nolima_lc_ebg_rag_ebg}

\paragraph{Setup.}
We examine whether LLMs answer more accurately when we give them a randomly sampled long-context haystack or a compact context formed by top-$k$ ranked retrieval. In the long-context setting, we provide the model with the Book~1 haystack at a target character length and insert the needle at predefined depths. This setting corresponds to randomly sampled haystacks: the long-context construction procedure embeds the relevant evidence within a larger context and determines the surrounding distractors and needle positions. In contrast, the retrieval setting replaces the full randomly sampled haystack with a retrieval-ranked context. For each query, we compute cosine similarity between the query embedding and each chunk embedding in the corresponding vector database, rank all chunks by similarity score, and select the minimum top-$k$ prefix that contains all gold needle evidence for that condition. We call this value the best top-$k$.

Because long-context and retrieval inputs differ substantially in structure, we compare them under approximately matched character budgets rather than matched nominal context labels. For each Book~1 haystack, we first compute the total character count of the best top-$k$ retrieved chunks. We then select the long-context condition whose haystack character count most closely matches that retrieval character count. For example, the best top-$k$ retrieved context for the 1M-character haystack contains 217,152 characters; we therefore compare this retrieval condition against the 200K-character long-context condition, since the two inputs match most closely in character budget. Throughout, ``context length'' denotes character count rather than token count.

\begin{table}[t]
\centering
\small
\resizebox{\columnwidth}{!}{
\begin{tabular}{cccc}
\toprule
\textbf{LC} & \multicolumn{2}{c}{\textbf{Full-coverage ($k$)}} & \textbf{LC (matched)} \\
\cmidrule{2-3}
\textbf{(Chars)} & \textbf{(Vectors)} & \textbf{(Chars)} & \textbf{(Chars)} \\
\midrule
200K & 176 & 45{,}742  & 50K  \\
400K & 340 & 72{,}493  & 70K  \\
500K & 422 & 93{,}694  & 90K  \\
600K & 530 & 113{,}052 & 100K \\
1M   & 920 & 217{,}152 & 200K \\
\bottomrule
\end{tabular}
}
\caption{\textbf{Character-budget matching procedure}. For each long-context (LC) haystack used during the retrieval stage on book $B^c_1$, we compute the minimum top-$k$ such that the retrieved set contains all required needles and evidence facts for every sample. We then measure the character count of this retrieved set and match it to the LC condition with the closest character count. The columns report, from left to right, the original LC haystack size, the minimum top-$k$ (in vectors), its character count, and the matched LC budget used for comparison.}
\label{tab:context_matching}
\end{table}

We use the same answer+evidence prompt as the long-context EBG setting, constructed from the selected top-$k$ chunks and gold evidence metadata. Models produce an answer with cited evidence. We denote retrieval and matched long-context answer accuracy as $\mathrm{Ans}_{\mathrm{RAG}}$ and $\mathrm{Ans}_{\mathrm{LC}}$, respectively. The primary comparison is the accuracy delta:

\[
\Delta
=\mathrm{Ans}_{\mathrm{RAG}} -
\mathrm{Ans}_{\mathrm{LC}}
.
\]
A positive value means that top-$k$ ranked retrieval outperforms randomly sampled long-context prompting under the matched character budget, while a negative value means that randomly sampled long-context prompting outperforms top-$k$ ranked retrieval.

Table~\ref{tab:context_matching} summarizes the character-budget matching procedure. The table reports, for each original Book~1 haystack length, the best top-$k$ value, the character count of the corresponding retrieved context, and the matched long-context character budget we use for comparison.

\begin{figure}[t]
    \centering
    \includegraphics[width=\linewidth]{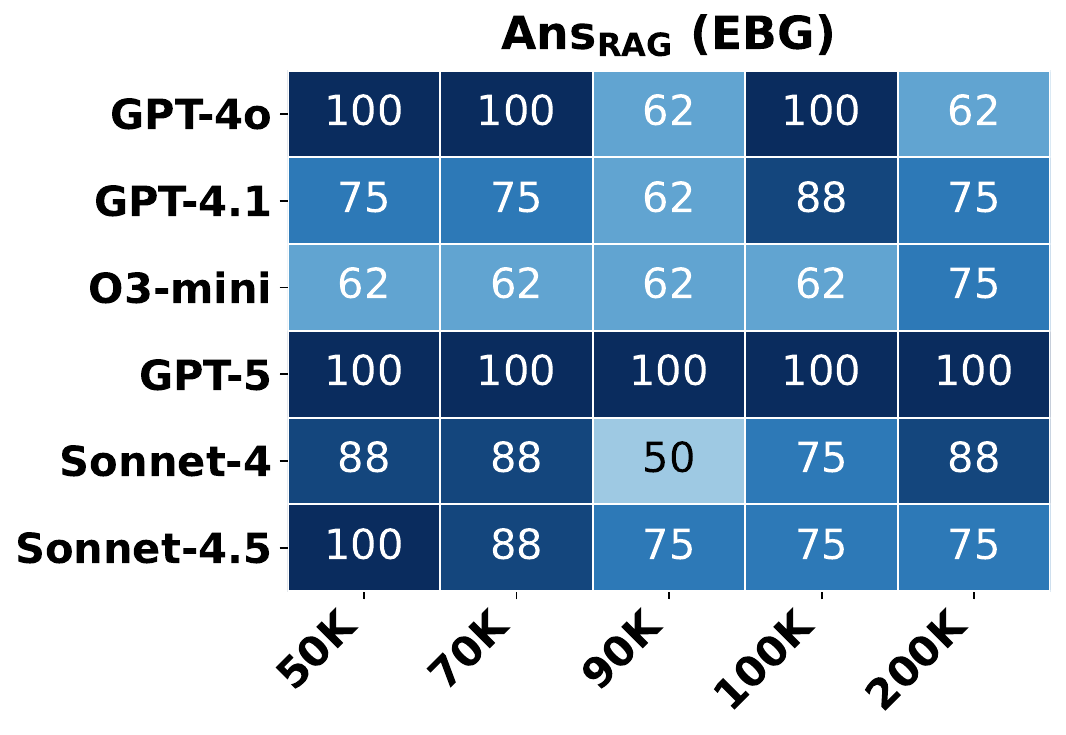}
    \caption{
    \textbf{Retrieval answer accuracy under the EBG prompting for the matched top-$k$ retrieval contexts}.
    Columns denote the long-context character budgets we use for comparison after matching each retrieved
    top-$k$ context to the nearest long-context haystack length. Rows denote the evaluated models.
    Each cell reports answer accuracy, $\mathrm{Ans}_{\mathrm{RAG}}$, after we condition the model
    on the retrieved top-$k$ chunks rather than the randomly sampled long-context haystack. Higher ($\uparrow$) values are better.
    }
    \label{fig:rag_answer_heatmap}
\end{figure}

\begin{figure}[t]
    \centering
    \includegraphics[width=\linewidth]{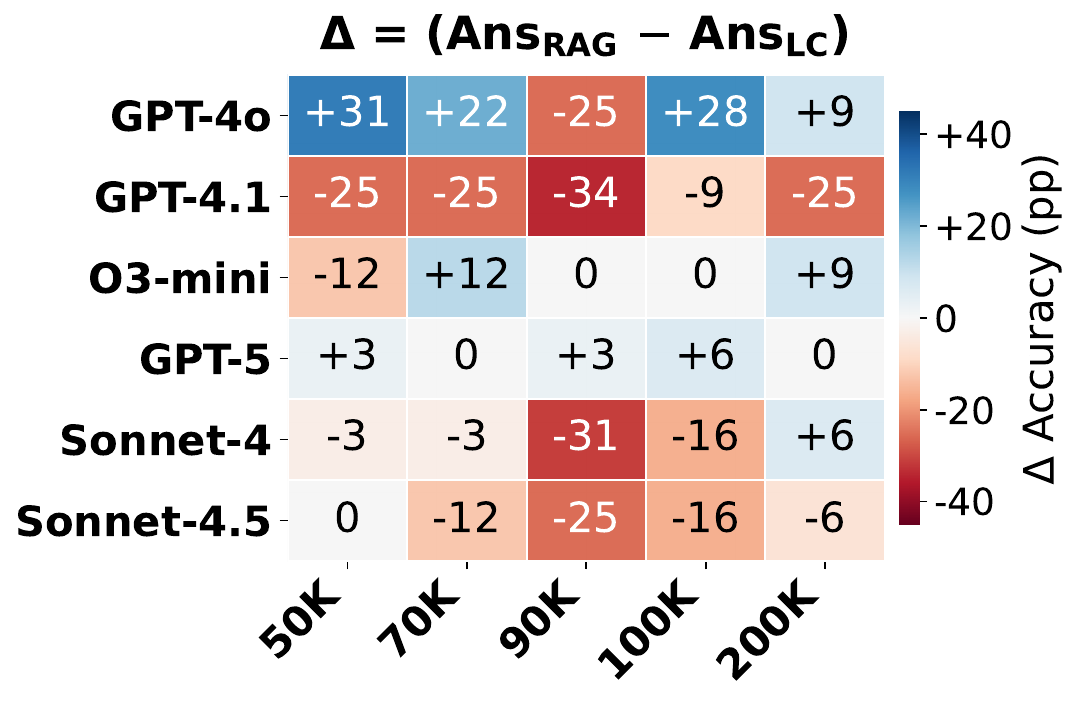}
    \caption{\textbf{Answer accuracy difference  between RAG-EBG and LC-EBG}, compared under approximately matched character budgets (Table~\ref{tab:context_matching}). Positive values indicate that RAG-EBG outperforms LC-EBG; negative values indicate that LC-EBG outperforms RAG-EBG. Higher~($\uparrow$) positive values are better for RAG.}
    \label{fig:delta_ans}
\end{figure}

Figure~\ref{fig:rag_answer_heatmap} reports retrieval answer accuracy under the answer+evidence prompt, and Figure~\ref{fig:delta_ans} reports the resulting $\Delta$ difference values across models and matched context lengths.

\paragraph{Observations.}
Figure~\ref{fig:delta_ans} shows that the relative advantage of top-$k$ ranked retrieval versus randomly sampled long-context prompting depends strongly on the model. Since each cell reports
$\Delta =\mathrm{Ans}_{\mathrm{RAG}}-\mathrm{Ans}_{\mathrm{LC}}$, positive values indicate that the top-$k$ retrieved context performs better, whereas negative values indicate that the randomly sampled long-context haystack performs better.

Overall, top-$k$ ranked retrieval competes strongly with long-context prompting under the matched character-budget setting. For GPT-4o, retrieval outperforms long-context prompting in four of the five matched budgets: 50K ($+31$\,pp), 70K ($+22$\,pp), 100K ($+28$\,pp), and 200K ($+9$\,pp). GPT-4o therefore benefits substantially when we replace the larger randomly sampled haystack with a compact retrieval-ranked context. GPT-5 also shows near parity between the two settings, with deltas close to zero across all budgets ($+3$, $0$, $+3$, $+6$, and $0$\,pp). For GPT-5, once the retrieved top-$k$ context contains the relevant evidence, the additional surrounding haystack context provides little or no answer-accuracy benefit.

In contrast, GPT-4.1 consistently favors the randomly sampled long-context haystack. Its deltas remain negative at every matched budget, ranging from $-9$\,pp at 100K to $-34$\,pp at 90K. GPT-4.1 therefore exploits information from the broader haystack context more effectively than from the compact top-$k$ context, even when the retrieved context contains the necessary gold evidence. The Claude models show a similar but weaker long-context preference. Sonnet-4 improves under long-context prompting at 50K, 70K, 90K, and 100K, with the largest gain at 90K ($-31$\,pp), but retrieval performs slightly better at 200K ($+6$\,pp). Sonnet-4.5 also favors long-context prompting at most matched budgets, especially 90K ($-25$\,pp) and 100K ($-16$\,pp), while matching retrieval at 50K and showing only a small long-context advantage at 200K ($-6$\,pp). O3-mini swings unpredictably across budgets in both directions: long-context prompting helps at 50K ($-12$\,pp), retrieval helps at 70K and 200K ($+12$\,pp and $+9$\,pp), and the two settings tie at 90K and 100K.

% \begin{table}[t!]
% \centering
% \small
% \resizebox{\linewidth}{!}{
% \begin{tabular}{lll}
% \toprule
% \textbf{Dataset} & \textbf{Source} & \textbf{License} \\
% \midrule
% Synthetic & Ours & \href{https://creativecommons.org/licenses/by/4.0/}{CC By 4.0} \\
% \textsc{NoLiMa}-based        & \citet{modarressi2025nolima} & \href{https://github.com/adobe-research/NoLiMa?tab=License-1-ov-file}{ADOBE RESEARCH LICENSE} \\
% \textsc{MultiHop-RAG}        & \citet{tang2024multihop} & \href{https://opendatacommons.org/licenses/by/1-0/}{ODC-By} \\
% \textsc{QASPER-MP}           & \citet{dasigi-etal-2021-qasper} & \href{https://creativecommons.org/licenses/by/4.0/}{CC By 4.0}\\
% \bottomrule
% \end{tabular}
% }
% \caption{Datasets used in this work, along with their source references and licenses.}
% \label{tab:dataset_sources}
% \end{table}

A notable pattern emerges at the 90K budget, which favors long-context prompting most strongly. Four models show large negative deltas: GPT-4o ($-25$,pp), GPT-4.1 ($-34$,pp), Sonnet-4 ($-31$,pp), and Sonnet-4.5 ($-25$,pp). Evidence completeness therefore does not guarantee retrieval superiority. Even when the top-$k$ context contains the required needles, the model may benefit from broader discourse, surrounding distractors, or positional structure of the original haystack. Retrieval can therefore alter the contextual relationships through which evidence is interpreted, making compression effectiveness model-dependent. Conversely, strong positive deltas for GPT-4o and near-zero deltas for GPT-5 show that compact retrieval removes distractor burden without hurting accuracy for some models.

Overall, the results do not support a single universal winner between randomly sampled haystacks and top-$k$ ranked retrieval. They instead reveal a model-specific tradeoff. Top-$k$ retrieval outperforms long-context prompting when compact evidence-focused contexts help the model avoid distraction, as GPT-4o demonstrates most clearly. Long-context prompting outperforms retrieval when models benefit from the broader haystack context, as GPT-4.1 and the Claude models demonstrate. GPT-5 emerges as the most retrieval-robust model in this comparison, showing that a sufficiently strong model answers equally well from the compact retrieved context and the matched randomly sampled haystack.

\subsection{Embedding-RAG vs Reranker-RAG under EBG Prompting}
\label{sec}

The preceding retrieval analysis (Table \ref{tab:topk_budget_by_pipeline_full}) shows that Qwen3-Reranker-8B can rank required evidence substantially closer to the top of the retrieved context than embedding-based similarity search. However, improved retrieval ranking does not necessarily translate into better end-to-end generation: even when the same evidence is available, its position within the retrieved context may affect a model's ability to identify, reason over, and correctly attribute that evidence. Motivated by this distinction, we evaluate whether the retrieval-order improvements from reranking translate into higher joint \texttt{Ans+Evid} accuracy when the amount of retrieved context is held fixed.

\begin{table*}[h!]
\centering
\small
\resizebox{\textwidth}{!}{
\begin{tabular}{lcccccccccc}
\toprule
& \multicolumn{2}{c}{\textbf{50K}} & \multicolumn{2}{c}{\textbf{70K}} & \multicolumn{2}{c}{\textbf{90K}} & \multicolumn{2}{c}{\textbf{100K}} & \multicolumn{2}{c}{\textbf{200K}} \\
\cmidrule(lr){2-3} \cmidrule(lr){4-5} \cmidrule(lr){6-7} \cmidrule(lr){8-9} \cmidrule(lr){10-11}
\textbf{Model} & \textbf{Emb-RAG} & \textbf{Reranker-RAG} & \textbf{Emb-RAG} & \textbf{Reranker-RAG} & \textbf{Emb-RAG} & \textbf{Reranker-RAG} & \textbf{Emb-RAG} & \textbf{Reranker-RAG} & \textbf{Emb-RAG} & \textbf{Reranker-RAG} \\
\midrule
Sonnet-4.5 & 100.0 & 100.0 & 87.5  & 100.0 & 75.0  & 100.0 & 75.0  & 100.0 & 75.0  & 87.5  \\
GPT-4.1           & 75.0  & 100.0 & 75.0  & 87.5  & 62.5  & 87.5  & 87.5  & 75.0  & 75.0  & 87.5  \\
GLM-5.2           & 62.5  & 50.0  & 62.5  & 75.0  & 75.0  & 75.0  & 50.0  & 75.0  & 50.0  & 75.0  \\
GPT-OSS-120B      & 62.5  & 75.0  & 75.0  & 87.5  & 87.5  & 87.5  & 62.5  & 100.0 & 50.0  & 87.5  \\
\bottomrule
\end{tabular}
}
\caption{\textbf{Comparison of Embedding-RAG and Reranker-RAG \texttt{Ans} accuracy under EBG prompting.} Columns are indexed by the LC-matched retrieved-context budgets from Table~\ref{tab:context_matching}; for example, \textbf{200K} denotes the retrieval budget matched to the \textbf{1M} LC haystack. Each cell reports answer accuracy over $n=8$ samples (4 needles $\times$ one-hop/two-hop). Emb: Embedding-RAG. Higher ($\uparrow$) values indicate better performance.}
\label{tab:embedding_vs_fixedtopk_ansonly}
\end{table*}
\begin{table*}[h!]
\centering
\small
\setlength{\tabcolsep}{4pt}
\resizebox{\textwidth}{!}{
\begin{tabular}{lllll}
\toprule
\textbf{Dataset}
& \textbf{Source}
& \textbf{License}
& \textbf{Eligible / Constructed Pool}
& \textbf{Samples Used} \\
\midrule

Synthetic
& Ours
& \href{https://creativecommons.org/licenses/by/4.0/}{CC BY 4.0}
& 66,834 distractor facts; 10 two-hop needles
& 100 \\

\textsc{NoLiMa}-based
& \citet{modarressi2025nolima}
& \href{https://github.com/adobe-research/NoLiMa?tab=License-1-ov-file}
        {Adobe Research License}
& 4 needle variants $\times$ 4 insertion depths
& 16 \\

\textsc{MultiHop-RAG}
& \citet{tang2024multihop}
& \href{https://opendatacommons.org/licenses/by/1-0/}{ODC-By}
& 2,255 answerable queries
& 50, 200 \\

\textsc{QASPER-MP}
& \citet{dasigi-etal-2021-qasper}
& \href{https://creativecommons.org/licenses/by/4.0/}{CC BY 4.0}
& Dev split: 1,005 questions over 281 papers
& 50, 200 \\

\bottomrule
\end{tabular}
}
\caption{\textbf{Dataset statistics.}
For the synthetic benchmark, we construct 66,834 unique single-sentence
distractor facts and curate 10 two-hop needles, evaluating 100
independently generated trials per context length.
For the \textsc{NoLiMa}-based benchmark, we use four primary needle variants
at four insertion depths, yielding 16 needle-placement conditions per context
length per model.
For \textsc{MultiHop-RAG}, removing 301 \texttt{null\_query} examples leaves
2,255 answerable queries; we evaluate 50 queries with proprietary models and
200 with open-weight models.
For \textsc{QASPER-MP}, experiments use the QASPER development split containing 1,005 questions over 281 papers. Each gold paper is paired with
ten distractor papers. We evaluate 50 queries (38 answerable and 12
unanswerable) with proprietary models and 200 queries (150 answerable and
50 unanswerable) with open-weight models.
}
\label{tab:dataset_sources}
\end{table*} 

\paragraph{Setup.}
We compare Embedding-RAG and Reranker-FixedTopk-RAG across four LLMs. For each LC-matched budget in Table~\ref{tab:context_matching}, Embedding-RAG retrieves full-coverage top-$k$ chunks by cosine similarity. Reranker-RAG uses \emph{the same chunks} and preserves the top-$k$ and character budget, but re-orders them using \texttt{Qwen3-Reranker-8B} relevance scores. No chunks are added or removed; the pipelines differ only in evidence ordering. This comparison isolates whether placing relevant evidence earlier improves generation rather than conflating ranking quality with context volume. With evidence availability and context size fixed, differences can be attributed to ranking and reveal whether models benefit from prioritizing relevant chunks. We evaluate both pipelines with our EBG prompt over $n=8$ examples, comprising four needles under one-hop and two-hop queries. Results appear in Table~\ref{tab:embedding_vs_fixedtopk_ansonly}.

\paragraph{Observations.}
Table~\ref{tab:embedding_vs_fixedtopk_ansonly} shows that reranking improves end-to-end answer accuracy over embedding-based retrieval, with gains particularly pronounced at larger context budgets. Sonnet-4.5 benefits consistently from reranking, reaching 100\% accuracy from 50K--100K and retaining an advantage at 200K. GPT-OSS-120B exhibits the largest gain at longer contexts, improving from 62.5\% to 100\% at 100K and from 50\% to 87.5\% at 200K. GPT-4.1 and GLM-5.2 also improve in most settings, although reranking is not uniformly beneficial: GPT-4.1 drops from 87.5\% to 75\% at 100K, while GLM-5.2 drops from 62.5\% to 50\% at 50K and remains unchanged at 90K. 

Overall, across the 20 model--budget combinations, Reranker-RAG improves over Embedding-RAG in 15 cases, ties in three, and underperforms in only two, indicating that \textbf{improved evidence ordering can translate into better downstream answer generation even when the retrieved context budget is fixed.}
 The larger gains at higher budgets suggest that retrieval quality remains important even when more context is available, particularly as irrelevant passages increasingly compete for the reader's attention. Because both pipelines contain identical chunks, these gains isolate ordering rather than evidence availability, suggesting that readers are sensitive to where relevant evidence appears within the retrieved context.

\begin{table*}[h!]
\centering
\small
\resizebox{\linewidth}{!}{
\begin{tabular}{llll}
\toprule
\textbf{Model} & \textbf{Provider} & \textbf{Source URL} & \textbf{License} \\
\midrule
GPT-4o        & OpenAI    & \url{https://platform.openai.com/docs/models/gpt-4o}       & Proprietary \\
GPT-4.1       & OpenAI    & \url{https://platform.openai.com/docs/models/gpt-4.1}      & Proprietary \\
o3-mini       & OpenAI    & \url{https://platform.openai.com/docs/models/o3-mini}      & Proprietary \\
GPT-5         & OpenAI    & \url{https://platform.openai.com/docs/models/gpt-5}        & Proprietary \\
GPT-5.5       & OpenAI    & \url{https://platform.openai.com/docs/models/gpt-5.5}      & Proprietary \\
Sonnet-4   & Anthropic & \url{https://www.anthropic.com/claude/sonnet}          & Proprietary \\
Sonnet-4.5 & Anthropic & \url{https://www.anthropic.com/claude/sonnet}          & Proprietary \\
Opus-4.6   & Anthropic & \url{https://www.anthropic.com/claude/opus}            & Proprietary \\
Gemini-2.5-Fl  & Google    & \url{https://ai.google.dev/gemini-api/docs/models/gemini-2.5-flash}      & Proprietary \\
Gemini-3-Fl    & Google    & \url{https://docs.cloud.google.com/gemini-enterprise-agent-platform/models/gemini/3-flash}      & Proprietary \\
\midrule
GLM-5.2 & Fireworks & \url{https://fireworks.ai/models/fireworks/glm-5p2} & \href{https://huggingface.co/datasets/choosealicense/licenses/blob/main/markdown/mit.md}{MIT}\\
GPT-OSS-120B & Fireworks & \url{https://fireworks.ai/models/fireworks/gpt-oss-120b} & \href{https://huggingface.co/datasets/choosealicense/licenses/blob/main/markdown/apache-2.0.md}{Apache-2.0}\\
Nemo-3.5-L & Fireworks & \url{https://fireworks.ai/models/fireworks/nemotron-lightning-3p5-30b-a3b} & \href{https://openmdw.ai/license/1-1/}{OpenMDW-1.1} \\
\midrule
Qwen3-Reranker-8B & Fireworks & \url{https://fireworks.ai/models/fireworks/qwen3-reranker-8b} & \href{https://huggingface.co/datasets/choosealicense/licenses/blob/main/markdown/apache-2.0.md}{Apache-2.0}\\
\midrule
text-embedding-3-large & OpenAI & \url{https://developers.openai.com/api/docs/models/text-embedding-3-large} & Proprietary \\
text-embedding-3-small & OpenAI & \url{https://developers.openai.com/api/docs/models/text-embedding-3-small} & Proprietary \\
\bottomrule
\end{tabular}
}
\caption{Models benchmarked in this work, along with their providers, source URLs, and licenses.}
\label{tab:models}
\end{table*}

\section{Dataset Source and Statistics}\label{app:dataset_source_info}
Table~\ref{tab:dataset_sources} summarizes all datasets used in this work along with their sources and licenses. We construct or adapt all datasets from publicly available sources. We construct the synthetic two-hop haystack entirely using iterative LLM prompting for distractor generation and manually curated needles, and we release it under CC BY 4.0. The NoLiMa-based dataset adapts the haystack construction procedure and needle set from \citet{modarressi2025nolima}, which is released under the Adobe Research License. We use \textsc{MultiHop-RAG} \citep{tang2024multihop} as released under the ODC-By license, and re-map gold evidence facts to line-level labels for EBG evaluation. We derive \textsc{QASPER-MP} from the \textsc{QASPER} dev split \citep{dasigi-etal-2021-qasper}, released under CC BY 4.0, and extend it to a multi-paper setting by pairing each gold paper with ten distractor papers and rewriting questions with short realistic paper cues.

\section{Model Information}\label{app:model_information}

We evaluate ten proprietary frontier and three open-weight models via their APIs. Table~\ref{tab:models} lists each model along with its provider, source URL, and license. All models were queried at temperature 0 (or the provider-equivalent default) with no additional fine-tuning or adaptation.

% This setup enables comparison across model families while minimizing variation introduced by decoding choices. We use the same evaluation protocol across models so that observed differences primarily reflect their long-context and retrieval capabilities.

\section{Artifact Use and Intended Use}\label{app:artifact_use}
We use all existing datasets and models consistently with their intended use. \textsc{MultiHop-RAG}, \textsc{QASPER}, and \textsc{NoLiMa} are used strictly for research evaluation, consistent with their release conditions. All proprietary models are accessed via official APIs in accordance with provider terms of service. All derived datasets introduced in this work are intended exclusively for research use and do not conflict with the access conditions of the original artifacts.

\section{Disclosure of AI Use}\label{app:disclaimer_of_ai_use}
We used AI assistants (large language models) at several stages of this work. We used them to review and debug code for our experimental pipelines and evaluation scripts, to polish and refine the writing of the manuscript by improving clarity, tightening phrasing, and catching grammatical errors, and to suggest alternative phrasings for titles, figure captions, and section headings. We developed all research ideas, experimental designs, benchmark constructions, analyses, and conclusions ourselves. We verified all AI-assisted code changes through testing, reviewed and edited all AI-assisted text, and take full responsibility for the final content of the paper.

% Same models/haystacks/pipelines as table_embedding_vs_fixedtopk.tex, but
% answer_metric only (not combined answer+evidence). Only one cell differs
% from the combined-accuracy table: claude-sonnet-4-5-20250929 at the 1M/200K
% column, Embedding-RAG, 62.5 (combined) -> 75.0 (answer-only) -- one test
% got the answer right but the cited evidence line wrong. Every other cell
% is identical between the two tables (answer-only >= combined, as expected).

% \clearpage
% \newpage
% \input{appendix}

\end{document}